\documentclass[draftcls,onecolumn,11pt]{IEEEtran}

\usepackage{amsmath,enumerate}
\usepackage{cite}
\usepackage{lipsum}
\usepackage[justification=centering]{caption}
\usepackage{amssymb,graphicx,subfigure}
\usepackage[bottom]{footmisc}
\usepackage[dvipsnames]{xcolor}

\usepackage{array,multirow}
\newlength{\Oldarrayrulewidth}

\newtheorem{thm}{Theorem}[section]
\newtheorem{lem}[thm]{Lemma}

\usepackage{algorithm,algpseudocode}
\algnewcommand\Input{\item[{\textbf{Input:}}]}
\algnewcommand\Output{\item[{\textbf{Output:}}]}
\algnewcommand\Initialize{\item[{\textbf{Initialize:}}]}
\algnewcommand{\return}[1]{
  \State \textbf{return:}
  \Statex \hspace*{\algorithmicindent}\parbox[t]{.8\linewidth}{\raggedright #1}
}

\newcommand{\supp}{\operatorname{supp}}

\newcommand{\abs}{\operatorname{abs}}

\begin{document}

\title{Tree Search Network for Sparse Regression}

\author{Kyung-Su~Kim,~Sae-Young Chung\\
School of Electrical Engineering, Korea Advanced Institute of Science and Technology\\
Email: {kyungsukim, schung}@kaist.ac.kr}


%

%


\maketitle

\begin{abstract}
We consider the classical sparse regression problem of recovering a sparse signal $x_0$ given a measurement vector $y = \Phi x_0+w$.
We propose a tree search algorithm driven by the deep neural network for sparse regression (TSN).  TSN improves the signal reconstruction performance of the deep neural network designed for sparse regression by performing a tree search with pruning. 
It is observed in both noiseless and noisy cases, TSN recovers synthetic and real signals with lower complexity than a conventional tree search and is superior to existing algorithms by a large margin for various types of the sensing matrix $\Phi$, widely used in sparse regression.
\end{abstract}

\begin{IEEEkeywords}
sparse regression, deep neural network, tree search, extended support estimation, long short-term memory
\end{IEEEkeywords}

\section{Introduction}
\label{intr} 
The sparse linear regression (SR), referred to as compressed sensing (CS) \cite{donoho2006compressed}, has received much attention in many machine learning and signal processing applications\footnote{Some examples are feature selection \cite{JRD16,kale2017adaptive}, signal reconstruction \cite{zhou2017sparse,caiafa2017unified,metzler2017learned}, denoising \cite{metzler2016denoising}, and super-resolution imaging \cite{heckel2016super,dai2017sparse}.}. Its goal is to recover $k$-sparse\footnote{$x_0$ is $k$-sparse if it has at most $k$ nonzero elements.} signal vector $x_0\in \mathbb{K}^{n}$ and its support $\Omega$ from under-sampled measurement vector $y \in \mathbb{K}^{m}$ such that $m \leq n$ and $y = \Phi x_0+w$, where $\Omega$ denotes the set of nonzero elements in $x_0$, $\Phi \in \mathbb{K}^{m \times n}$ is a known sensing matrix, and $w \in \mathbb{K}^{m}$ is the noise vector. 

\subsection{Existing deep neural network for SR}\label{aam1}
Deep neural networks (DNNs) have contributed to notable performance improvements in fields such as image processing \cite{he2016deep,ren2017faster}, natural language processing  \cite{chung2016character}, and reinforcement learning \cite{mnih2015human}. Consequently, intense research has been devoted to the development of a tailored DNN for SR (DNN-SR) to estimate sparse signals; it outputs an estimate of $k$-sparse signal $x_0$ or its support $\Omega$ from measurement vector $y$.  Gregor and LeCun proposed a DNN-SR structure by observing that the iterations in each layer of a feed-forward neural network can represent the update step in an existing SR algorithm called iterative shrinkage and thresholding (ISTA)  \cite{gregor2010learning}. Subsequent studies addressed variants of the DNN-SR architecture based on the unfolding process in the context of this algorithm to improve performance by incorporating a nonlinear activation function \cite{kamilov2016learning,mahapatra2017deep}, reducing the training complexity using shared parameters over the DNN layers \cite{hershey2014deep}, and developing a structured sparse and low-rank model \cite{sprechmann2015learning}. In addition, there are some recent works to study theoretical properties of learned variants of ISTA \cite{moreau2017understanding,giryes2018tradeoffs,chen2018theoretical}. Similarly, learned variants of approximate message passing (AMP) for SR have been studied \cite{tramel2016approximate,borgerding2017amp,metzler2017learned} by exploiting the unfolding process. On the other hand, a DNN-SR structure based on the alternating direction method of multipliers was proposed for generating a transform matrix to filter magnetic resonance imaging data \cite{sun2016deep,yang2017admm} and a DNN-SR architecture based on a generative model has been proposed \cite{mardani2017deep,bora2017compressed}. Besides, the correlation among the different sparse vectors was used in the DNN-SR proposed in \cite{palangi2016distributed,palangi2016exploiting} when multiple measurement vectors share a common support. Lately, He et al. suggested \cite{he2017bayesian} that the update step for sparse Bayesian learning (SBL) \cite{wipf2004sparse} can be formed into a gated feedback long short-term memory (GFLSTM) network, which is a widely used recurrent DNN structure \cite{chung2015gated}. This DNN-SR structure showed a comparable performance to that of SBL in the case when the sensing matrix consists of columns with high correlation.  

\subsection{Scope and contribution}

As the studies mentioned in Section \ref{aam1} showed, DNN can be exploited to solve the SR problem and demonstrate potential to outperform existing SR algorithms. However, there is not enough research showing that DNN-SR is not limited to image processing and enables uniform recovery of synthesis sparse signals with better performance than existing SR methods. On the other hand, deep learning combined with optimization techniques based on tree search has shown better performance than existing methods without tree search. Typical examples include AlphaGo \cite{silver2016mastering}, which applies Monte Carlo tree search to deep reinforcement learning, and a deep reinforcement neural network \cite{guo2014deep} trained by data generated from an offline Monte Carlo tree search to outperform deep Q-networks. 
Motivated by these works, we first propose a tree search algorithm driven by deep neural network for SR (TSN) to improve the performance of DNN-SR.\footnote{Once the support is determined, the problem of estimating $x_0$ reduces to a standard overdetermined linear inverse problem, which can be easily solved. Therefore, we focus on a type of DNN-SR that recovers the true support $\Omega$, such that it takes $y$ as its input and outputs a probability vector $v:=(v_1,...,v_n)$ whose $|\Omega|$-largest indices represent the estimate of $\Omega$. TSN performs a tree search to find support $\Omega$ of $x_0$ based on a trained DNN-SR of the abovementioned type.} 

Note that the tree search in TSN is applied to DNN-SR as a post-processing framework. That is, we trained a single DNN-SR  independent of the tree search and used the single network to generate all the nodes in the search tree in TSN. TSN has the following three main features, i.e., DNN-based index selection for tree search and pruning the tree.

\begin{figure}
  \begin{center}
    \includegraphics[width=8cm, height=5cm]{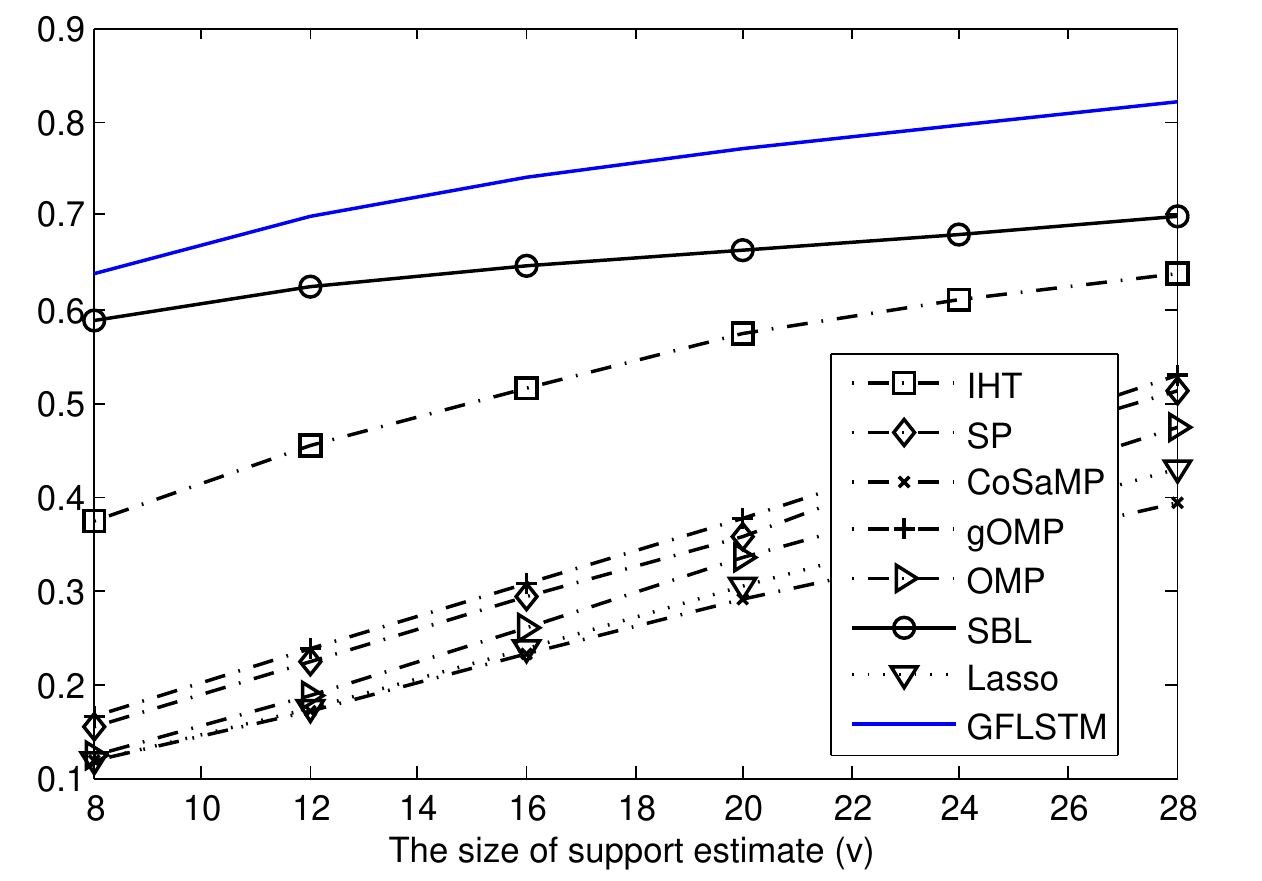}
  \end{center}
\footnotesize
  \caption{\footnotesize Support recovery rate of existing SR algorithms in terms of $\mathbb{E}(|\Omega \cap \Sigma|/|\Omega|)$, where $\Sigma$ is a support estimate of size $v$, obtained by each algorithm. We note that GFLSTM \cite{he2017bayesian} outperforms others irrespective of $v$. The simulation setting is the same as that in the noiseless case when $|\Omega|=8$ in Section \ref{rvm}.}
\label{intro_fig}
\end{figure}
\begin{itemize}
\item 
(The DNN-based index selection for tree search) In the conventional tree search, e.g., multipath matching pursuit (MMP) \cite{kwon2014multipath}, to estimate the support, each parent node representing a partial support estimate generates its child nodes through the index selection based on orthogonal matching pursuit (OMP): selecting indices according to the largest correlations measured by the inner product with the residual vector. In this study, we consider the DNN-based index selection in addition to the OMP-based index selection for generating  child nodes. 
Figure \ref{intro_fig} shows that the percentage of true indices among the support estimate of size $v$, obtained by the trained DNN-SR, is higher than those using other conventional SR methods. This implies that a true index can be included in the child nodes with a higher probability by using the DNN-based index selection than the OMP-based approach. {He et al. showed in \cite{he2017bayesian} that the percentage of true indices among the m-largest predicted DNN outputs (the loose accuracy) is significantly higher than those obtained by using other SR methods. This observation also supports our argument.} Thus, given that the DNN-based tree search can find the support with a few child nodes, combining DNN and tree search can improve the performance over using them separately.

\begin{figure*}
\begin{center}
{\includegraphics[width=15cm,height=6.5cm]{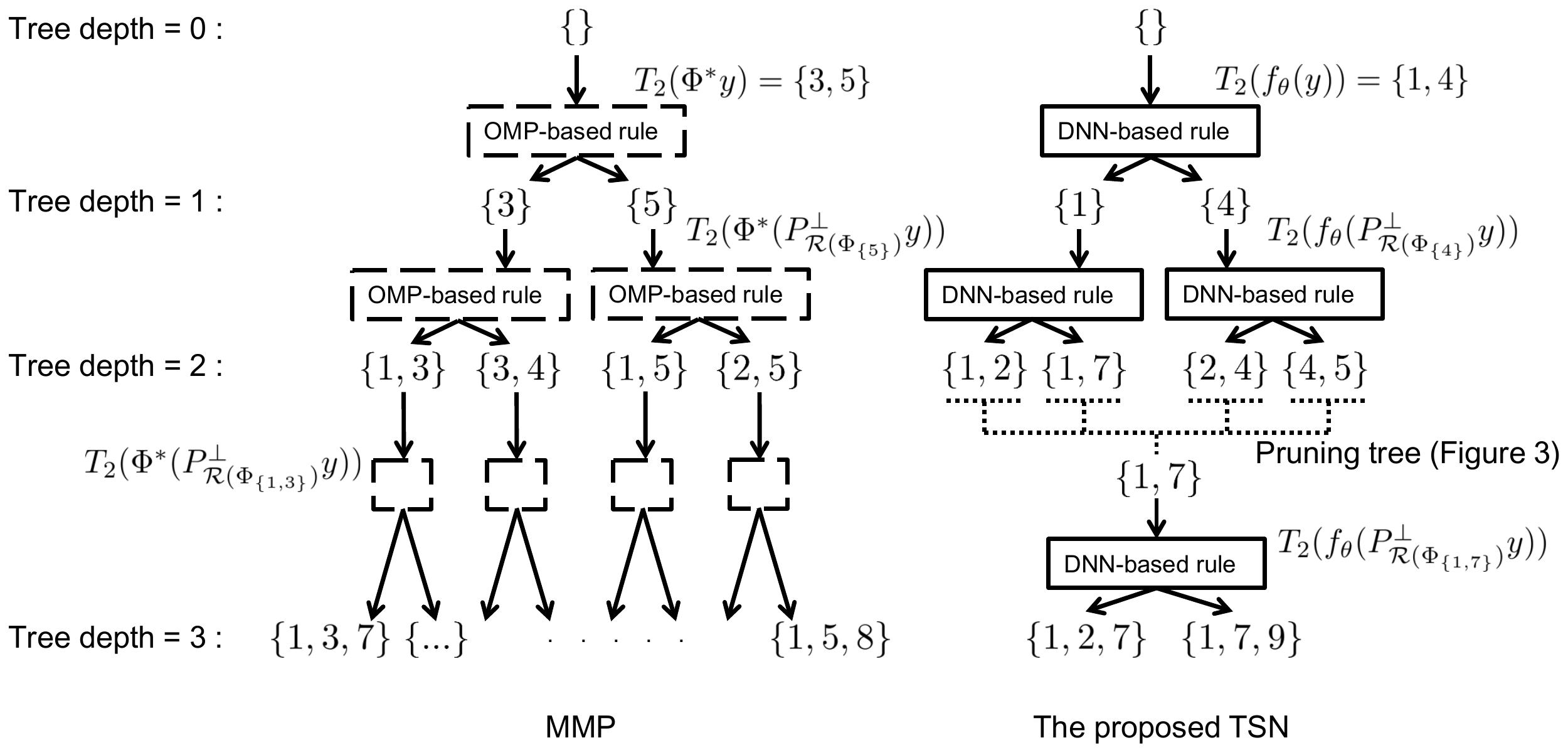}}
\caption{Comparison of the proposed TSN to the convectional tree search for SR (MMP) in the case when $q \,(=2)$ child nodes are generated at each parent node}
\label{fig_comp}
\end{center}
\end{figure*}

\item (Pruning the search tree) Suppose that $q$ child nodes are generated at each parent node of the search tree. Then, the number of nodes in the tree exponentially increases with the tree depth $d$ to approximately $\sum_{i=0}^{d-1}q^i$. Hence, to reduce the complexity of the tree search, we propose a pruning method to remove leaf nodes at certain depths, except for those showing the smallest signal errors. For instance, preserving one node at every depth $t \ll d$ through the pruning method, the number of nodes in the resulting tree with depth $d$ is approximately $(d/t) \cdot (\sum_{i=0}^{t-1} q^i)$. Given that the maximum exponent of $q$ decreases from $d$ to a constant $t$, the searching complexity is efficiently reduced, and it is linearly dependent on depth $d$.

\item (Generating the extended support estimation $\Psi$  of size $m-1$ via DNN-SR) Note that each node in the search tree in TSN has a partial support estimate $\Delta$ generated from its parent node via the DNN-based index selection. Then every node in TSN generates a support estimate of size $k$ by utilizing its parital support estimate $\Delta$ and the trained DNN-SR. To do this, each node generates an extended support estimate $\Psi$ of size $m-1$ via the DNN-based index selection at the first stage, and estimate the support by selecting $ k $ indices out of the set at the second stage. This two-stage process is a variant of the two-stage process shown in \cite{kim2019greedy}; the existing process selects an extended support estimate $\Psi$ of $m-1$ indices based on OMP, whereas the proposed process is based on DNN.

It is guaranteed theoretically and experimentally in \cite{kim2019greedy} that exploiting the extended support estimate $\Psi$ of size $m-1$, generated by the OMP-based rule, improves the performance for the support recovery in comparison to  the case without considering it, i.e., the case when the size of $\Psi$  is equal to $k$.\footnote{\cite{kim2019greedy} shows that this re-estimation method using an extended support estimate of size $ m-1 $ is guaranteed to further mitigate the sufficient conditions for SR algorithms based on the OMP-based index selection to restore the sparse signal.} We accepted this principle, i.e., selecting the $ m-1 $ incides instead of $k$ incides, for estimating the support at each node, but used a different index selection technique, i.e., the DNN-based index selection. As shown in Figure \ref{intro_fig} and experimental results in \cite{he2017bayesian}, the DNN-based index selection has shown better performance for the loose accuracy and lower complexity than OMP-based approaches. This supports our claim that combining DNN and the extended support estimation of size $m-1$ improves the recovery performance for the sparse signal. We demonstrated it in Section \ref{rvm}. 
\end{itemize}

We utilize the GFLSTM \cite{he2017bayesian} or learned vector AMP (LVAMP) \cite{borgerding2017amp} as the DNN-SR used in TSN to demonstrate that TSN improves the performance of a typical DNN-SR.
Experimental results also suggest that TSN significantly improves the recovery performance of DNN-SR, has a lower complexity than the conventional tree search, and outperforms existing SR algorithms in both noiseless and noisy cases with various types of the sensing matrix $\Phi$. For example, the maximal sparsity (i.e., the maximal size $|\Omega|$ of target support) to uniformly recover $x_0$ using TSN is two times larger than those using SBL and the GFLSTM network from simulations with a Gaussian sensing matrix of dimension $20 \times 100$ and noiseless measurements. These tests were based on all synthesis sparse signals. It suggests that TSN can improve performance in various domains using SR. In this paper, to provide its examples, we evaluated performance of nonorthogonal multiple access (NOMA) in communication system and image restoration without using image training data, and showed the superiority of TSN.

\section{Notation}\label{not}
$\mathbb{K}$ denotes the real $\mathbb{R}$ or complex $\mathbb{C}$ field and  $\mathbb{N}$ denotes the set of natural numbers. The set $\{i,i+1,...,j\}$ is denoted by $\{i:j\}$. For a matrix $A:=[a_1,...,a_n] \in \mathbb{K}^{m \times n}$, submatrices of $A$ with columns indexed by $J \subseteq \{1:n\}$ and rows indexed by $Q \subseteq \{1:m\}$ are denoted by $A_J$ and $A^Q$, respectively.  $\mathcal{R}(A)$ denotes the range space spanned by the columns of $A$. $A^{\top}$ ($A^{*}$) denotes the (Hermitian) transpose of $A$. $P^{\perp}_{\mathcal{R}(A)}$ denotes the projection matrix onto the orthogonal complement of $\mathcal{R}(A)$. For a vector $v:=(v_1,...,v_n) \in \mathbb{K}^{n}$, $\abs(v)$ denotes a vector $z$ whose element $z_i$ is the absolute value of $v_i$ and the Frobenius norm of $v$ is denoted by $\left \| v \right \|$. $\supp(v)$, the support of $v$, denotes the index set of nonzero elements in $v$. $T_l(v)$ is the operator whose output is an index set of size $l$ with the $l$-largest absolute values in the input vector $v:=(v_1,...,v_n)$.

\section{High-level description of TSN}
\label{hld}
The propposed TSN recovers the target signal $x_0$ and its support $\Omega$ given $y$ and $\Phi$ by utilizing a trained DNN-SR defined by function $f_{\theta}(\cdot): \mathbb{K}^{m} \rightarrow \mathbb{T}^n$ as its input, where $\theta$ is the set of training parameters in the network and $\mathbb{T}^n$ represents $(n-1)$-dimensional probability simplex $\{v\in \mathbb {R} ^{n}| v_{1}+\dots +v_{n}=1 \textup{ and } v_{i}\geq 0 \textup{ for }i \in \{1:n\} \}$. For sparse vector $z \in \mathbb{K}^{n}$ and measurement vector $h=\Phi z \in \mathbb{K}^{m}$, given sensing matrix $\Phi$, function $f_{\theta}(\cdot)$ takes vector $h$ as its input and is trained to return vector $v =(v_1,...,v_n)= f_{\theta}(h) \in \mathbb{T}^n$ such that $v_i = 1/|\supp(z)|$ for $i \in \supp(z)$ and $v_i = 0$ for $i \notin \supp(z)$. Each element $v_i$ of vector $v$ indicates the probability that index $i$ belongs to the support of $z$. For instance, if $n$ is 4 and the support of $z$ is $\{1,3\}$, function $f_{\theta}(\cdot)$ is trained to return output vector $f_{\theta}(h)$ equal to $(1/2,0,1/2,0)$. The detailed process to train DNN $f_{\theta}(\cdot)$ used in TSN and its effect are shown in Appendix \ref{tr1}.

In the rest of this section, we introduce main features of TSN to estimate the support $\Omega$. The detailed process of TSN is shown in Algorithm \ref{alg5} in Appendix \ref{tsns}. 

\begin{figure*}
\begin{center}
{\includegraphics[width=15cm,height=6cm]{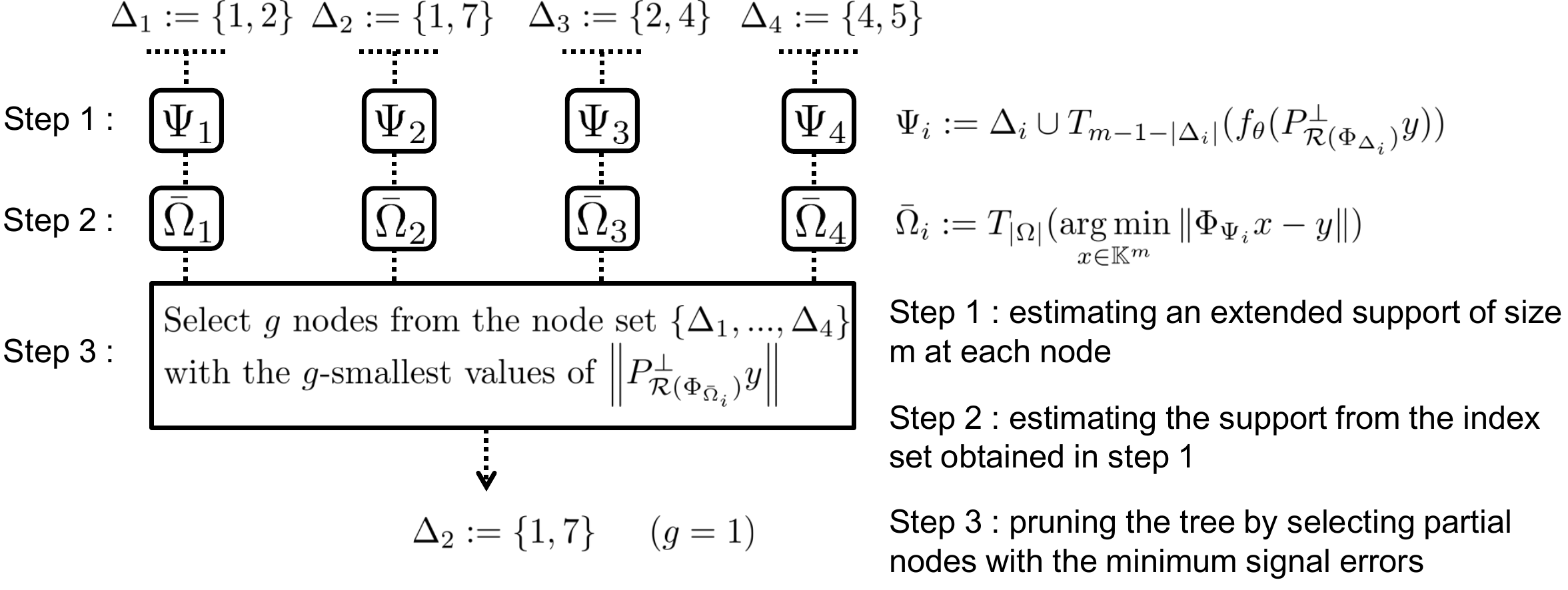}}
\caption{Description of the pruning algorithm in TSN for the noiseless case when the number $g$ of the remaining nodes after pruning is set to 1}
\label{fig_prune}
\end{center}
\end{figure*}

\subsection{The DNN-based index selection in TSN} 
\label{fig_comp_exp}

Suppose that there exists a partial support estimate $\Delta$ of the target signal $x_0$, which corresponds to a parent node of the tree. Then, to find the remaining support $\Omega \setminus \Delta$ outside $\Delta$, we expand a branch from the node by generating multiple partial support estimates (child nodes) $\{\Delta_{1},\Delta_{2},...,\Delta_{q}\}$ given $\Delta$, where $q$ represents the number of child nodes. Each index set $\Delta_{i} = h_i \cup \Delta$ for $i \in \{1:q\}$ is obtained by adding element $h_i$ to $\Delta$ where $\Theta := \{h_1,...,h_q\}$ is an estimate of the remaining support $\Omega \setminus \Delta$ with size $q$.  This estimate is obtained by considering the residual vector $P^{\perp}_{\mathcal{R}(\Phi_{\Delta})}y$, the orthogonal complement of the columns in $\Phi$, indexed by $\Delta$ onto $y$,  as the trained DNN-SR input and selecting the $q$-largest elements $T_q(f_{\theta}(P^{\perp}_{\mathcal{R}(\Phi_{\Delta})}y))$ of its output. Figure \ref{fig_comp} shows an example of the DNN-based index selection in TSN compared to the conventional tree search for SR, e.g., MMP, when $q$ is set to $2$. Suppose that the parent node $\Delta$ is $\{5\}$ at the tree depth $1$ of MMP in Figure \ref{fig_comp}. Then, MMP generates $q\, (=2)$ child nodes as $\{1,5\}$ and $\{2,5\}$ by using the OMP-based index selection $T_q(\Phi^* P^{\perp}_{\mathcal{R}(\Phi_{\Delta})}y) = \{1,2\}$. Conversely, TSN generates $q$ child nodes as $\{2,4\}$ and $\{4,5\}$ by using the DNN-based index selection $T_q(f_{\theta}(P^{\perp}_{\mathcal{R}(\Phi_{\Delta})}y)) = \{2,5\}$  when  the parent node $\Delta$ is $\{4\}$.

Note that the DNN-based index selection uses one common DNN-SR $f_{\theta}(P^{\perp}_{\mathcal{R}(\Phi_{\Delta})}y)$ to create child nodes from each parent node with a different partial support estimate $\Delta$. In order for this trained DNN-SR to provide the remaining support $\Omega \setminus \Delta$ irrespective of the set $\Delta$ given at each parent node, we learn the DNN-SR $f_{\theta}(\cdot)$ by using Algorithm \ref{alg1} such that the following condition (\ref{abc_eq}) is satisfied for any index set $\Gamma \subseteq \{1:n\}$
\begin{align}\label{abc_eq}
\underset{\tilde x \in \mathbb{K}^n}{\min} \left \| P^{\perp}_{\mathcal{R}(\Phi_{\Gamma})}y - \Phi_{D}\tilde x^{D} \right \|=0,
\end{align}  
where $D:= T_q(f_{\theta}(P^{\perp}_{\mathcal{R}(\Phi_{\Gamma})}y))$ is the index set of size $q$, obtained by the DNN-based index selection. Given that $D$ satisfying condition (\ref{abc_eq}) where $\Gamma=\Delta$ includes the true remaining support $\Omega \setminus \Delta$ from Lemma \ref{lem1}, therefore, a true remaining index, which is not in the parent node, is added to $|\Omega \setminus \Delta|$ of its $q$ child nodes if the DNN-SR is ideally trained. The proof of Lemma \ref{lem1} is shown in Appendix \ref{apena}.

\begin{lem}\label{lem1}
Suppose that $|\Omega|<m$ and every $m$ columns in $\Phi$ exhibit full rank. For any pair $(D,\Gamma)$ of index sets in $\{1:n\}$ satisfying (\ref{abc_eq}) such that $|D| < m$, if $P^{\perp}_{\mathcal{R}(\Phi_{\Gamma})}y$ is uniformly sampled from $\mathcal{R}(\Phi_{\Omega \setminus \Gamma})$, the index set $D$ includes $\Omega \setminus \Gamma$ ($D \supseteq \Omega \setminus \Gamma$) almost surely. 
\end{lem}

\subsection{Pruning the search tree in TSN}

\label{fig_prune_exp}
TSN prunes the search tree to reduce its complexity, as its example is shown in Figures \ref{fig_comp} and \ref{fig_prune}. In Figure \ref{fig_comp}, pruning the tree in TSN is executed by remaining $g$ ($=$1) nodes $(\Delta_2)$ having the minimum signal errors among all nodes $\Delta_1:=\{1,2\},\Delta_2:=\{1,7\},\Delta_3:=\{2,4\},\Delta_4:=\{4,5\}$ given at tree depth $2$. This process consists of the following three steps, which are illustrated in Figure \ref{fig_prune}.\footnote{In Figure \ref{fig_prune}, an input $z$ of TSN, detailed in  Appendix \ref{tsns}, is set to 1.}

In the first step, an extended support estimate $\Psi_i:=\Delta \cup T_{m-1-|\Delta|}(f_{\theta}(P^{\perp}_{\mathcal{R}(\Phi_{\Delta})}y))$ of size $m-1$ is generated at each node $\Delta_i$ by using the trained DNN-SR and  the residual vector $P^{\perp}_{\mathcal{R}(\Phi_{\Delta})}y$. In the second step, a $k$-support estimate\footnote{The $k$-support, $\Omega(k)$, of $x_0$ denotes any index set satisfying $\Omega(k) \supseteq \Omega$ and $|\Omega(k)|=k$.} $\bar \Omega_i$ is obtained by selecting $k$ indices from each extended support estimate $\Psi_i$ through the ridge regression suggested in \cite{wipf2004sparse}. A detailed description of the ridge regression is shown in Appendix \ref{mwhy}.  In the noiseless case ($w=0$), given that this ridge regression is equal to the least-squares regression, $\bar \Omega_i$ is obtained by $T_k(\underset{x \in \mathbb{K}^n}{\arg\min} \left\| \Phi_{\Psi_i}x^{\Psi_i}-y\right\|)$. In the third step, the signal error of each node is calculated as the residual norm $\left\| P^{\perp}_{\mathcal{R}(\Phi_{\bar \Omega_i})}y \right\|$ by using the $k$-support estimate $\bar \Omega_i$, and then $g$ nodes having the minimum errors are selected.

The least-squares method ($\underset{x \in \mathbb{K}^{n}}{\arg\min} \left\| \Phi_{\Psi_i}x^{\Psi_i}-y\right\|$) provides $x_0$ as the unique solution in the noiseless case if $ \Psi_i \supseteq \Omega$, $|\Psi_i| < m$, and $\Phi_{\Psi_i}$ has full column rank. Note that the probability of satisfying $\Psi_i \supseteq \Omega$ increases with the dimension of $\Psi_i$. In addition, if $ \Psi_i \supseteq \Omega$ holds, the inversion problem of SR can be simplified as a problem where $\Phi$ is replaced by submatrix $\Phi_{\Psi_i}$. Therefore, we set the dimension of extended support estimate $\Psi_i$ to $m-1$; a further detailed information is shown in Appendix \ref{mwhy}.

If the extended support estimate $\Psi_i$ of $m-1$ is obtained by the OMP-based index selection shown in \cite{kim2019greedy}, then the first and second steps correspond to an existing SR algorithm called two-stage orthogonal
subspace matching pursuit with sparse Bayesian learning (TSML). Therefore, these two steps can be interpreted as a modification of TSML using DNN. The experimental results in Section \ref{rvm} show that this modification reduces the complexity of TSN and improves performance, thus demonstrating its validity.


\section{Numerical experiments}
\label{ne}
In this section, we verify the performance of TSN against some conventional SR algorithms, namely, generalized orthogonal matching pursuit (gOMP) \cite{wang2012generalized}\footnote{ Three indices were selected per iteration in gOMP.}, compressive sampling matched pursuit (CoSaMP) \cite{blanchard2014greedy}, subspace pursuit (SP) \cite{dai2009subspace}, iterative hard thresholding (IHT) \cite{blumensath2009iterative}, MMP \cite{kwon2014multipath}\footnote{MMP$_1$ and MMP$_2$ shown in this section are the MMP depth first (MMP-DF) algorithms \cite{kwon2014multipath} such that the number $N_{\textup{max}}$ of path candidates are $L^{s}$ and $500$, respectively, with the same expansion number $L$ equal to $4$. Thus, MMP$_1$ is the MMP-DF with full tree search.}, SBL \cite{wipf2004sparse,zhang2011sparse}, and basis pursuit denoising (Lasso) \cite{chen2001atomic}. 
We also compared TSN to some state of the art DNN-based SR algorithms, namely, GFLSTM \cite{he2017bayesian}, a learned variant of ISTA (LISTA) \cite{gregor2010learning}, learned AMP (LAMP) \cite{borgerding2017amp}, and LVAMP \cite{borgerding2017amp}. 
 We set the DNN-SR used in TSN to GFLSTM to evaluate that TSN improves the performance of its target DNN-SR.

\subsection{Real-valued case}
\label{rvm}
\subsubsection{Experimental settings}
\label{se}
Let $\mathcal{N}(a,b)$ denote the real Gaussian distribution with mean $a$ and variance $b$. We sample $\Phi \in \mathbb{K}^{m \times n}$ such that its elements independently follow $\mathcal{N}(0,1)$ and its columns are $l_2$-normalized. Support $\Omega$ of $x_0$ is generated from a uniform distribution, and signal vector $x_0$ is obtained such that each of its nonzero elements is independently and uniformly sampled from $-1$ to $1$, excluding the interval from $-$0.1 to 0.1. To consider noisy signals, the average signal-to-noise ratio (SNR) per sample ($\textup{SNR} := \mathbb{E}\left \| \Phi x_0 \right \|^2  / \, \mathbb{E}\left \|w \right \|^2 $) is defined as the ratio between the power of the measured signal and that of noise.
Each element of the noise vector $w$ follows $\mathcal{N}(0,\sigma^2_w)$, where $\sigma_w$ is dependent on the given SNR environment.

For the DNN-SR $f_\theta(\cdot)$ used in TSN, we used GFLSTM network for sparse regression, as suggested in \cite{he2017bayesian}, where the hidden unit size, the number of unfolding steps, and the layer size are set to $425$, $11$, and $2$, respectively. Further details on this network are provided in \cite{he2017bayesian}.
To train the GFLSTM $f_\theta(\cdot)$, we used Algorithm \ref{alg1} in Appendix \ref{tr1} whose input $(k_1,k_2,s_d,s_b,n_e)$ and $v_{\textup{SNR$_{\textup{dB}}$}}$ are set to $(1,10,6 \cdot 10^5,250,400)$ and a fixed SNR in decibels (dB), respectively. We used RMSprop optimization with learning rate $\eta_i$ of $0.001$ for epoch $i \leq 250$ and $0.001/4^j$ for epoch $i$ from $201+50j$ to $250+50j$ ($j \in \{1:3\}$). 

\begin{figure*}[t]
\scriptsize
\begin{center}
\subfigure[$\mathbb{P}( \hat x  = x_0 )$]{\includegraphics[width=6.2cm, height=3.9cm]{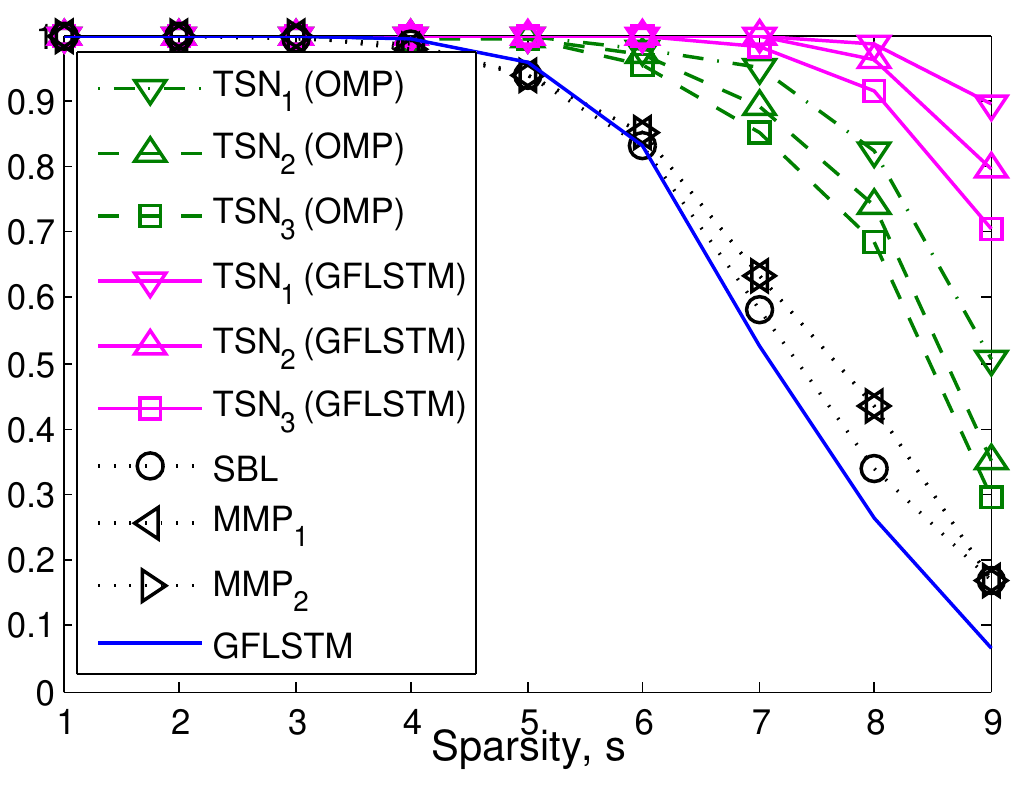}}
\subfigure[Execution time]{\includegraphics[width=6.2cm, height=3.9cm]{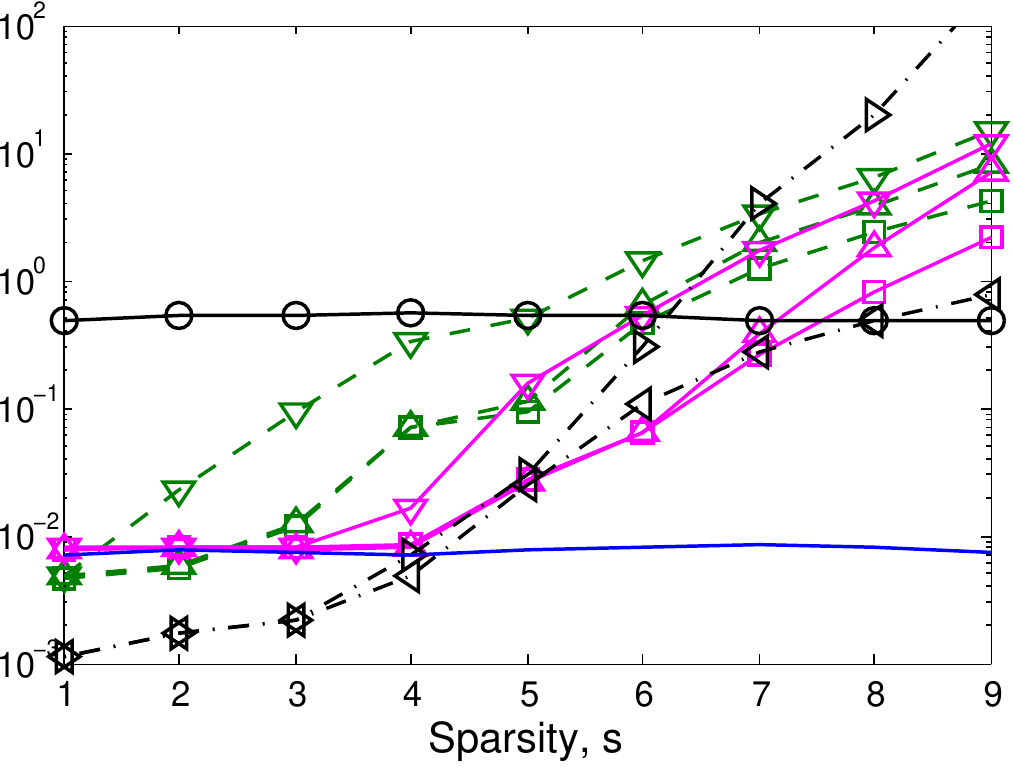}}
\subfigure[$\mathbb{P}( \hat x  = x_0 )$]{\includegraphics[width=6.2cm, height=3.9cm]{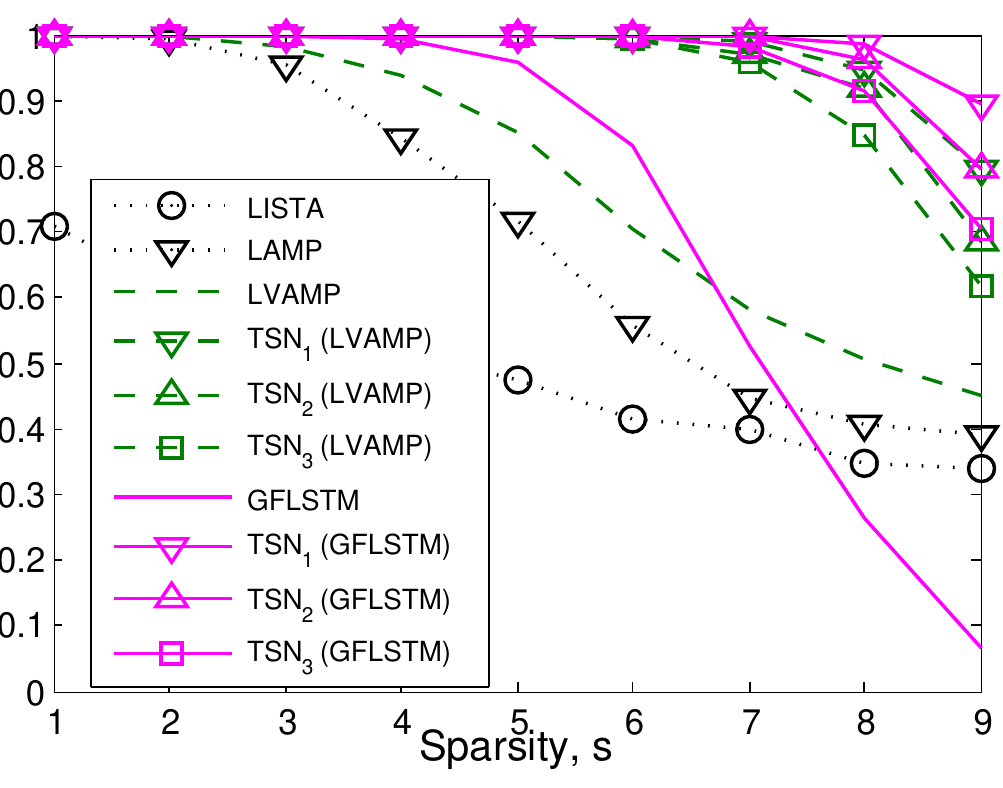}}
\subfigure[Execution time]{\includegraphics[width=6.2cm, height=3.9cm]{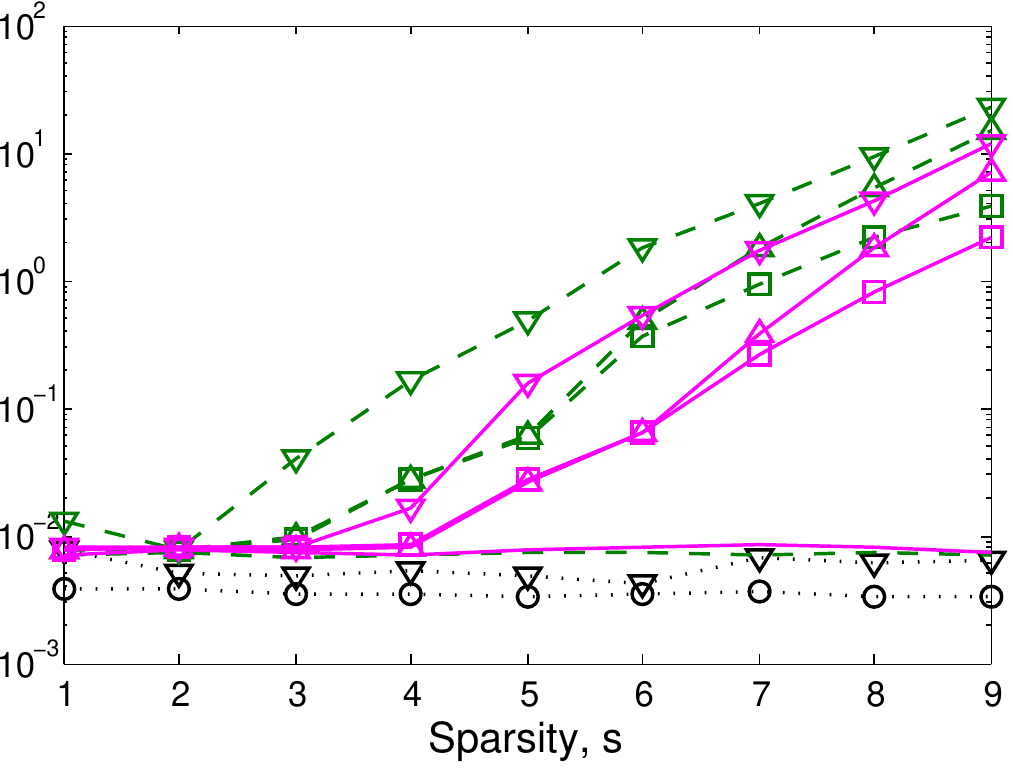}}
\caption{Performance comparison of TSN without or with DNN-based index selection and other state-of-the-art  DNN-based SR algorithms in the noiseless case.}
\label{ompreal}
\end{center}
\end{figure*} 
\begin{figure*}[t]
\scriptsize
\begin{center}
\subfigure[$\mathbb{P}( \hat x  = x_0 )$]{\includegraphics[width=3.7cm, height=3.9cm]{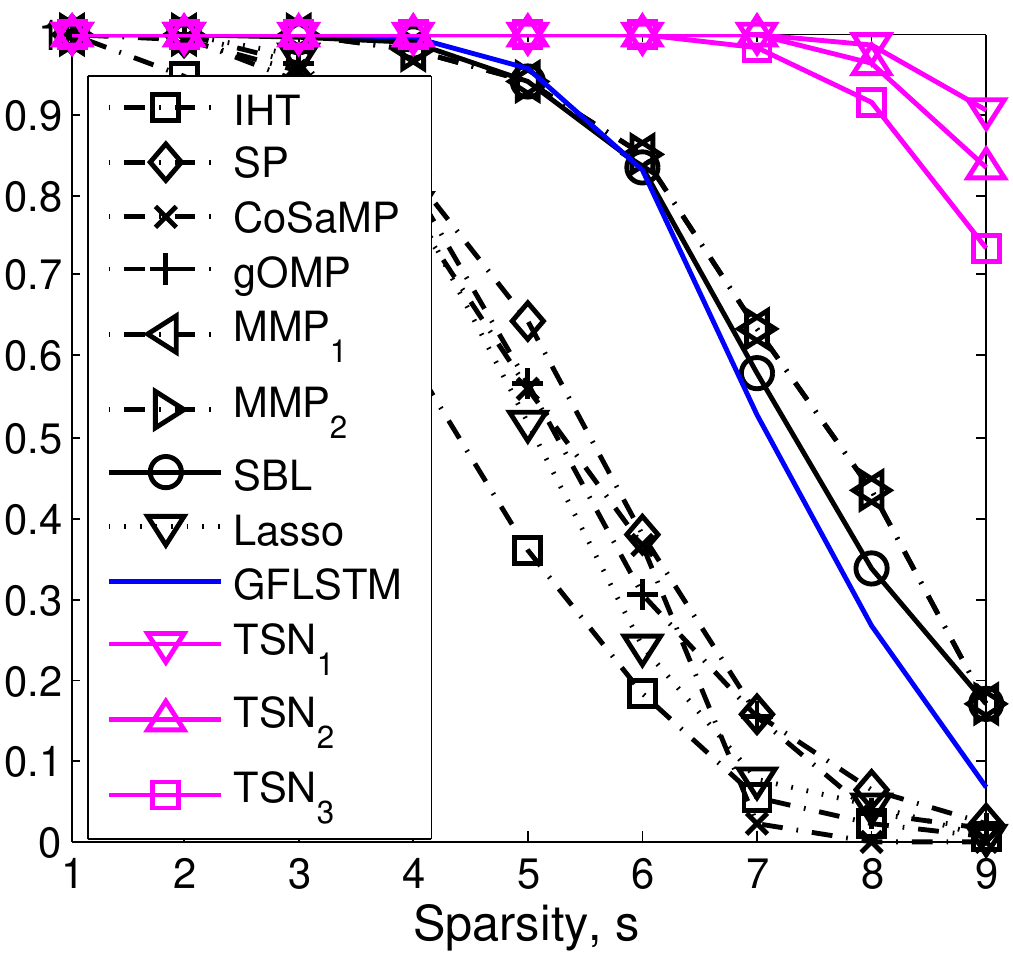}}
\subfigure[$\mathbb{E}(\frac{\left \| x_0 - \hat x\right \|}{\left \| x_0 \right \|} )$]{\includegraphics[width=3.7cm, height=3.9cm]{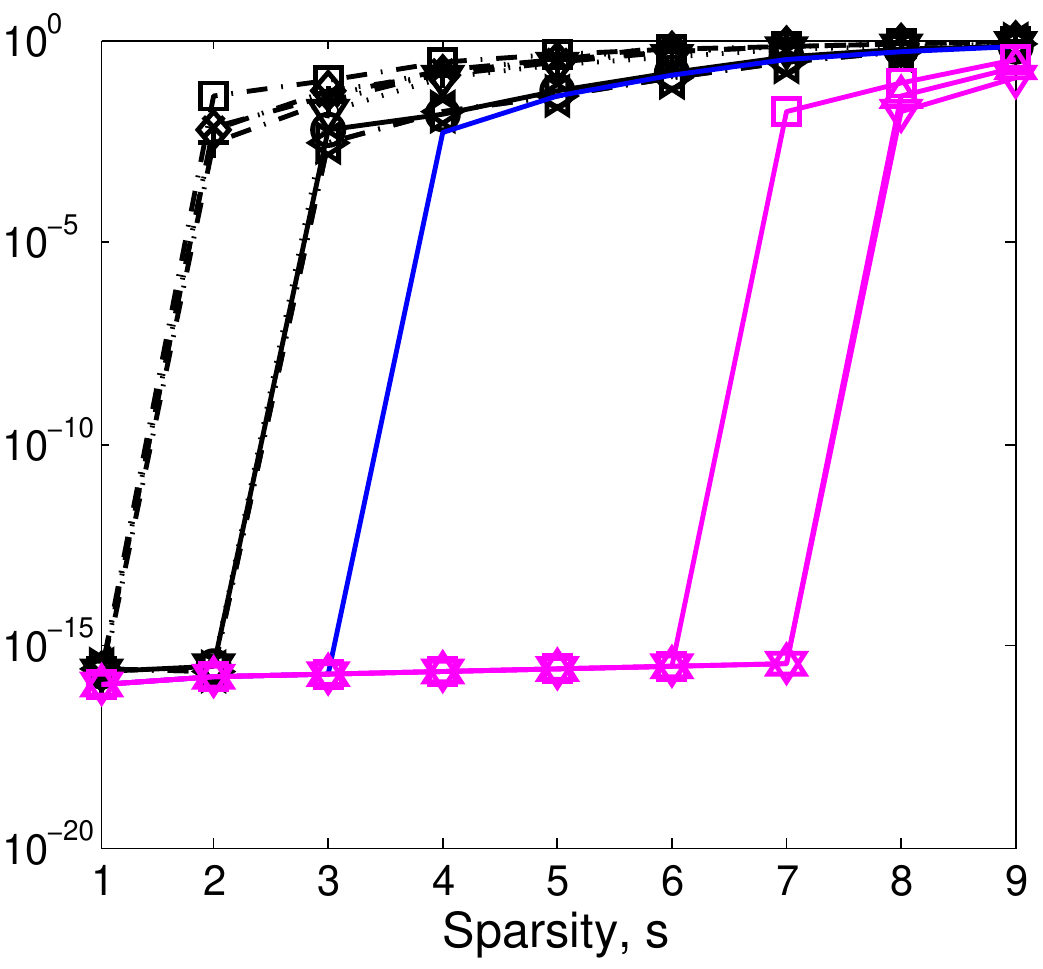}}
\subfigure[Execution time]{\includegraphics[width=3.7cm, height=3.9cm]{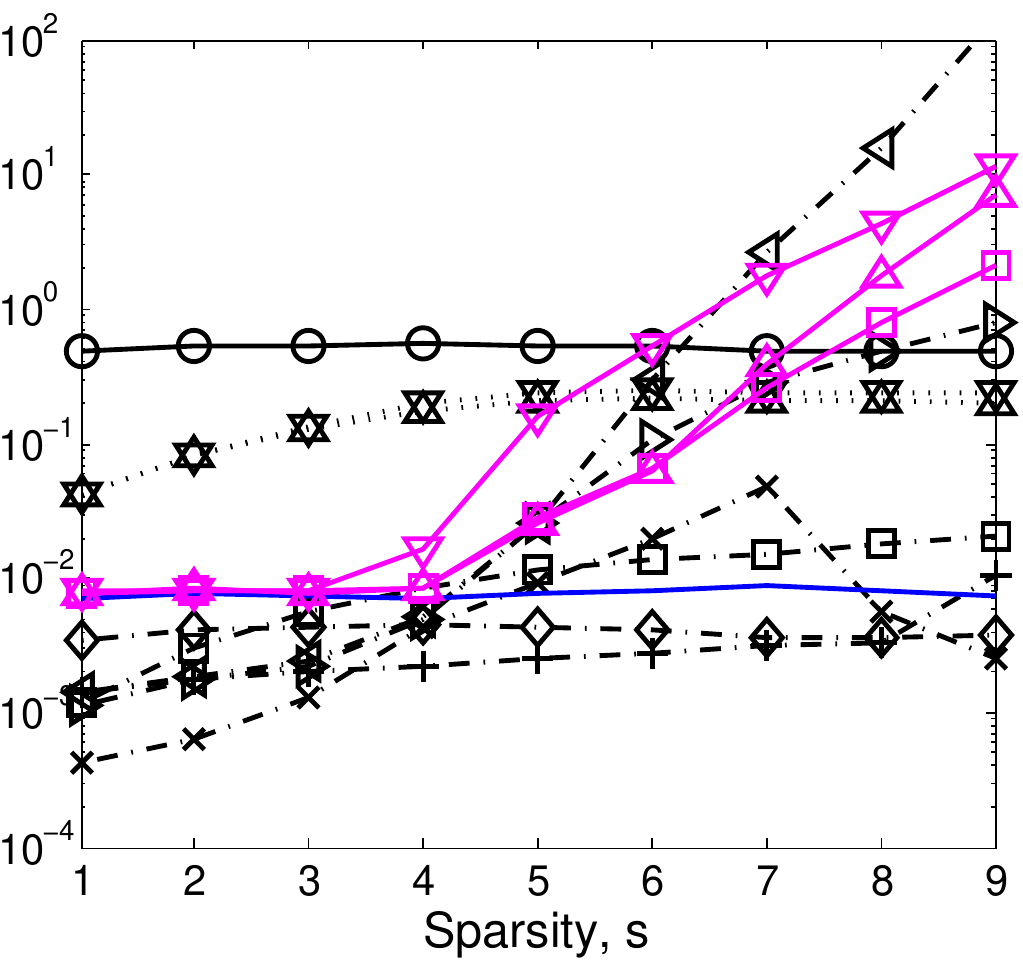}}
\subfigure[$\mathbb{E}(\frac{\left \| x_0 - \hat x\right \|}{\left \| x_0 \right \|} )$]{\includegraphics[width=3.7cm, height=3.9cm]{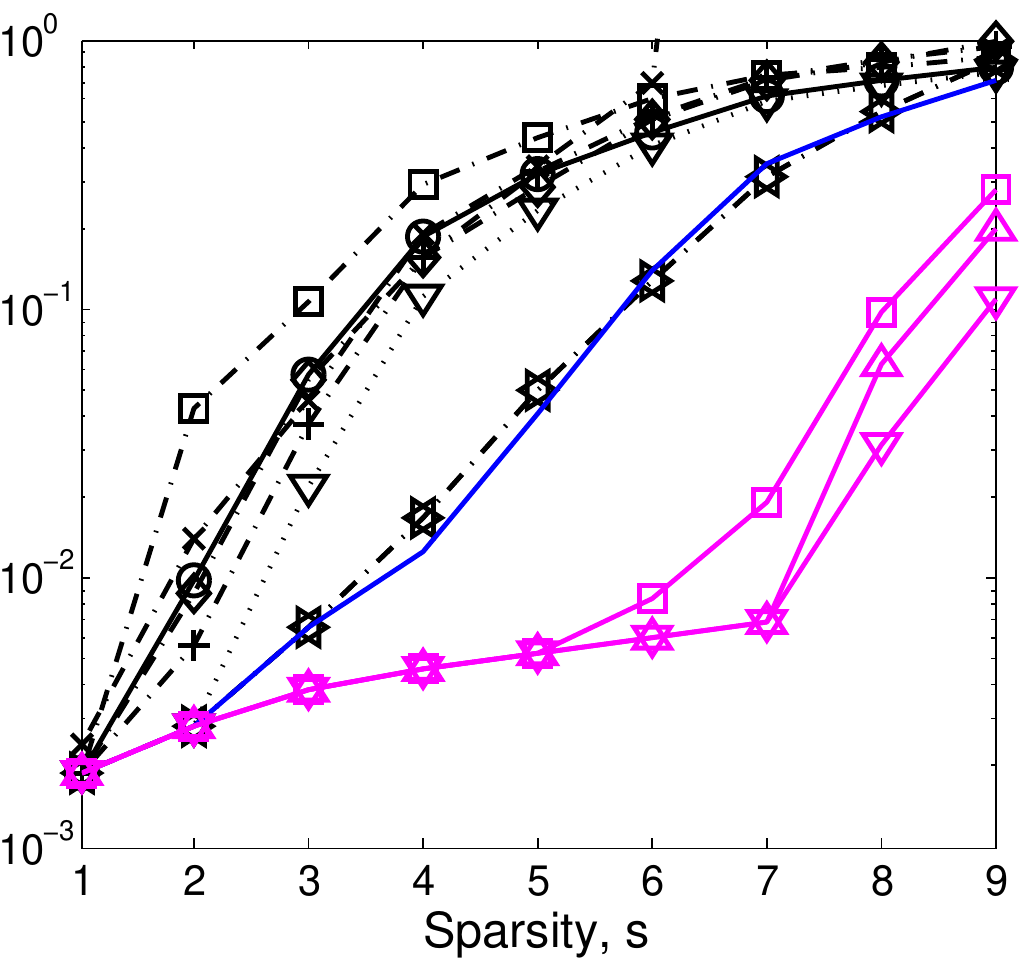}}
\subfigure[Execution time]{\includegraphics[width=3.7cm, height=3.9cm]{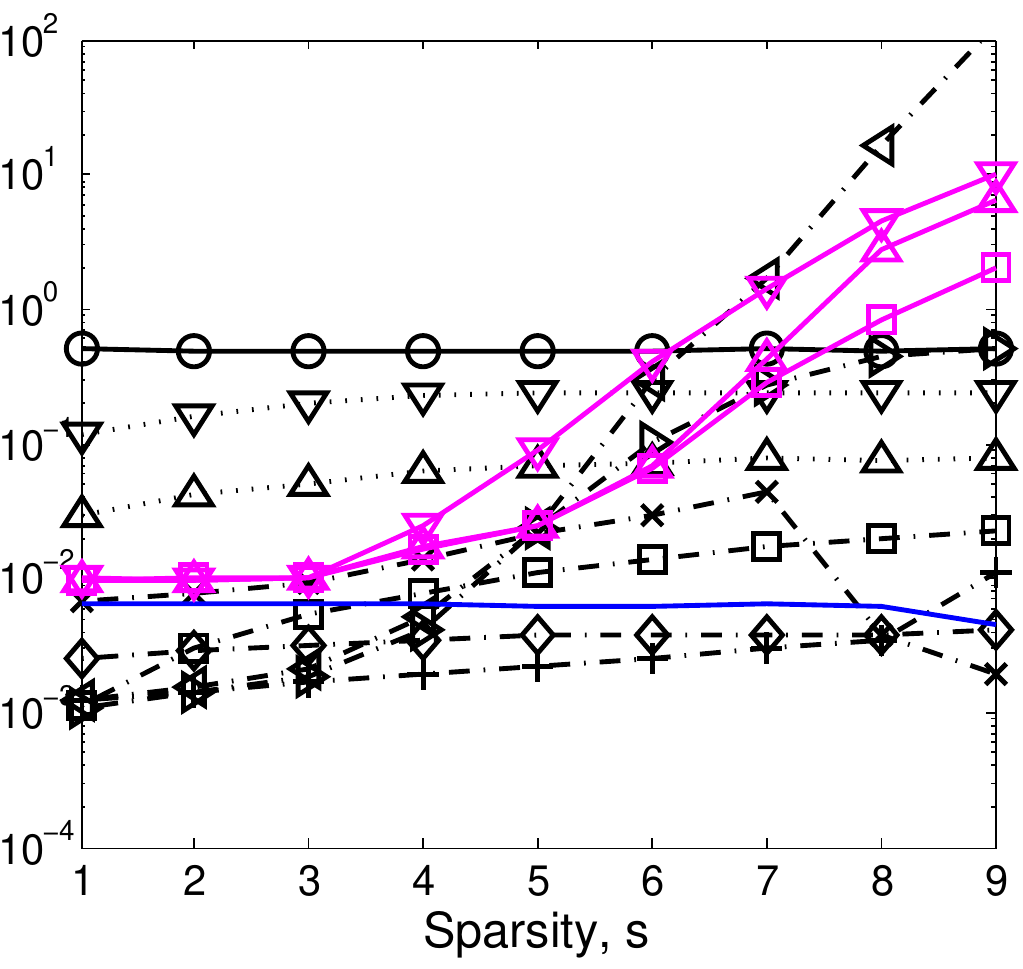}}
\subfigure[$\mathbb{P}(\frac{\left \| \Phi (\hat x - x_0) \right \|}{\left \| w \right \|} \leq 1)$]{\includegraphics[width=3.7cm, height=3.9cm]{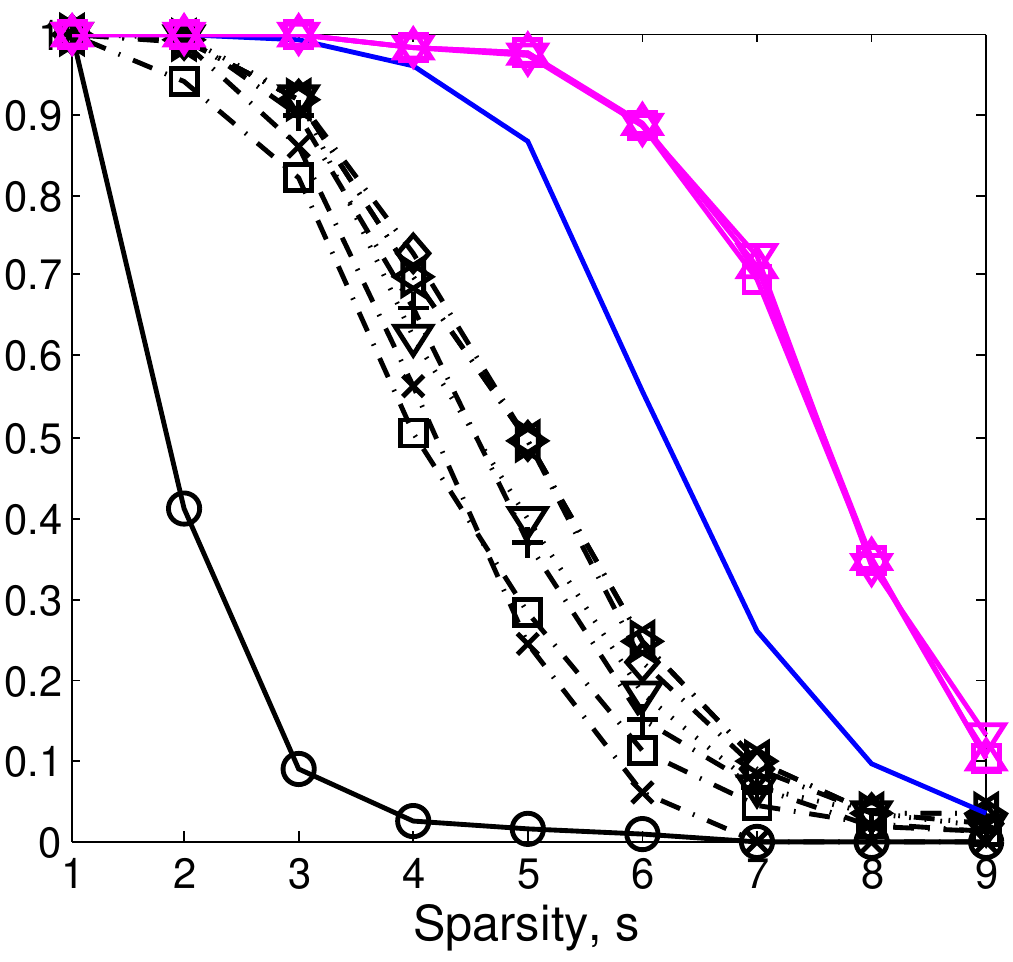}}
\subfigure[$\mathbb{E}(\frac{\left \| x_0 - \hat x\right \|}{\left \| x_0 \right \|} )$]{\includegraphics[width=3.7cm, height=3.9cm]{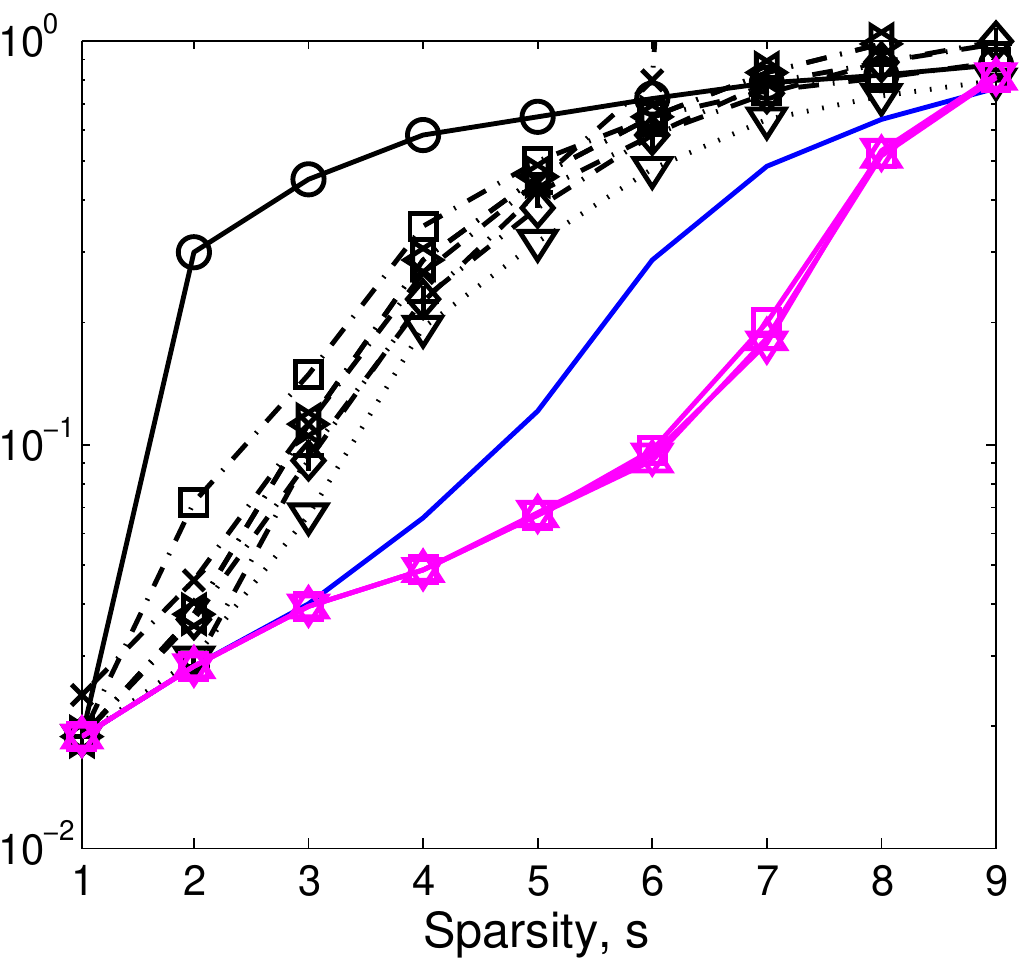}}
\subfigure[Execution time]{\includegraphics[width=3.7cm, height=3.9cm]{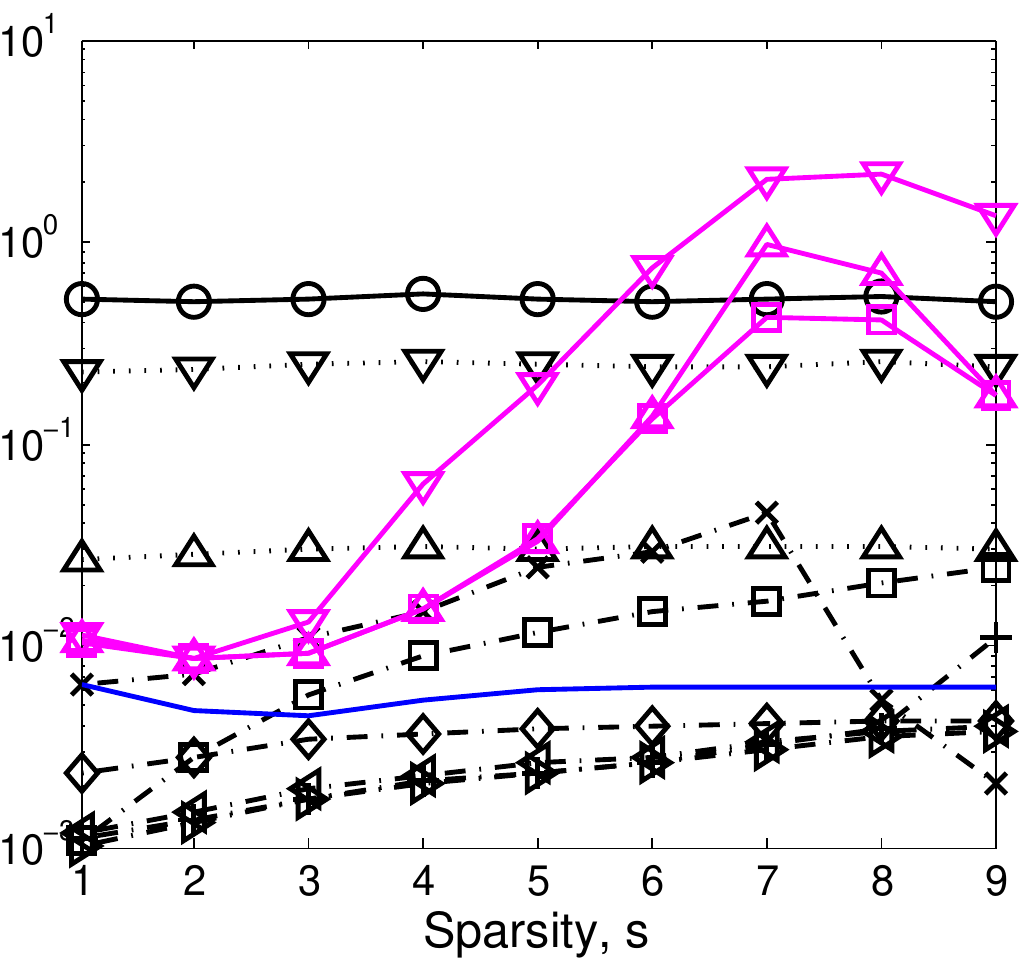}}
\caption{Performance comparison to conventional SR algorithms ((a)-(c): noiseless case,  (d)-(e): noisy case (SNR = $25$ dB),  (f)-(h): noisy case (SNR = $5$ dB)) }
\label{realfig}
\end{center}
\end{figure*}

We varied sparsity $s:=|\Omega|$ of $x_0$ from $1$ to $9$ and set input value $k$ in TSN to $9$; the true sparsity $s$ was not given but only its maximal value of 9 was given to TSN, whereas the sparsity $s$ was given to the other evaluation algorithms.   We evaluated the proposed TSN (Algorithm \ref{alg5} in Appendix \ref{tsns}) with three examples, namely, TSN$_i$ $:=$ TSN($y,\Phi,k, \boldsymbol{\tau}_i$), where the following set $\boldsymbol{\tau}_i$ of input parameters of TSN$_i$ is used for $i \in \{1:3\}$:
\begin{align}\nonumber
\scriptsize
&\boldsymbol{\tau}_1 :(q,z,\epsilon,\boldsymbol{l},\boldsymbol{g},t_{\textup{max}}) = (m,1,\bar \epsilon,(3,1),(60,1),\infty) \\\nonumber
&\boldsymbol{\tau}_2:(q,z,\epsilon,\boldsymbol{l},\boldsymbol{g},t_{\textup{max}}) = (m,1,\bar \epsilon,(2,1)^2,(60,1)^2,\infty)\\\nonumber
&\boldsymbol{\tau}_3:(q,z,\epsilon,\boldsymbol{l},\boldsymbol{g},t_{\textup{max}}) = (m,1,\bar \epsilon,(2,1)^2,(60,1)^2,5)
\end{align}
where $\bar \epsilon:= \max[(\left \| y\right \| \cdot 10^{-v_\textup{SNR$_{\textup{dB}}$}/20}),10^{-5}]$ is the signal error bound,  $t_{\textup{max}}$ is the maximum running time of TSN in second, $(\boldsymbol{l},\boldsymbol{g})$ determines how many nodes are left at  particular tree depths via the pruning method (Appendix \ref{tsns}), and $(a,b)^2:=(a,b,a,b)$.

The performance was evaluated according to two metrics, namely, the expected value of the normalized distance between $x_0$ and its estimate $\hat x$ obtained from the evaluated algorithm, $\mathbb{E}(\left \| x_0 - \hat x\right \| /\left \| x_0 \right \| )$, and the probability where the error between the target signal $x_0$ and its estimate $\hat x$ is smaller than the noise magnitude, i.e., $\mathbb{P}(\left \| \Phi \hat x - \Phi x_0 \right \| \leq \left \| w \right \|)$.\footnote{The second metric $\mathbb{P}(\left \| \Phi \hat x - \Phi x_0 \right \| \leq \left \| w \right \|)$ in the noiseless case is equal to the rate of successful signal recovery, $\mathbb{P}(x_0 = \hat x)$, in our simulation setting, as every set of $s$ columns of $\Phi$ has the rank $s$ for $ s \leq m/2$.}

\subsubsection{Performance evaluation} 
Figures \ref{ompreal}(a) and (b) show the performance of a modified TSN that replaces the DNN-based index selection with the OMP-based index selection in terms of signal recovery rate and execution time. The proposed TSN and its variant are denoted by TSN (GFLSTM) and TSN (OMP), respectively, in Figures \ref{ompreal}(a) and (b). The modified TSN using the OMP-based index selection outperforms MMP at a comparable complexity, and thus the proposed tree search, even omitting the DNN, is superior to the existing tree search. Still, the figures show that TSN with the DNN-based selection has a lower complexity and outperforms its modified one using the OMP-based selection, thus confirming the effectiveness of jointly exploiting tree search and DNN-based index selection.
We also compared TSN to other widely used DNN-based SR algorithms, including LISTA, LAMP, and LVAMP, obtaining the results shown in Figures \ref{ompreal}(c) and (d). LVAMP outperforms both LISTA and LAMP and retrieves a comparable performance to GFLSTM. TSN uniformly improves the performance of GFLSTM and LVAMP by setting the DNN used in TSN to GFLSTM and LVAMP, respectively, and TSN using GFLSTM has a lower complexity and outperforms TSN using LVAMP. In addition, TSN$_i$ shows better performance than TSN$_j$, whereas the complexity of TSN$_i$ is larger than that of TSN$_j$ for $i,j$ in $\{1:3\}$ such that $i<j$.\footnote{The total number of nodes in the search tree of TSN$_j$ is smaller than that of TSN$_i$.} 
Given that the performance does not notably improve when the number of layers in LISTA, LAMP, and LVAMP is larger than 12, we considered 12 layers.
\begin{figure*}[t]
\scriptsize
\begin{center}
\subfigure[Gaussian matrix]{\includegraphics[width=5.0cm, height=3.9cm]{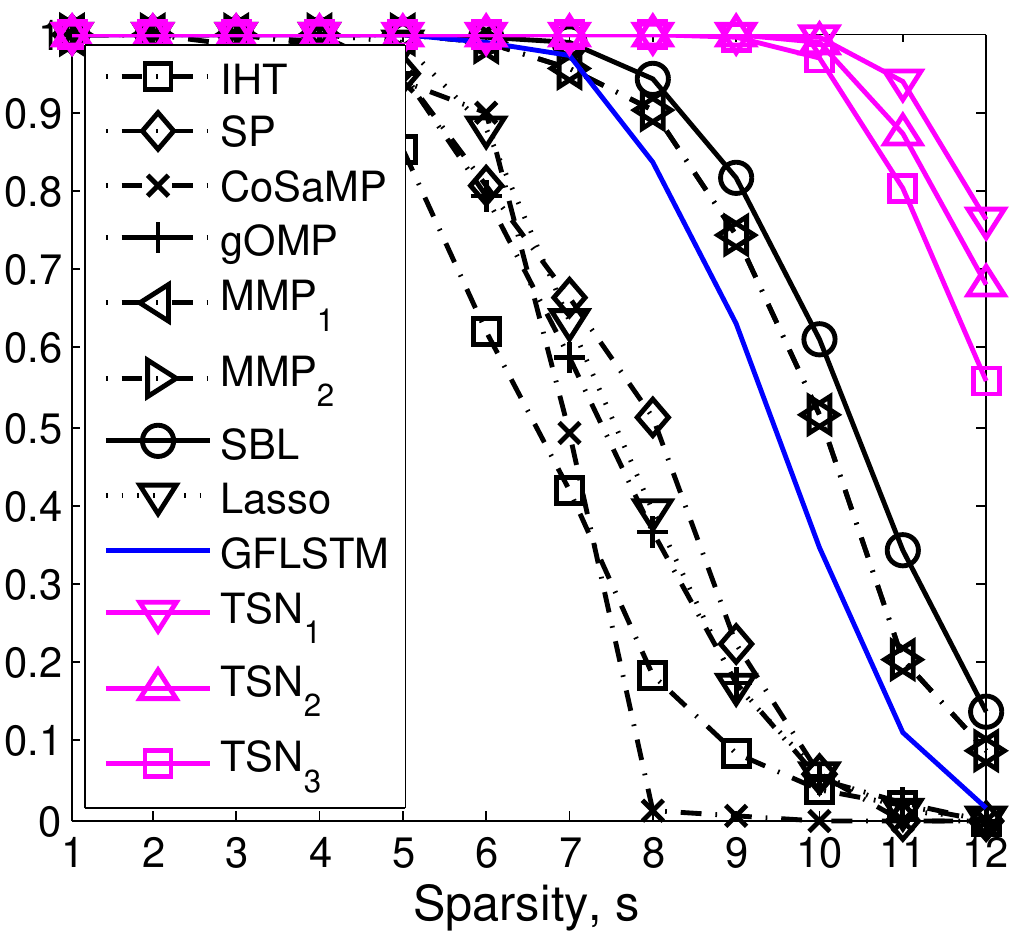}}
\subfigure[DFT matrix]{\includegraphics[width=5.0cm, height=3.9cm]{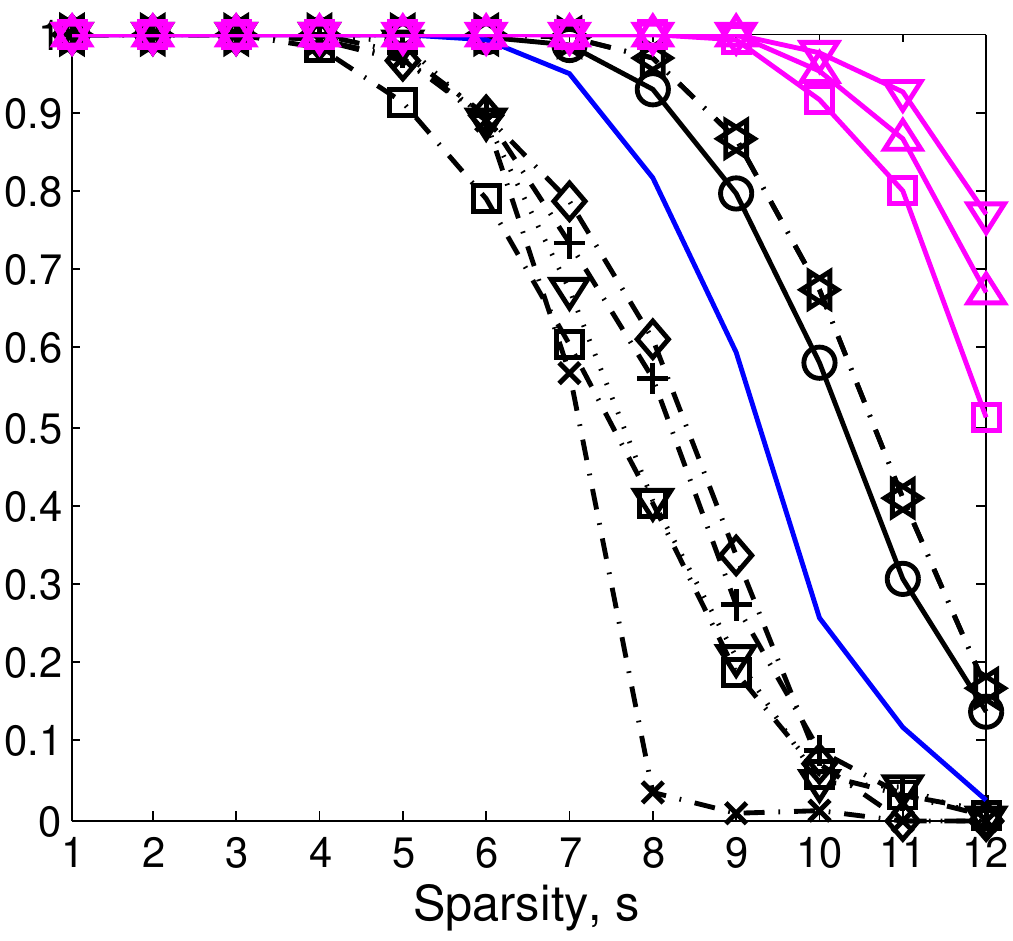}}
\subfigure[Matrix with correlated columns]{\includegraphics[width=5.0cm, height=3.9cm]{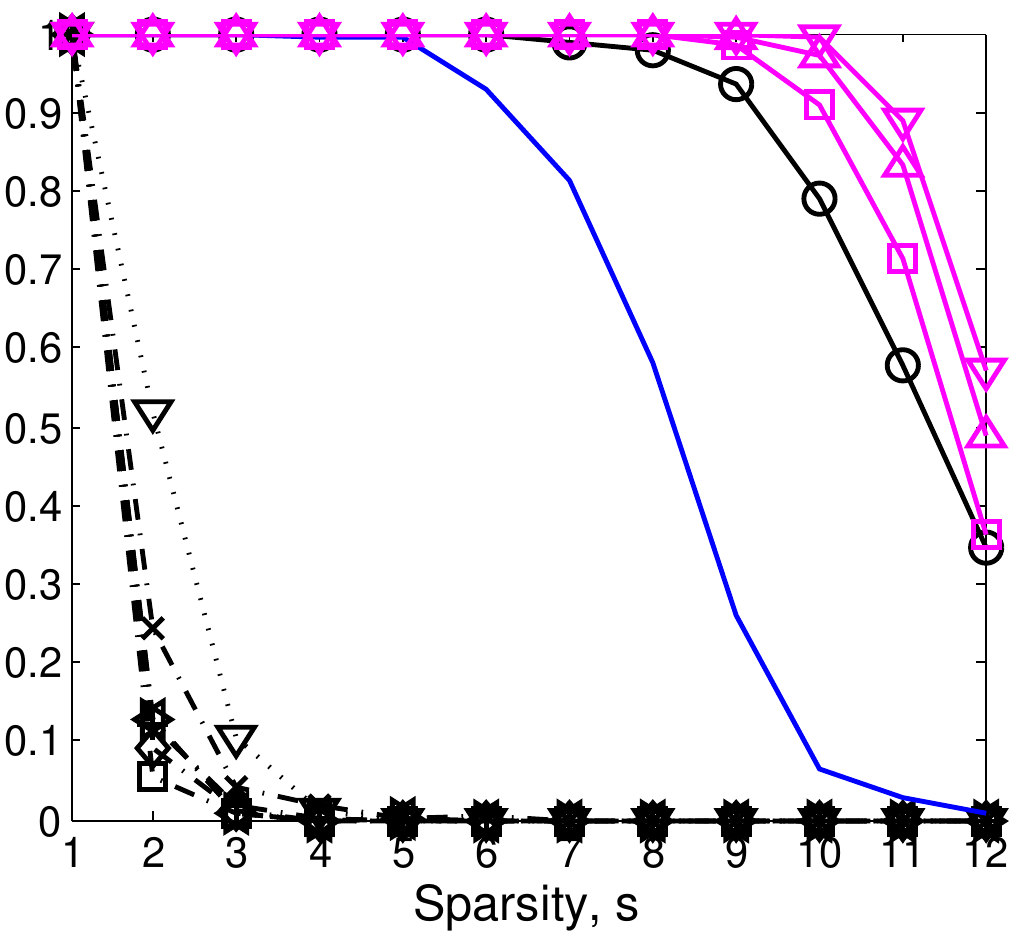}}
\subfigure[Gaussian matrix]{\includegraphics[width=5.0cm, height=3.9cm]{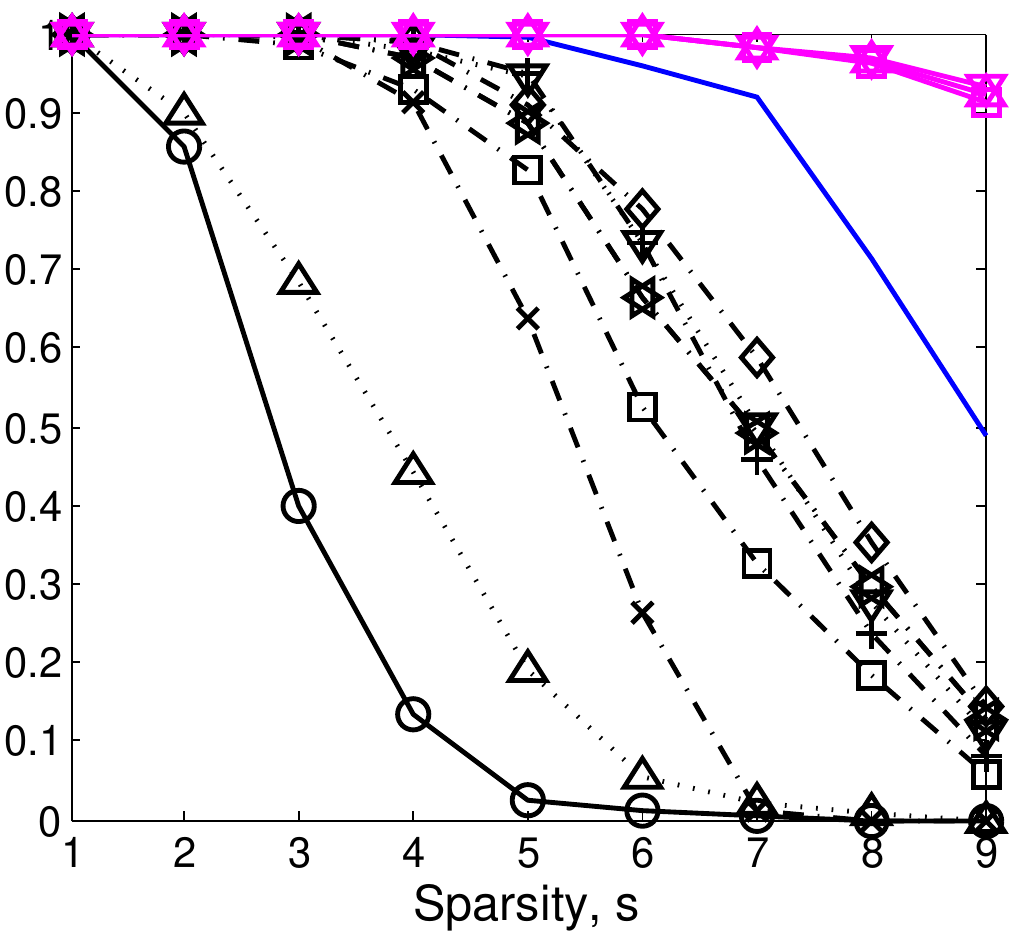}}
\subfigure[DFT matrix]{\includegraphics[width=5.0cm, height=3.9cm]{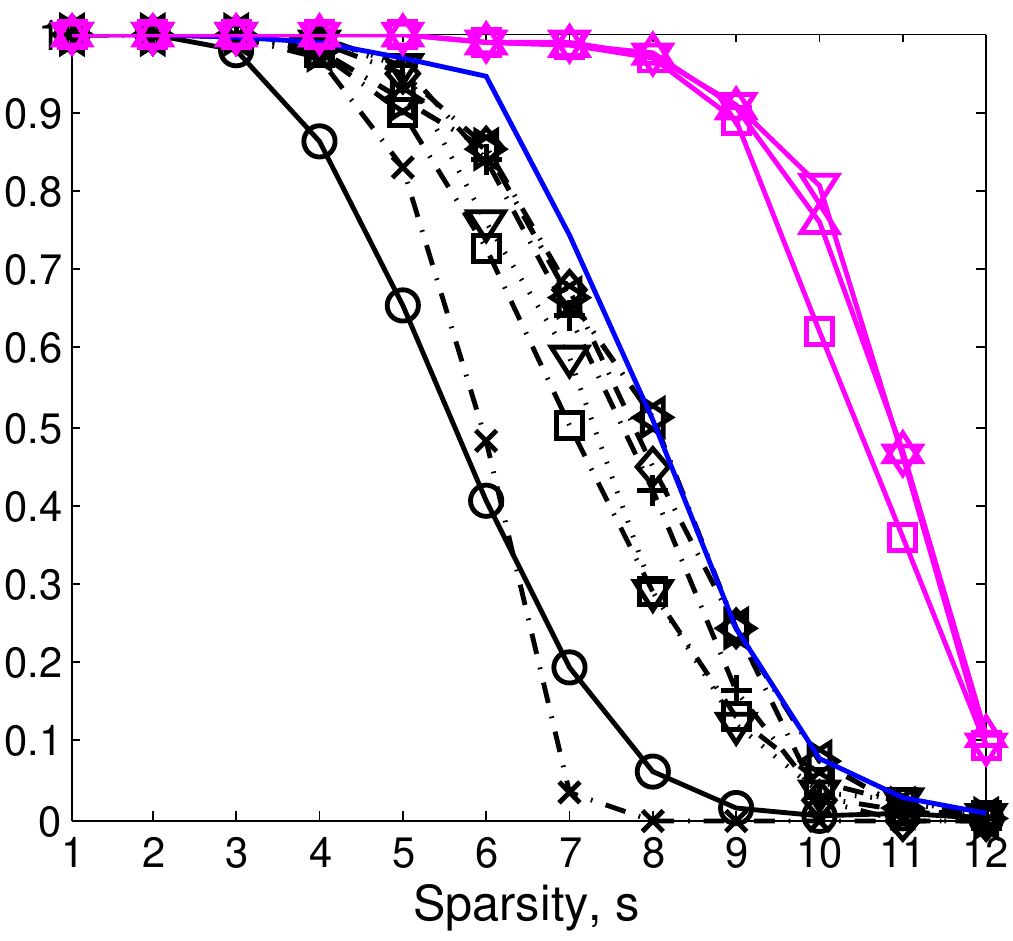}}
\subfigure[Matrix with correlated columns]{\includegraphics[width=5.0cm, height=3.9cm]{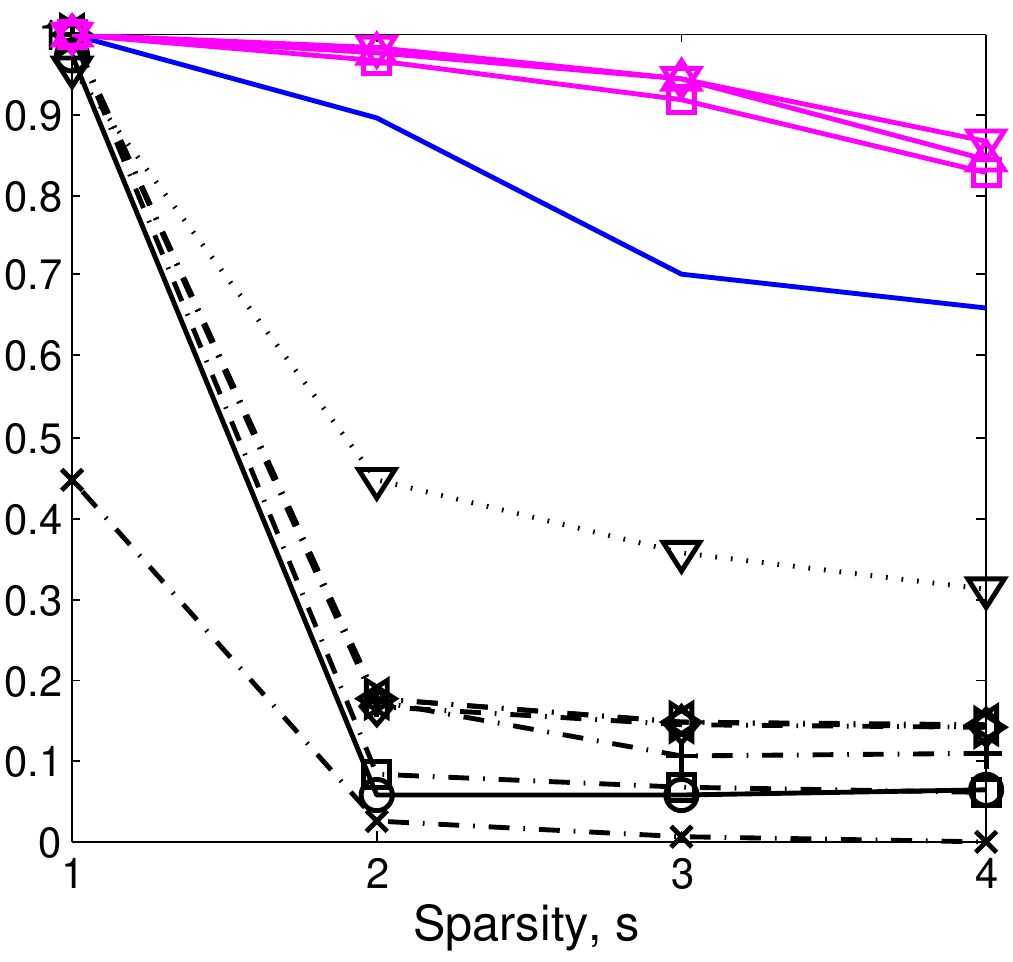}}
\caption{Signal recovery rate $\mathbb{P}(\left \| \Phi \hat x - \Phi x_0 \right \| \leq \left \| w \right \|)$ of each algorithm given complex-valued measurements ((a)-(c): noiseless case,  (d)-(f): noisy case (SNR = $5$ dB))}
\label{main_cpxfig}
\end{center}
\end{figure*}

Figures \ref{realfig}(a) and (b) show the performance comparison to conventional SR algorithms without using DNN in terms of signal recovery rate and  signal error, respectively, whereas Figure \ref{realfig}(c) shows the algorithm execution time\footnote{We simulated TSN and GFLSTM by using TensorFlow running on an E5-2640 v4 2.4GHz CPU endowed with a Nvidia Titan X GPU; the other algorithms were implemented in MATLAB.}, in the noiseless case. The results in Figures \ref{realfig}(a) and (b) show that TSN exhibits the best recovery performance of the signal $x_0$ and its support $\Omega$ in the whole sparsity region for the noiseless case.\footnote{We observe that MMP$_1$ does not improve the performance of MMP$_2$  though the tree of MMP$_1$ is extended from that of MMP$_2$.} In particular, it is observed that the maximum sparsity $s$ of $x_0$ for TSN to uniformly recover $x_0$ is 6 or 7, which is larger than two times those for other algorithms, by measuring the maximum sparsity with the signal error below $10^{-10}$ in Figure \ref{realfig}(b).  Figure \ref{realfig}(c) shows that TSN algorithms depicted in Figures \ref{realfig}(a) and (b) have execution times under 1 second for the true sparsity of up to $6$, which is around their phase-transition points shown in Figure \ref{realfig}(b).

The performance is compared in the noisy cases in Figures \ref{realfig}(d)--(h). These results show that TSN outperforms other algorithms to recover $x_0$ for most sparsity region even in the noisy case. 
Note that  from Figures \ref{realfig}(d) and (e), there exists a tradeoff between the recovery performance and complexity, as similarly shown in Figures \ref{realfig}(a)--(c), of TSN$_i$ for $i \in \{1:3\}$ in the case where SNR is $25$ dB.\footnote{As $\mathbb{P}(\left \| \Phi \hat x - \Phi x_0 \right \| \leq \left \| w \right \| )$ when SNR is $25$ dB is observed similarly with Figure \ref{realfig}(a), we omitted to show this plot.}  It is observed that the execution time of TSN$_3$ is under 2 seconds for sparsity $s \leq 9$ in Figure \ref{realfig}(e). Figures \ref{realfig}(f)--(h) show that in low SNR case ($5$ dB), TSN$_3$ takes the smallest execution time, under $1$ second in the whole sparsity region, with a similar recovery performance of $x_0$ among TSN$_i$ for $i \in \{1:3\}$.  The smaller SNR, the easier it is to find a $k$-sparse vector $z$ such that $\left \| \Phi z - \Phi x_0 \right \| \leq \left \| w \right \|$. For that reason, TSN$_3$, whose nodes in the tree are fewer than those of TSN$_1$ and TSN$_2$,  is sufficient for providing a signal estimate existing within the noise bound $\left \| w \right \|$ in low SNR case, as observed in Figures \ref{realfig}(f)--(h).

\subsection{Complex-valued case}
\label{cvm}
Even if $x_0$ and $\Phi$ are complex, we observed that TSN improved its target DNN-SR and outperformed other SR algorithms with the same tendency as in Figure \ref{realfig} in both noiseless and noisy cases. We set $\Phi$ to a complex Gaussian matrix, a partial DFT matrix, and a complex matrix with highly correlated columns \cite{he2017bayesian,xin2016maximal}, which have been typically used in the CS application. Other DNN-SR algorithms except for TSN and GFLSTM, shown in Figure \ref{ompreal}, can not be extended to the complex-valued case, they are omitted from the performance comparison result. Figure \ref{main_cpxfig} demonstrates our argument and shows that in the noiseless case TSN almost achieves the ideal limit \cite{cohen2009compressed} of sparsity $(s=10)$ for the uniform recovery of $x_0$ regardless of the type of $\Phi$.  Details including the setting parameters, the noisy case (SNR = $25$ dB), and the complexity are shown in Appendix \ref{apenb}.

\subsection{Scalability}
We evaluated the scaling of performance and complexity in the TSN at different sizes of sensing matrix $\Phi$.
Table \ref{table0} lists the maximum sparsity and execution time of each algorithm at which the recovery rate is at least 95\%. TSN notably outperforms the other methods even when $(m,n)$ increases beyond $(20, 100)$, and its performance and complexity scale well with $n$.\footnote{The execution time for TSN sometimes decreases with $n$ following the trend of $s_{0.95}$.} The setting parameters are detailed in Appendix \ref{scal_apx}.

We observed that as $m$ approaches $n$ and $m$ increases, the complexity versus performance of TSN relatively increases. It implies that there exists a trade-off between performance and computational complexity of TSN in the case when $ m $ is large enough. To show this, results with matrix size $m\times n$ reaching $1500 \times 10^4$ are detailed in Appendix \ref{scal_apx}. Thus, at least for SR problems with a small value of $m$, regardless of the value of $n$, we can conclude that TSN has a higher performance versus complexity than other SR algorithms. Note that the smaller the sampling rate $m/n$, the more difficult it is to recover the signal in the SR problem. Our study suggests that TSN enables to derive the state-of-the-art result to solve the SR problem for this difficult case.

\begin{table} 
\centering
\caption{The maximum sparsity $s_{0.95}$ of algorithm at which the exact support recovery rate is at least 95 percent (the execution time of algorithm when the sparsity $|\Omega|$ is $s_{0.95}$)}
\tiny

\begin{tabular}{|m{0.1cm}|m{0.1cm}|m{0.1cm}|m{0.1cm}|m{0.1cm}|m{0.1cm}|m{0.1cm}|m{0.1cm}|m{0.1cm}|m{0.1cm}|m{0.1cm}|m{0.1cm}|m{0.1cm}|}
  \hline
  \multicolumn{1}{|c|}{Alg $\setminus$ $(m/n)$} & \multicolumn{1}{c|}{$20/100$} 
& \multicolumn{1}{c|}{$20/400$}  & \multicolumn{1}{c|}{$20/800$} & \multicolumn{1}{c|}{$20/3200$} & \multicolumn{1}{c|}{$20/10^4$}  & \multicolumn{1}{c|}{$40/100$}
& \multicolumn{1}{c|}{$40/400$} & \multicolumn{1}{c|}{$40/800$} & \multicolumn{1}{c|}{$40/3200$} & \multicolumn{1}{c|}{$40/10^4$} \\\hline 
   \multicolumn{1}{|c|}{TSN} & \multicolumn{1}{l|}{$\boldsymbol{7}\,(0.7)$} 
& \multicolumn{1}{l|}{$\boldsymbol{5}\,(2.2)$}& \multicolumn{1}{l|}{$\boldsymbol{4}\,(2.1)$} & \multicolumn{1}{l|}{$\boldsymbol{3}\,(2.08)$} & \multicolumn{1}{l|}{$\boldsymbol{2}\,(0.8)$}  & \multicolumn{1}{l|}{$\boldsymbol{17}\,(3.2)$} 
& \multicolumn{1}{l|}{$\boldsymbol{10}\,(2.9)$} & \multicolumn{1}{l|}{$\boldsymbol{8}\,(3.4)$} & \multicolumn{1}{l|}{$\boldsymbol{6}\,(1.9)$} & \multicolumn{1}{l|}{$\boldsymbol{5}\,(3.1)$} \\\cline{1-11}   
   \multicolumn{1}{|c|}{SBL} & \multicolumn{1}{l|}{$4\,(0.5)$}
& \multicolumn{1}{l|}{$3\,(1.1)$}& \multicolumn{1}{l|}{$2\,(1.8)$} & \multicolumn{1}{l|}{${1}\,(6.8)$}   & \multicolumn{1}{l|}{${1}\,(25.7)$} & \multicolumn{1}{l|}{$13\,(1.5)$} 
& \multicolumn{1}{l|}{$7\,(2.1)$} & \multicolumn{1}{l|}{$6\,(3.6)$} & \multicolumn{1}{l|}{${3}\,(13.1)$} & \multicolumn{1}{l|}{${2}\,(52.3)$}   \\\cline{1-11}  
   \multicolumn{1}{|c|}{MMP} & \multicolumn{1}{l|}{$4\,(0.004)$} 
& \multicolumn{1}{l|}{$3\,(0.003)$}& \multicolumn{1}{l|}{$2\,(0.002)$} & \multicolumn{1}{l|}{${2}\,(0.003)$} & \multicolumn{1}{l|}{${1}\,(0.003)$} & \multicolumn{1}{l|}{$11\,(0.06)$} 
& \multicolumn{1}{l|}{$7\,(0.03)$} & \multicolumn{1}{l|}{$6\,(0.03)$} & \multicolumn{1}{l|}{${4}\,(0.05)$} & \multicolumn{1}{l|}{${3}\,(0.04)$}  \\\cline{1-11}  
  \multicolumn{1}{|c|}{Lasso} & \multicolumn{1}{l|}{$2\,(0.1)$} 
& \multicolumn{1}{l|}{$2\,(0.1)$}& \multicolumn{1}{l|}{$2\,(0.2)$} & \multicolumn{1}{l|}{${1}\,(0.1)$} & \multicolumn{1}{l|}{${1}\,(0.3)$} & \multicolumn{1}{l|}{$9\,(0.2)$} 
& \multicolumn{1}{l|}{$5\,(0.2)$} & \multicolumn{1}{l|}{$4\,(0.2)$} & \multicolumn{1}{l|}{${3}\,(0.3)$} & \multicolumn{1}{l|}{${3}\,(0.5)$} \\\cline{1-11}  
  \multicolumn{1}{|c|}{gOMP} & \multicolumn{1}{l|}{$2\,(0.003)$} 
& \multicolumn{1}{l|}{$2\,(0.005)$}& \multicolumn{1}{l|}{$2\,(0.008)$} & \multicolumn{1}{l|}{${1}\,(0.02)$} & \multicolumn{1}{l|}{${1}\,(0.08)$}  & \multicolumn{1}{l|}{$9\,(0.004)$} 
& \multicolumn{1}{l|}{$5\,(0.008)$} & \multicolumn{1}{l|}{$4\,(0.007)$} & \multicolumn{1}{l|}{${3}\,(0.04)$} & \multicolumn{1}{l|}{${3}\,(0.16)$}  \\\cline{1-11}  
   \multicolumn{1}{|c|}{IHT} & \multicolumn{1}{l|}{$1\,(0.001)$} 
& \multicolumn{1}{l|}{$1\,(0.001)$}& \multicolumn{1}{l|}{$1\,(0.001)$} & \multicolumn{1}{l|}{${1}\,(0.004)$} & \multicolumn{1}{l|}{${1}\,(0.008)$} & \multicolumn{1}{l|}{$6\,(0.006)$} 
& \multicolumn{1}{l|}{$4\,(0.007)$} & \multicolumn{1}{l|}{$3\,(0.007)$} & \multicolumn{1}{l|}{${2}\,(0.01)$} & \multicolumn{1}{l|}{${2}\,(0.03)$}  \\\cline{1-11}  
   \multicolumn{1}{|c|}{GFLSTM} & \multicolumn{1}{l|}{$4\,(0.02)$}
& \multicolumn{1}{l|}{$2\,(0.02)$}& \multicolumn{1}{l|}{$2\,(0.02)$} & \multicolumn{1}{l|}{${1}\,(0.02)$} & \multicolumn{1}{l|}{${1}\,(0.02)$}  & \multicolumn{1}{l|}{$11\,(0.02)$} 
& \multicolumn{1}{l|}{$6\,(0.02)$} & \multicolumn{1}{l|}{$4\,(0.02)$} & \multicolumn{1}{l|}{${2}\,(0.02)$} & \multicolumn{1}{l|}{${1}\,(0.02)$}  \\\cline{1-11}  
   \multicolumn{1}{|c|}{LVAMP} & \multicolumn{1}{l|}{$3\,(0.02)$} 
& \multicolumn{1}{l|}{$2\,(0.02)$}& \multicolumn{1}{l|}{$1\,(0.02)$} & \multicolumn{1}{l|}{${1}\,(0.02)$} & \multicolumn{1}{l|}{${1}\,(0.02)$}  & \multicolumn{1}{l|}{$11\,(0.02)$} 
& \multicolumn{1}{l|}{$2\,(0.02)$} & \multicolumn{1}{l|}{$1\,(0.02)$} & \multicolumn{1}{l|}{${1}\,(0.02)$} & \multicolumn{1}{l|}{${1}\,(0.02)$}   \\\cline{1-11}  
\end{tabular}
\label{table0}
\end{table}
\label{struc}
\begin{figure} 
\scriptsize
\begin{center}
\subfigure[]{\includegraphics[width=3.7cm, height=3.9cm]{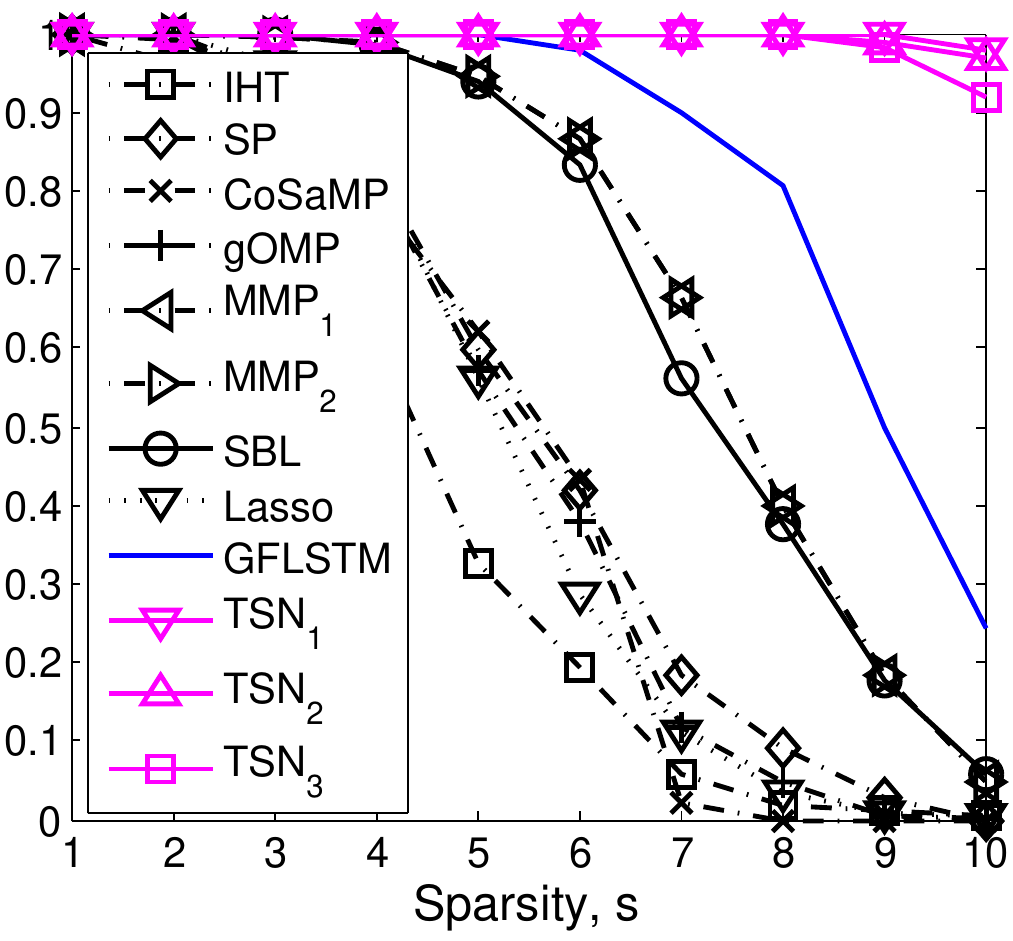}}
\subfigure[]{\includegraphics[width=3.7cm, height=3.9cm]{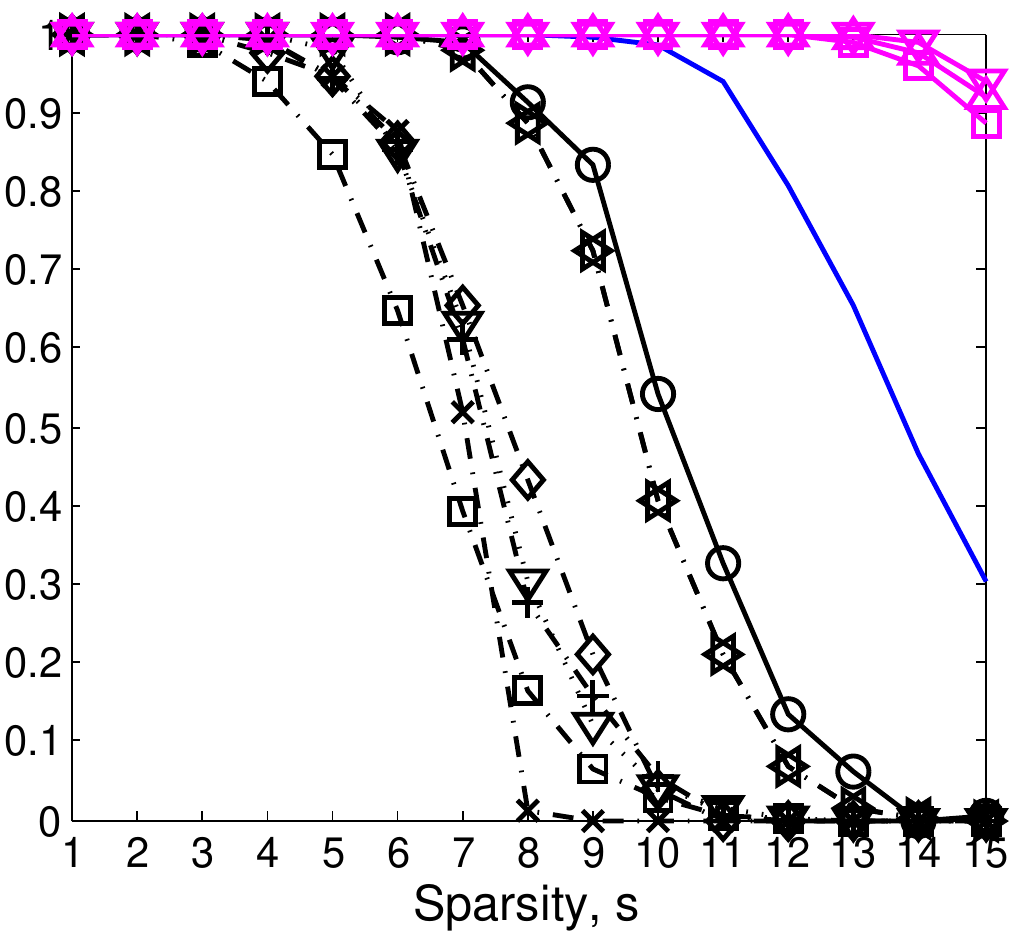}}
\subfigure[]{\includegraphics[width=3.7cm, height=3.9cm]{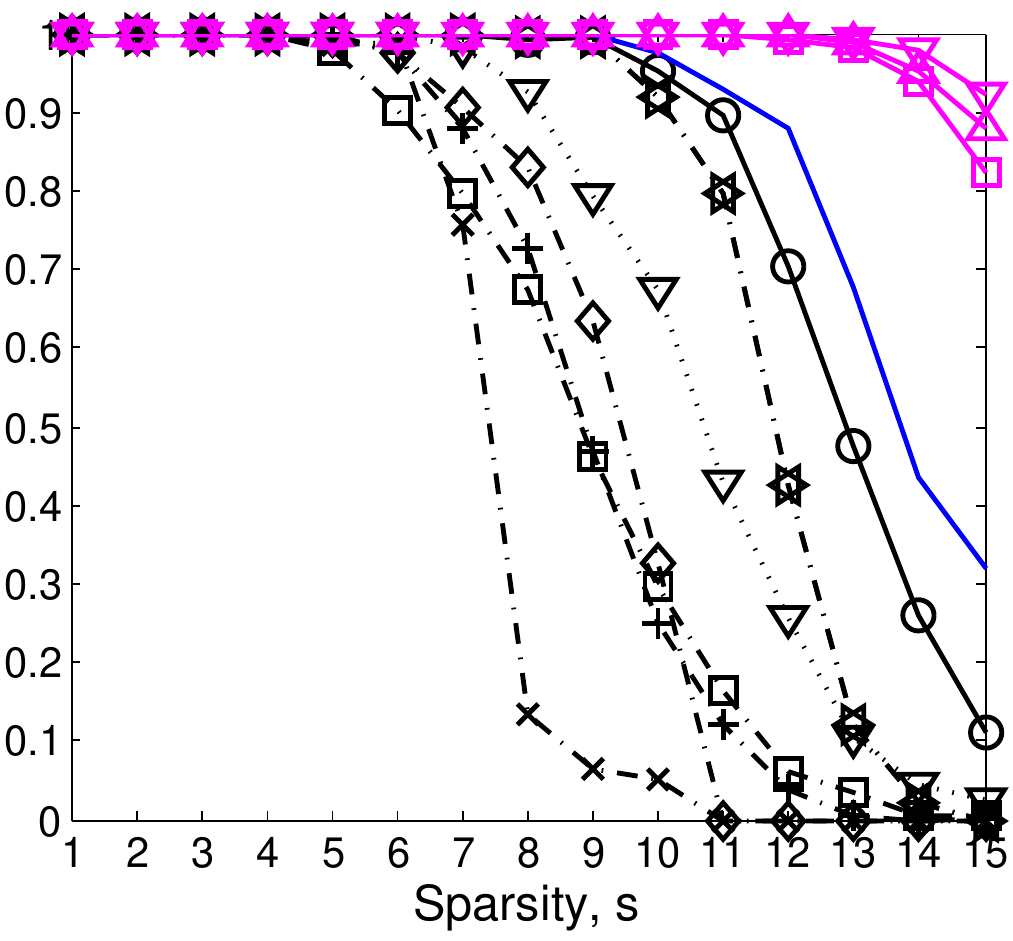}}
\subfigure[]{\includegraphics[width=3.7cm, height=3.9cm]{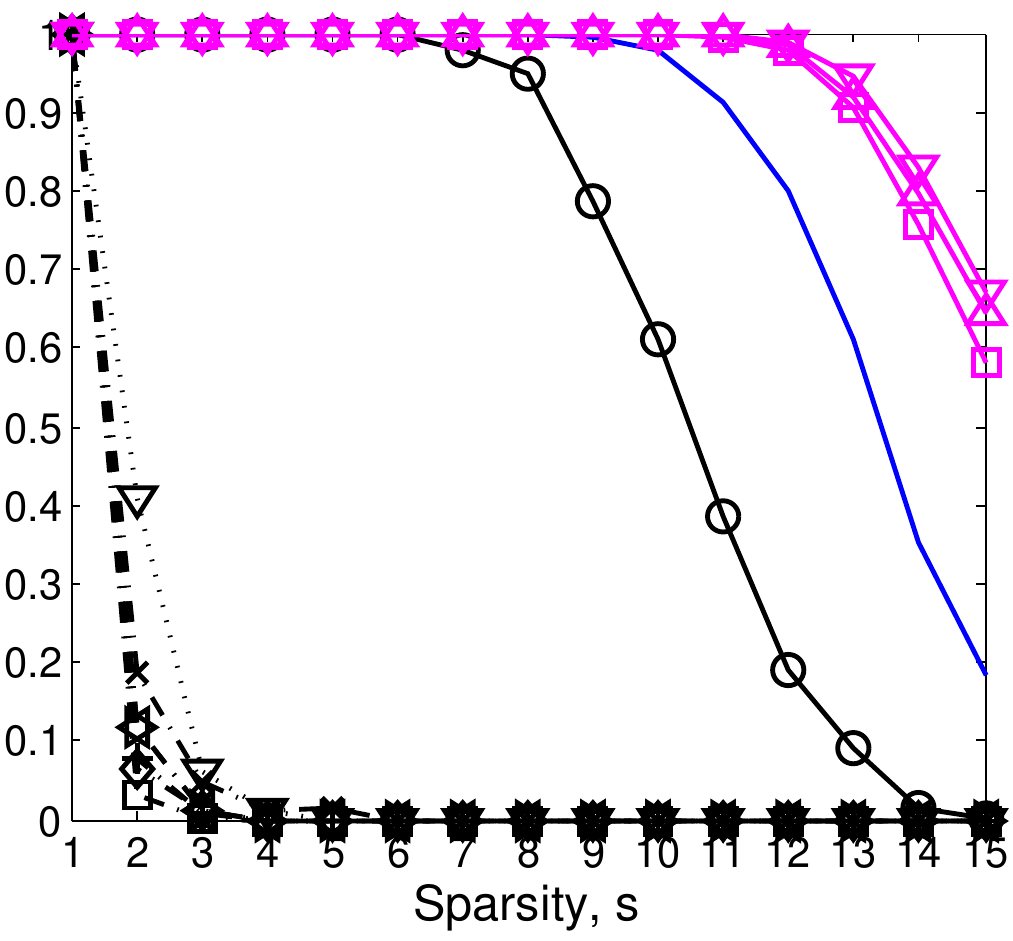}}
\caption{Signal recovery rate $\mathbb{P}( \hat x  = x_0 )$ in the noiseless case ((a): the real Gaussian matrix,  (b): the complex Gaussian matrix,  (c): the partial DFT matrix,  (d): the matrix with correlated columns)}
\label{main_posfig}
\end{center}
\end{figure} 

\subsection{Estimation for structured sparse signal}
TSN can further improve its recovery performance when the signal distribution $p_{x_0}(\cdot)$ has an additional structure as TSN is based on DNN-SR trained by data sampled from that distribution. 
To provide an example (Figure \ref{main_posfig}), we added a condition to the signal distribution used in Sections \ref{rvm} and \ref{cvm}, such that all nonzero elements of $x_0$ are positive. Figure \ref{main_posfig}(a) shows the performance improvement of TSN compared to Figure \ref{realfig}(a). It is observed in Figure \ref{main_posfig}(a) that TSN almost achieves the ideal limit of sparsity $(s=10)$ for the uniform recovery of $x_0$, while the maximum sparsity for perfect signal recovery through SBL is 3; TSN performs nearly three times better than SBL. Similarly, with the real case, the performance improvement of TSN is observed in the complex case when the real and imaginary parts of nonzero values in $x_0$ are positive. Figures \ref{main_posfig}(b), \ref{main_posfig}(c), and \ref{main_posfig}(d) show the performance improvement of TSN compared to those in Figures \ref{main_cpxfig}(a),  \ref{main_cpxfig}(b), and \ref{main_cpxfig}(c), respectively. More details including the setting parameters and the complexity are shown in Appendix \ref{apenb}. 

\section{Application}
\subsection{Nonorthogonal multiple access}
We evaluated TSN using NOMA \cite{wang2016dynamic,wunder2014compressive} on an AWGN channel. We assumed that number $n$ of total users in the cell is $100$, number $s:=|\Omega|$ of active users is $10$, and number $m$ of measurements is $20$. The spreading sequence of each user is set to each column of a $20 \times 100$ DFT matrix. We used the Bose-–Chaudhuri-–Hocquenghem code with codeword length $7$ and message length $4$ and applied quadrature phase shift keying modulation. Table \ref{table_noma} shows that TSN identifies the $10$ active users with probability above 90\% and block error rate below $1$\%, whereas the other SR algorithms have identification rate below 50\% and block error rate above $10$\%, regardless of SNR, i.e., E$_\textup{b}$/N$_0$. Thus, TSN reduces block error rate to less than one-tenth. Note that TSN has a complexity comparable  with SBL and MMP. This shows the superiority of TSN in a real application. More details are presented in Appendix \ref{app_noma}. 
\label{Application}
\begin{table}[t]
\centering
\footnotesize
\caption{ Tuple $(u,b,t)$ of each algorithm under NOMA with $10$ active users. The tuple contains perfect recovery rate $u$ ($0\leq u \leq1$) for all active users, block error rate $b$ ($0\leq b \leq1$), and execution time $t$ (s)}
\begin{tabular}{|m{0.2cm}|m{0.2cm}|m{0.2cm}|}
  \hline
  \multicolumn{1}{|c|}{ Alg. $\setminus$ E$_\textup{b}$/N$_0$} & \multicolumn{1}{c|}{0  (dB)} & \multicolumn{1}{c|}{20  (dB)} \\\hline
   \multicolumn{1}{|c|}{TSN} & \multicolumn{1}{c|}{$\boldsymbol{(0.93,0.01,3.1)}$} & \multicolumn{1}{c|}{$\boldsymbol{(0.98,0.002,3.7)}$}    \\\cline{1-3}  
   \multicolumn{1}{|c|}{GFLSTM} & \multicolumn{1}{c|}{$(0.36,0.1,0.03)$} & \multicolumn{1}{c|}{$(0.45,0.1,0.03)$}  \\\cline{1-3} 
   \multicolumn{1}{|c|}{SBL} & \multicolumn{1}{c|}{$(0.07,0.22,3.2)$} & \multicolumn{1}{c|}{$(0.27,0.21,3.1)$}  \\\cline{1-3} 
   \multicolumn{1}{|c|}{MMP$_{1}$} & \multicolumn{1}{c|}{$(0.18,0.28,3.8)$} & \multicolumn{1}{c|}{$(0.21,0.27,3.8)$}  \\\cline{1-3} 
   \multicolumn{1}{|c|}{MMP$_{2}$} & \multicolumn{1}{c|}{$(0.18,0.28,37.6)$} & \multicolumn{1}{c|}{$(0.21,0.27,39.2)$}  \\\cline{1-3} 
   \multicolumn{1}{|c|}{Lasso} & \multicolumn{1}{c|}{$(0.15,0.2,0.2)$} & \multicolumn{1}{c|}{$(0.16,0.2,0.2)$}  \\\cline{1-3} 
   \multicolumn{1}{|c|}{IHT} & \multicolumn{1}{c|}{$(0.06, 0.24, 0.11)$} & \multicolumn{1}{c|}{$(0.1, 0.23, 0.11)$}  \\\cline{1-3} 
   \multicolumn{1}{|c|}{gOMP} & \multicolumn{1}{c|}{$(0.01,0.4,0.06)$} & \multicolumn{1}{c|}{$(0.01,0.4,0.05)$}  \\\cline{1-3} 
\end{tabular}
\label{table_noma}
\end{table}

\subsection{Image reconstruction: MNIST and OMNIGLOT}
Table \ref{table_mnist} and Table \ref{table_omni} show the performance for reconstructing MNIST and OMNIGLOT \cite{lake2015human} images of $28 \times 28$ pixels in the noisy case with SNR = $25$ dB, respectively.\footnote{We omitted to show the signal error of CoSaMP, because it has larger than 3.} We randomly sampled $20$ MNIST images with sparsity equal to $215$ and OMNIGLOT images with sparsity equal to $160$. The sampled images are compressed by using real Gaussian sensing matrix $\Phi \in \mathbb{R}^{420 \times 784}$. TSN recovers the MNIST and OMNIGLOT image with signal error $0.13$ and $0.17$, whereas the others have signal errors above $0.2$ and $0.4$, respectively. Thus, TSN reduces the error rate of reconstructing MNIST and OMNIGLOT images to less than half. The running time of TSN is less than one-tenth of those of SBL and MMP. Note that the DNN for TSN was trained without exploiting MNIST or OMNIGLOT image data. For the purpose of fair comparison, we excluded  SR algorithms, e.g., learned denoising-based AMP (LDAMP)  \cite{metzler2017learned}, which are required for  real image data in the learning process. More details including the setting parameters and some reconstructed images are presented in Appendix \ref{app_ir}.

\begin{table}[t]
\footnotesize
\caption{Performance for MNIST image recovery in terms of signal error and signal recovery rate when SNR is $25$ dB and sparsity of the target MNIST image is $215$}
\begin{center}
\begin{tabular}{|p{5cm}|p{5cm}|p{2cm}|p{2cm}|}
  \hline
  \multicolumn{1}{|c|}{Algorithm} & \multicolumn{1}{c|}{${\left \| x_0 - \hat x\right \|}/{\left \| x_0 \right \|}$} & \multicolumn{1}{c|}{$|\hat \Omega \cap \Omega|/|\Omega|$} & \multicolumn{1}{c|}{Running time} \\\hline
   \multicolumn{1}{|c|}{TSN} & \multicolumn{1}{c|}{$\boldsymbol{0.1301}$} & \multicolumn{1}{c|}{$\boldsymbol{0.8577}$}      & \multicolumn{1}{c|}{11.47}    \\\cline{1-4}    
   \multicolumn{1}{|c|}{GFLSTM} & \multicolumn{1}{c|}{ 0.2605} & \multicolumn{1}{c|}{0.7805} & \multicolumn{1}{c|}{0.091} \\\cline{1-4}  
   \multicolumn{1}{|c|}{LVAMP} & \multicolumn{1}{c|}{ 0.7419} & \multicolumn{1}{c|}{0.5493} & \multicolumn{1}{c|}{1.457}  \\\cline{1-4}
   \multicolumn{1}{|c|}{SBL} & \multicolumn{1}{c|}{0.7488} & \multicolumn{1}{c|}{0.5513}  & \multicolumn{1}{c|}{ 116.3
} \\\cline{1-4}  
   \multicolumn{1}{|c|}{MMP} & \multicolumn{1}{c|}{1.0178} & \multicolumn{1}{c|}{0.4884} & \multicolumn{1}{c|}{ 391.9} \\\cline{1-4}  
   \multicolumn{1}{|c|}{Lasso} & \multicolumn{1}{c|}{ 0.4684} & \multicolumn{1}{c|}{0.7342}  & \multicolumn{1}{c|}{2.622} \\\cline{1-4}  
   \multicolumn{1}{|c|}{IHT} & \multicolumn{1}{c|}{ 0.6452} & \multicolumn{1}{c|}{0.6422} & \multicolumn{1}{c|}{ 0.267
}  \\\cline{1-4}  
   \multicolumn{1}{|c|}{gOMP} & \multicolumn{1}{c|}{1.0627} & \multicolumn{1}{c|}{0.4403}  & \multicolumn{1}{c|}{6.948
} \\\cline{1-4}   
   \multicolumn{1}{|c|}{SP} & \multicolumn{1}{c|}{ 0.8126} & \multicolumn{1}{c|}{0.5028} & \multicolumn{1}{c|}{ 0.233
}  \\\cline{1-4}  
   \multicolumn{1}{|c|}{CoSaMP} & \multicolumn{1}{c|}{-} & \multicolumn{1}{c|}{0.4561} & \multicolumn{1}{c|}{ 0.064
}  \\\cline{1-4}  
\end{tabular}
\label{table_mnist}
\end{center}
\end{table}

\begin{table}
\footnotesize
\caption{Performance for OMNIGLOT image recovery in terms of signal error and signal recovery rate when SNR is $25$ dB and sparsity of the target OMNIGLOT image is $160$}
\begin{center}
\begin{tabular}{|p{5cm}|p{5cm}|p{2cm}|p{2cm}|}
  \hline
  \multicolumn{1}{|c|}{Algorithm} & \multicolumn{1}{c|}{${\left \| x_0 - \hat x\right \|}/{\left \| x_0 \right \|}$} & \multicolumn{1}{c|}{$|\hat \Omega \cap \Omega|/|\Omega|$} & \multicolumn{1}{c|}{Running time} \\\hline
   \multicolumn{1}{|c|}{TSN} & \multicolumn{1}{c|}{$\boldsymbol{0.1741}$} & \multicolumn{1}{c|}{$\boldsymbol{0.9256}$}      & \multicolumn{1}{c|}{2.106}    \\\cline{1-4}    
   \multicolumn{1}{|c|}{GFLSTM} & \multicolumn{1}{c|}{0.7008} & \multicolumn{1}{c|}{0.6289} & \multicolumn{1}{c|}{0.054} \\\cline{1-4}  
   \multicolumn{1}{|c|}{LVAMP} & \multicolumn{1}{c|}{0.7654} & \multicolumn{1}{c|}{0.6039} & \multicolumn{1}{c|}{0.855}  \\\cline{1-4}
   \multicolumn{1}{|c|}{SBL} & \multicolumn{1}{c|}{0.7665} & \multicolumn{1}{c|}{0.6359}  & \multicolumn{1}{c|}{197.1} \\\cline{1-4}  
   \multicolumn{1}{|c|}{MMP} & \multicolumn{1}{c|}{0.9506} & \multicolumn{1}{c|}{0.5811} & \multicolumn{1}{c|}{265.1} \\\cline{1-4}  
   \multicolumn{1}{|c|}{Lasso} & \multicolumn{1}{c|}{0.4082} & \multicolumn{1}{c|}{0.8557}  & \multicolumn{1}{c|}{4.205} \\\cline{1-4}  
   \multicolumn{1}{|c|}{IHT} & \multicolumn{1}{c|}{0.6961} & \multicolumn{1}{c|}{0.6991} & \multicolumn{1}{c|}{0.969}  \\\cline{1-4}  
   \multicolumn{1}{|c|}{gOMP} & \multicolumn{1}{c|}{1.0096} & \multicolumn{1}{c|}{0.4475}  & \multicolumn{1}{c|}{11.49} \\\cline{1-4}   
   \multicolumn{1}{|c|}{SP} & \multicolumn{1}{c|}{ 0.7732} & \multicolumn{1}{c|}{0.6811} & \multicolumn{1}{c|}{1.277}  \\\cline{1-4}  
   \multicolumn{1}{|c|}{CoSaMP} & \multicolumn{1}{c|}{-} & \multicolumn{1}{c|}{0.3225} & \multicolumn{1}{c|}{ 0.552}  \\\cline{1-4}  
\end{tabular}
\label{table_omni}
\end{center}
\end{table}

\section{Conclusion}
\label{conc}
We propose a post-processing framework of DNN, i.e., TSN, to improve the performance of the target DNN in SR. The proposed TSN is featured by performing a tree search for support retrieval via the DNN-based index selection and extended support estimation. Experimental results demonstrate that TSN is superior to its target DNN-SR and traditional SR algorithms to uniformly recover synthesis sparse signals using diverse types of the sensing matrix, 
especially when the signal is difficult to be recovered, i.e., when the sampling rate $m/n$ is sufficiently small. This result implies that the performance limitation of SR can be resolved by applying the dynamic programming (tree search) to DNN.

Given that TSN solves the common inverse problem in CS and exploits a characteristic inherent in the signal distribution, we expect it to be utilized in various CS applications to enhance the signal reconstruction performance. 
We have verified its validity by testing two typical applications of SR.

\appendices
\section{The training algorithm for DNN-SR used in TSN and its effect} 
\label{trd}

\begin{algorithm}
\footnotesize
   \caption{TrainDNN($\Phi,k_1,k_2,s_d,s_b,n_e,v_\textup{SNR$_{\textup{dB}}$}$)} 
\label{alg1}
\begin{algorithmic}[1]
   \Input{ $\Phi \in \mathbb{K}^{m \times n}$, $(k_1,k_2,s_d,s_b,n_e) \in \mathbb{N}^4$, $v_\textup{SNR$_{\textup{dB}}$} \in \mathbb{R}$}
 \For{$l=1$ {\bfseries to} $n_{e}$}
    \For{$i=1$ {\bfseries to} $s_d$}
    \State $x_i  \gets v \in \mathbb{K}^{n} \textup{ s.t. } s_v:=|\supp(v)| \in \mathbb{N}$ is uniformly sampled from $\{k_{1}:k_{2}\}$ and nonzero elements of $v$ are sampled from the probability distribution $p_{x_0}(\, \, \cdot \mid s_v)$ of the signal vector $x_0$ whose number of nonzero elements is $|\supp(v)|$.
\State  $\alpha \gets $  uniformly sampled from $0$ to $1$
\State $o_i \in \mathbb{K}^{m} \gets $ uniformly sampled from sphere $\mathbb{S}_m$ of dimension $m-1$  with radius 1.
\State $o_i \gets \alpha\cdot \beta \cdot o_i$ where $\beta := \left \| z_i \right \| \cdot 10^{- v_\textup{SNR$_{\textup{dB}}$}/20}$ and $z_i := \Phi x_i$
\State $y_i \gets z_i+o_i$
    \EndFor

\For{$q=1$ {\bfseries to} $\lceil s_d / s_b \rceil$}
\State $L(\theta) \gets \sum\limits_{p \in \Gamma}[ \textup{ce(}f_{\theta}(y_p),\textup{u}_n(\supp(x_p))) /|\Gamma| ]$ where $\Gamma:=[(q-1)\cdot  s_b+1: \min(q \cdot  s_b, s_d)]$
\State $\eta_q \gets \textup{Set the learning rate for the gradient update}$
\State $\theta \gets \textup{Update}(L(\theta), \eta_q)$
\EndFor
\EndFor
\State {\bfseries return} DNN-SR $f_{\theta}(\cdot): \mathbb{K}^{m} \rightarrow \mathbb{T}^n$ where $\theta$ is the set of its training parameters 
\end{algorithmic}
\end{algorithm}

\label{tr1}
We propose Algorithm \ref{alg1} to train the DNN-SR defined by $f_{\theta}(\cdot)$ with its training parameter $\theta$. The algorithm considers noisy training data whose sparsity is in a certain range $\{k_{1}:k_{2}\}$ to improve robustness to noise and recover $x_0$ whose sparsity is unknown but its range is given as $\{k_{1}:k_{2}\}$. To reduce the difference between prediction error in training and test data, this method is based on the online learning method proposed in \cite{he2017bayesian} such that training data is updated for each epoch. 
 
In Algorithm \ref{alg1}, $n_{e}$ is the number of epochs, $s_b$ is the size of data batch, $s_d$ is the size of training data per epoch, and $v_\textup{SNR$_{\textup{dB}}$}$ is SNR in decibels. We assume that sparsity $|\Omega|$ of $x_0$ is unknown, and its lower and upper bounds are given as $k_1$ and $k_2$, respectively. The training process uses these bounds to generate the training data.
For each epoch, steps 3--7 generate training data $(x_i,y_i)_{i=1}^{s_d}$  given sensing matrix $\Phi$, where $x_i$ and $y_i$ are the synthetic signal vector and its measurement vector, respectively.
Specifically, step 3 generates signal vector $x_i$ from the conditional probability of signal vector $x_0$, given its sparsity $\bar s$ uniformly sampled from $\{k_{1}:k_{2}\}$, which is expressed as
\begin{align}\nonumber
p_{x_0}(x \mid \bar s):= \frac{p_{x_0}(x) \cdot 1(|\supp(x)|=\bar s)}{\sum_{\tilde x \in \mathbb{K}^n \textup{ s.t. } |\supp(\tilde x)|=\bar s}p_{x_0}(\tilde x)}, 
\end{align}
where $p_{x_0}(x)$ is the distribution of $x_0$ and $1(\cdot)$ is the indicator function that outputs $1$ if its input statement is true, otherwise $0$. Measurement vector $y_i$ in step 7 corresponds to $\Phi x_i$ plus a vector $o_i$ randomly sampled such that its magnitude ranges from 0 to the expected noise magnitude $\beta$ determined by the SNR. 
Thus, the objective of Algorithm \ref{alg1} is to train DNN-SR such that for any sparse vector $\bar x$ whose sparsity is  in $\{k_{1}:k_{2}\}$, the network outputs the support of $\bar x$ from any input measurement vector $\bar y$ in the Euclidean ball of radius $ \bar \beta$, centered around $\Phi \bar x$, such that $\left \| \bar y - \Phi \bar x \right \|  \leq \bar \beta$, where $\bar \beta := \left \| \Phi \bar x\right \| \cdot 10^{- v_\textup{SNR$_{\textup{dB}}$}/20}$.

To satisfy the objective,  we define the following loss function in step 10, which is minimized when each signal support in training data is equal to its estimate generated by the trained DNN-SR output:
\begin{align}\nonumber
L(\theta):= \mathbb{E}_{(y,x)} [ \textup{ce(}f_{\theta}(y),\textup{u}_n(\supp(x))) ],
\end{align}
where $\textup{u}_n(\Delta)$ with index set $\Delta \subseteq \{1:n\}$ is the function that returns an $n$-dimensional vector $v \in \mathbb{R}^n$, whose support is $\Delta$ and nonzero elements are equal to $1/|\Delta|$, and $\textup{ce}(a,b):= -\frac{1}{n} \sum_{i=1}^{n} b_i \log \, a_i $ is the cross entropy function for $n$-dimensional vectors $a$ and $b$. From loss function $L(\theta)$, parameter $\theta$ of the DNN-SR is updated in steps 11--12 through function $\textup{Update}(L(\theta), \eta)$, where $\eta $ is the learning rate. In Sections \ref{ne} and \ref{Application}, an adaptive method for stochastic gradient descent, called RMSprop, is used for $\textup{Update}(L(\theta), \eta)$. 

Optimizing DNN-SR by generating noisy training data contributes to the performance improvement of DNN-SR compared to no consideration of noisy training data. To show this, two GFLSTM networks were trained by using noiseless and noisy synthetic data with SNR equal to $5$ dB, respectively. {That is, these two GFLSTM networks were trained by Algorithm \ref{alg1} whose input $v_\textup{SNR$_{\textup{dB}}$}$ is set to $\infty$ and $5$, respectively.} Then, Table \ref{table1} compares the performance of both trained networks with sparsity $s$ of $x_0$ varied from 1 to 6 and noisy test data (SNR = $5$ dB), and then demonstrates this argument.
\begin{table}
\centering
\footnotesize
\caption{Probability where the true support $\Omega$ is equal to the top $|\Omega|$ indices of the trained DNN-SR output when SNR = $5$ dB}
\begin{tabular}{|m{0.2cm}|m{0.2cm}|m{0.2cm}|m{0.2cm}|m{0.2cm}|m{0.2cm}|m{0.2cm}|m{0.2cm}|m{0.2cm}|}
  \hline
  \multicolumn{1}{|c|}{Data type $\setminus$ $|\Omega|$ } & \multicolumn{1}{c|}{$1$} & \multicolumn{1}{c|}{$2$} & \multicolumn{1}{c|}{$3$}  & \multicolumn{1}{c|}{$4$} & \multicolumn{1}{c|}{$5$} & \multicolumn{1}{c|}{$6$} \\\hline 
   \multicolumn{1}{|c|}{Noisy data} & \multicolumn{1}{l|}{$\boldsymbol{1.00}$} & \multicolumn{1}{l|}{$\boldsymbol{1.00}$}  & \multicolumn{1}{l|}{$\boldsymbol{0.98}$}& \multicolumn{1}{l|}{$\boldsymbol{0.94}$} & \multicolumn{1}{l|}{$\boldsymbol{0.76}$} & \multicolumn{1}{l|}{$\boldsymbol{0.45}$} \\\cline{1-7} 
   \multicolumn{1}{|c|}{Noiseless data} & \multicolumn{1}{l|}{$0.92$} & \multicolumn{1}{l|}{$0.81$} & \multicolumn{1}{l|}{$0.75$}& \multicolumn{1}{l|}{$0.71$} & \multicolumn{1}{l|}{$0.56$} & \multicolumn{1}{l|}{$0.31$} \\\cline{1-7} 
\end{tabular}
\label{table1}
\end{table}

\section{TSN specification}
\label{tsns}

The proposed TSN is detailed in Algorithm \ref{alg5}, whose objective is to recover $k$-sparse signal $x_0$ and its support $\Omega$ from measurement vector $y$ and sensing matrix $\Phi$. Let us define $\Omega(k)$, called the $k$-support of $x_0$, as any index set satisfying $\Omega(k) \supseteq \Omega$ and $|\Omega(k)|=k$.

TSN is designed by combining three algorithms, i.e., Algorithms \ref{alg2}, \ref{alg3}, and \ref{alg4} presented in the sequel, and a tree search to determine $\Omega(k)$. Algorithm \ref{alg2} (expand) generates multiple index sets as partial estimates of $\Omega(k)$ from an input index set. Each set generated by Algorithm \ref{alg2} corresponds to a node in the tree, and hence this algorithm allows expanding the tree by generating multiple child nodes from a given parent node. Algorithm \ref{alg3} (prune) reduces the number of index sets, which represent leaf nodes in the tree, to a fixed value $g$ for reducing complexity and allows to terminate TSN in an intermediate step. Algorithm \ref{alg4} (initialize) is executed before the tree search to provide an initial estimate of $\Omega(k)$, which is iteratively updated through the tree search to determine the final estimate of $\Omega(k)$. The three algorithms are detailed in Section \ref{suba} and integrated into the proposed TSN in Section \ref{tsnd}. 
\subsection{Algorithms composing TSN}
\label{suba}

%

\begin{algorithm}
\footnotesize
   \caption{Expand($y,\Phi, \Gamma,q$)}
\label{alg2}
\begin{algorithmic}[1]
   \Input { $y \in \mathbb{K}^{m}, \Phi \in \mathbb{K}^{m \times n}, \Gamma \subseteq \{1:n\},q \in \mathbb{N}$, a trained DNN-SR $f_{\theta}(\cdot): \mathbb{K}^{m} \rightarrow \mathbb{T}^n$}
\State $v:=(v_1,..,v_n) \gets  f_{\theta}(P^{\perp}_{\mathcal{R}(\Phi_{\Gamma})}y)$
\State $u \gets \min(n-|\Gamma|, q)$
\State $Z:=\{z_1,...,z_u\} \gets \{ \textup{$u$-largest indices $i$ of $v_i$ s.t. $i \in \{1:n\} \setminus \Gamma$} \}$
    \For{$j=1$ {\bfseries to} $u$}
    \State $\bar \Delta_j \gets z_j  \cup \Gamma$
    \EndFor
\State {\bfseries return} The family $\boldsymbol{S}:=(\Delta_i)_{i=1}^{|\boldsymbol{S}|}$ of unique index sets in $\{\bar \Delta_1,...,\bar \Delta_u\}$
\end{algorithmic}
\end{algorithm}
\subsubsection{Algorithm \ref{alg2} (expand)} 
\label{ex1}
Algorithm \ref{alg2} takes partial support estimate $\Gamma \subseteq \{1:n\}$ of $\Omega(k)$ as one of its inputs. Step 1 generates probability vector $v:= f_{\theta}(P^{\perp}_{\mathcal{R}(\Phi_{\Gamma})}y)$ from the trained DNN-SR $f_{\theta}(\cdot)$ output and residual vector $P^{\perp}_{\mathcal{R}(\Phi_{\Gamma})}y$. Then, step 3 generates a remaining support estimate $Z:=\{z_1,...,z_u\} \subseteq \{1:n\} \setminus \Gamma$ of size $u$ by selecting positions of the $u$-largest elements in the vector $v$. Next, Algorithm \ref{alg2} generates family $\boldsymbol{S}:=(\Delta_i)_{i=1}^{|\boldsymbol{S}|}$ of index sets (child nodes) to expand the tree by adding $z_i$ in $Z$ to index set $\Gamma$ (parent node).

In the noiseless case, as described in Section \ref{tr1}, the DNN-SR $f_{\theta}(\cdot)$ in step 1 is trained to provide the index set $Z$ in step 3  such that the following condition is satisfied for a vector $\tilde x \in \mathbb{K}^{n}$.
\begin{align}\label{abc1}
\left \| P^{\perp}_{\mathcal{R}(\Phi_{\Gamma})}y - \Phi_{H}\tilde x^{H} \right \|  = 0
\end{align}  


Then, given that $Z$ satisfying (\ref{abc1}) includes the true remaining support $\Omega \setminus \Gamma$ from Lemma \ref{lem1}, the index in the true remaining support, which is not in the parent node, can be added to its child nodes through Algorithm \ref{alg2}.

\begin{algorithm} 
\footnotesize
   \caption{Prune($\boldsymbol{S},\boldsymbol{I},\check\Omega,\check r,y,\Phi,k,g,z,\epsilon$)}
\label{alg3}
\begin{algorithmic}[1]
   \Input { $y \in \mathbb{K}^{m}, \Phi \in \mathbb{K}^{m \times n}, \boldsymbol{S}:=(\Delta_i)^{|\boldsymbol{S}|}_{i=1}$ where $\Delta_i$ is a subset of $\{1:n\}$ for $i \in \{1:|\boldsymbol{S}|\}$, $\boldsymbol{I}:=(\Lambda_i, \bar r_i)^{|\boldsymbol{I}|}_{i =1}$ where $\Lambda_i$ is a subset of $\{1:n\}$ and $\bar r_i \in \mathbb{R}$ for $i \in \{1:|\boldsymbol{I}|\}$, $(k,g) \in \mathbb{N}^2$, $\epsilon \in \mathbb{R}$, $\check\Omega \subseteq \{1:n\}$, $\check r \in \mathbb{R}$, a trained DNN-SR $f_{\theta}(\cdot): \mathbb{K}^{m} \rightarrow \mathbb{T}^n$}
   \Initialize $ \Pi_{i}=\Delta_i $ for ${i \in \{1:|\boldsymbol{S}|\}}$, and $(\Pi_{i+|\boldsymbol{S}|}, r_{i+|\boldsymbol{S}|}) = (\Lambda_i,  \bar r_i) $ for ${i \in \{1:|\boldsymbol{I}|\}}$, $(t,e)=(1,0)$
  \Output $\boldsymbol{D}:=(\Delta_i, r_i)_{i \in \{1:|\boldsymbol{D}|\}}$ where $\Delta_i$ is a subset of $\{1:n\}$ for $i \in \{1:|\boldsymbol{D}|\}$, $e \in \{0,1\}$, $\check\Omega \subseteq \{1:n\}$, $\check r \in \mathbb{R}$

\While{$ t \leq  |\boldsymbol{S}| \textup{ and } e=0$} 
    \State $v:=(v_1,..,v_n) \gets  f_{\theta}(P^{\perp}_{\mathcal{R}(\Phi_{\Pi_t})}y)$
    \State $Z:=\{z_1,...,z_u\} \gets \{ \textup{$u$-largest indices $i$ of $v_i$ s.t. $i \in \{1:n\}\setminus \Pi_t$} \}$ where $u:= m -1 -|\Pi_t|$
    \State $\Psi_t \gets \Pi_t \cup Z$
\State $\bar{x}^{{\Psi_t}}  \gets (\Phi^{*}_{{\Psi_t}}\Phi_{{\Psi_t}}+ \eta^2 D({\gamma})^{-1})^{-1}\Phi^{*}_{{\Psi_t}} y  $  where $({\gamma},\eta)$ is obtained by $\underset{{{\bar \gamma} \in \mathbb{R}^{m},\bar\eta \in \mathbb{R}}}{\arg\min\limits} L(\Phi_{{\Psi_t}},y)$

    \State $\bar \Omega_{t} \gets \{ \textup{$k$-largest indices $i$ of $\abs(\bar x_i)$}\}$
    \State ${r_t} \gets \underset{x \in \mathbb{K}^n}{\min} \left \| \Phi_{{\bar\Omega}_{t}}x^{{\bar\Omega}_{t}} - y \right\|$
    \If{${r_t} \leq \epsilon$}  
	\State {$e \gets 1$}
    \EndIf
    \State $t \gets t+1$
    \EndWhile
\If{$g \neq 1$ or $e = 1$}
    \State $ \Theta  \gets  \{\textup{$g$-smallest indices $i$ of ${r_i}$ in $\{1:|\boldsymbol{S}|+|\boldsymbol{I}|\}$}\}$
    \State $ \boldsymbol{D}  \gets  (\Pi_i, r_i)_{ i \in \Theta }$
    \State $q \gets  \textup{the smallest index $i$ of $r_i$}$ in $\Theta$
    \If{$\max(\check r ,\epsilon) \geq r_{q}$} 
	\State $(\check \Omega, \check r  ) \gets (\bar \Omega_{q}, r_{q} )$
\EndIf
\Else
\State $ \Theta  \gets  \{\textup{$z$-smallest indices $i$ of ${r_i}$ in $\{1:|\boldsymbol{S}|+|\boldsymbol{I}|\}$}\}$
\State Define an index set $J:=\underset{ i \in \Theta}{\cup} \Pi_i$ and define a pair $(\dot \Omega,\dot r)$ of the $k$-support estimate and its signal error obtained by the same procedure in steps $5$--$7$ where $\Psi_t$ is replaced with $J$.
\State $ \boldsymbol{D}  \gets  (J, \dot r)$
    \If{$\max(\check r ,\epsilon) \geq \dot r$} 
	\State $(\check \Omega, \check r  ) \gets (\dot \Omega, \dot r )$
    \EndIf 
    \If{$\epsilon \geq \check r$} 
	\State $e \gets 1$
    \EndIf 
\EndIf 
\State {\bfseries return} ($e,\boldsymbol{D},\check \Omega, \check r $)

\end{algorithmic}
\end{algorithm}

\subsubsection{Algorithm \ref{alg3} (prune)}  
\label{mwhy}

Suppose that there exists a family $\boldsymbol{S}:=(\Delta_i)^{|\boldsymbol{S}|}_{i=1}$ of index sets and another family $\boldsymbol{I}:=(\Lambda_i, \bar r_i)^{|\boldsymbol{I}|}_{i=1}$ of pairs, where $\Lambda_i$ and $\bar r_i$ in each pair in $\boldsymbol{I}$ are an index set and a signal error obtained from the set $\Lambda_i$, respectively. From each $\Delta_i$ in family $\boldsymbol{S}$, Algorithm \ref{alg3} determines the $k$-support estimate and its signal error. Then, the algorithm reduces the number of index sets in $\boldsymbol{S}$ and $\boldsymbol{I}$ by selecting a family $\boldsymbol{D}$ of $g$ index sets with the smallest signal errors among their union $\{\Pi_1:=\Delta_1,...,\Pi_{|\boldsymbol{S}|}:=\Delta_{|\boldsymbol{S}|},\Pi_{|\boldsymbol{S}|+1}:=\Lambda_1,...,\Pi_{|\boldsymbol{S}|+|\boldsymbol{I}|}:=\Lambda_{|\boldsymbol{I}|}\}$. Given that the index sets in $\boldsymbol{S}$ and $\boldsymbol{I}$ represent leaf nodes in the search tree, the algorithm prunes the tree to reduce the number of nodes.

To obtain the signal error given partial support estimate $\Pi_t$, extended support estimate $\Psi_t$ of size $m-1$  for $t \in \{1:|\boldsymbol{S}|\}$ is obtained by adding remaining support estimate $Z$ of $u:=m-1-|\Pi_t|$ indices to index set $\Pi_t$ in steps 2--4. Set $Z \subseteq \{1:n\} \setminus \Pi_t$ in step 3 is obtained through the DNN-based index selection by selecting the position of the $u$-largest elements in probability vector $v:=f_{\theta}(P^{\perp}_{\mathcal{R}(\Phi_{\Pi_t})}y)$ generated from the trained DNN-SR and residual vector $P^{\perp}_{\mathcal{R}(\Phi_{\Pi_t})}y$.\footnote{Similar to step 3 of Algorithm \ref{alg2}, set $Z$ in step 3 of Algorithm \ref{alg3} includes the remaining support, $\Omega \setminus \Pi_t$, if DNN-SR defined by $f_{\theta}(\cdot)$ is ideally trained, as explained in Sections \ref{tr1} and \ref{ex1}.  
}  Steps 5--6 generate estimate $\bar \Omega_t$ of $\Omega(k)$ by selecting $k$ indices from extended support estimate $\Psi_t$. Specifically, step 5 corresponds to a ridge regression with parameters $(\gamma,\eta)$, which are obtained by minimizing the following cost
\begin{align}\label{sble}
(\gamma,\eta) = \underset{{\bar \gamma \in \mathbb{R}^{a},\bar \eta \in \mathbb{R}}}{\arg\min\limits} L(\bar \Phi,y):=\, \log |\Sigma|+ y^{\top}\Sigma^{-1}y,
\end{align}
where $\bar \Phi = \Phi_{\Psi_t} \in \mathbb{K}^{m \times a}$ is a submatrix of $\Phi$, whose columns are indexed by $\Psi_t$, $\Sigma = (\bar \eta^{-2} {\bar\Phi}^{*}{\bar\Phi}+ D(\bar \gamma))^{-1}$, and $D(\bar \gamma)$ is the diagonal matrix whose $i$-th diagonal element is $\gamma_i$ in $\bar \gamma=(\gamma_1,...,\gamma_a)^{\top}$. The ridge regression is derived through the Bayesian methodology \cite{wipf2004sparse} for solving linear inverse problems to determine vector $\bar x \in \mathbb{K}^n$ such that (\ref{ls1}) holds. 
\begin{align}\label{ls1}
y=\Phi \bar x+w \,\,\,\, \textup{ and }\,\,\,\,  \supp(\bar x) \subseteq \Psi_t
\end{align}
Furthermore, in the noisy case, the ridge regression considers the noise magnitude corresponding to $\eta$ to improve noise robustness compared to the least-squares regression. Next, we obtain an approximate solution of $(\ref{sble})$ using SBL  to reduce complexity. In the numerical experiments, we set the regularization parameter and maximum iteration number of  the SBL used in TSN to $\max[(\left \| y\right \| \cdot 10^{-v_\textup{SNR$_{\textup{dB}}$}/20})^2/m,10^{-4}]$ and $10$, respectively. In the noiseless case, parameters $(\gamma,\eta)$ are set to zero, and hence the ridge regression is equivalent to the least-squares regression, i.e., $(\Phi^{*}_{{\Psi_t}}\Phi_{{\Psi_t}})^{-1}\Phi^{*}_{{\Psi_t}} y=\underset{x}{\arg\min}\left \| \Phi_{\Psi_t} x - y \right \|$. The least-squares method has a lower complexity than SBL and provides $x_0$ as the unique solution of (\ref{ls1}) in the noiseless case ($w=0$) if $ \Psi_t \supseteq \Omega$, $|\Psi_t| \leq m$, and $\Phi_{\Psi_t}$ has full column rank. Note that the probability of satisfying $\Psi_t \supseteq \Omega$ increases with the dimension of $\Psi_t$. In addition, if $ \Psi_t \supseteq \Omega$ holds, the inversion problem of SR can be simplified as a problem where $\Phi$ is replaced by submatrix $\Phi_{\Psi_t}$. However, we also note that the signal error $\underset{x}{\min}\left \| \Phi_{\Psi_t} x - y \right \|$ estimated by using any $\Psi_t$ of size $m$  is equal to zero, even if $ \Psi_t \nsupseteq \Omega$. To avoid this trivial case and maximize the probability of satisfying $\Psi_t \supseteq \Omega$, we set the dimension of extended support estimate $\Psi_t$ to $m-1$.\footnote{The extended support is defined by an index set including $\Omega$. Some popular existing SR algorithms gOMP \cite{wang2012generalized}, SP \cite{dai2009subspace}, and CoSaMP \cite{needell2009cosamp} exploit an extended support estimate in the process to estimate $\Omega$. However, these algorithms do not utilize a DNN-based index selection. Besides, the size of the extended support estimate $\Psi_t$ in TSN is set to $m-1$, which is indepedent of $s$, to maximize the probability satisfying $ \Psi_t \supseteq \Omega$, whereas its size in gOMP, SP, and CoSaMP is $t \cdot \min(s,\left \lfloor {m/t} \right \rfloor)$ , $2s$, and $3s$, respectively,  for a constant $t$ smaller than $s$, where $s:=|\Omega|$ is the sparsity of $x_0$.}

Then, we calculate signal error $r_t:= \underset{x}{\min} \left \| \Phi_{{\bar \Omega_t}}x - y \right\|$ using $k$-support estimate $\bar \Omega_t$, which is obtained in step 7. If there exists a norm $r_i$ for $i \in \{1:|\boldsymbol{S}|\}$ smaller than a threshold $\epsilon$, $\bar\Omega_i$ is considered as the true $k$-support and Algorithm \ref{alg3} terminates by returning $\bar\Omega_i$ and setting Boolean parameter $e$ to true, indicating the successful termination of TSN. Otherwise, if the number $g$ of index sets remained after pruning is larger than 1, Algorithm \ref{alg3} goes to step 14 to select, from family $(\Pi)_{i=1}^{|\boldsymbol{S}|+|\boldsymbol{I}|}$ of index sets, set $\Theta$ of $g$ indices $i$ minimizing signal error $r_i$. Then, Algorithm \ref{alg3} returns set $\boldsymbol{D}$ composed of pairs $(\Pi_t, r_t)_{t \in \Theta}$ in step 15; Algorithm \ref{alg3}  only leaves $g$ index sets $(\Pi)_{i\in \Theta}$ from the $|\boldsymbol{S}|+|\boldsymbol{I}|$ index sets $(\Pi)_{i=1}^{|\boldsymbol{S}|+|\boldsymbol{I}|}$. Note that in step 16 a $k$-support estimate $\bar \Omega_{q}$, where $q$ is the argument $i$ with the minimum residual error $r_i$ for $i \in \Theta$, is selected among $g$ index sets $(\Pi)_{i\in \Theta}$. In addition, Algorithm \ref{alg3} takes pair $(\check\Omega,\check r)$ as its input, where $\check\Omega$ is an index set for the $k$-support estimate and $\check r$ is its signal error. Then, the algorithm replaces the input pair $(\check\Omega,\check r)$ by $(\bar \Omega_{q},r_q)$ in step 18, provided that signal error $r_q$ generated from $k$-support estimate $\bar \Omega_{q}$ is smaller than $\check r$. In the case where only one index set is remained after pruning, i.e., $g=1$, Algorithm \ref{alg3} generates  set $\Theta$ of $z$ indices $i$ minimizing signal error $r_i$ in step 21 and returns index set $J:=\underset{ i \in \Theta}{\cup} \Pi_i$ by unioning $z$ sets $(\Pi_i)_{i \in \Theta}$ indexed by $i \in \Theta$ in step 22; the input value $z$ of TSN represents the number of sets to be combined as one set. Similarly with step 18, Algorithm \ref{alg3} updates pair $(\check\Omega,\check r)$ by obtaining a $k$-support estimate $\dot \Omega$ and its signal error $\dot r$ from index set $J$ in steps 22 and 25.

\subsubsection{Algorithm \ref{alg4} (Initialize)} 
\begin{algorithm} 
\footnotesize
   \caption{Initialize($y,\Phi,k,\epsilon$)}
\label{alg4}
\begin{algorithmic}[1]
   \Input {$y \in \mathbb{K}^{m}, \Phi \in \mathbb{K}^{m \times n}$, $k \in \mathbb{N}$, $\epsilon \in \mathbb{R}$, a trained DNN-SR $f_{\theta}(\cdot): \mathbb{K}^{m} \rightarrow \mathbb{T}^n$}
   \Initialize $e=0$
\State $\Delta_{\textup{DNN}} \gets T_{k}(f_{\theta}(y)) $.
\State $\Delta_{\textup{OMP}} \gets \Gamma_{k}$  where $\Gamma_{i} := T_{1}(\Phi^* P^{\perp}_{\mathcal{R}(\Phi_{\Gamma_{i-1}})}y) \cup \Gamma_{i-1}$ for $i \in \{1:m-1\}$ and $\Gamma_{0}:=\{\}$ is the empty set
\If{$r_{\textup{DNN}}:=\underset{x \in \mathbb{K}^n}{\min} \left\| \Phi_{\Delta_{\textup{DNN}}}x^{\Delta_{\textup{DNN}}}-y\right\| \leq r_{\textup{OMP}}:=\underset{x \in \mathbb{K}^n}{\min} \left\| \Phi_{\Delta_{\textup{OMP}}}x^{\Delta_{\textup{OMP}}}-y\right\|$}
\State $\Psi \gets T_{m-1}(f_{\theta}(y))$ 
\Else
\State $\Psi \gets \Gamma_{m-1}$ where $\Gamma_{m-1}$ is defined in step 2
\EndIf
\State $\check{x}^{\Psi}  \gets (\Phi^{*}_{\Psi}\Phi_{\Psi}+ \eta^2 D(\gamma)^{-1})^{-1}\Phi^{*}_{\Psi} y  $  where $({\gamma},\eta)$ is obtained by $\underset{{{\bar\gamma} \in \mathbb{R}^{m},\bar\eta \in \mathbb{R}}}{\arg\min\limits} L(\Phi_{{\Psi}},y)$ in (\ref{sble})

    \State $\check \Omega \gets \{ \textup{$k$-largest indices of $|\check x_i|$ s.t. $i \in \Psi$} \}$
%
%
 \If{$\underset{x \in \mathbb{K}^n}{\min}\left \| \Phi_{\check \Omega} x^{\check \Omega} - y\right\| \leq \epsilon$} 
\State{$e \gets 1$}
\EndIf
\State {\bfseries return} $(\check \Omega \subseteq \{1:n\},$ $e \in \{0,1\})$
\end{algorithmic}
\end{algorithm}
Algorithm \ref{alg4} provides index set $\check \Omega$, an initial estimate for $\Omega(k)$ in TSN (Algorithm \ref{alg5}), given the tuple ($\Phi,y,k$) in the following three stages. 

The first stage (steps 1-2) aim to obtain two $k$-support estimates, $\Delta_{\textup{DNN}}$ and $\Delta_{\textup{OMP}}$, from the DNN-based and OMP-based selections, respectively.\footnote{In step 1, $k$-support estimate $\Delta_{\textup{DNN}}$ is obtained by selecting positions of the $k$-largest elements in probability vector $v:= f_{\theta}(y)$, which is generated from the trained DNN-SR $f_{\theta}(\cdot)$ output and the measurement vector $y$. In step 2, another $k$-support estimate $\Delta_{\textup{OMP}}$ is obtained by selecting $k$ indices via the OMP-based rule.} The second stage (steps 3--7) generates extended support estimate $\Psi$ of size $m-1$ from two $k$-support estimates. For this, the signal errors, $r_{\textup{DNN}}$ and $r_{\textup{OMP}}$, are obtained through estimates $\Delta_{\textup{DNN}}$ and $\Delta_{\textup{OMP}}$ to select one estimate among them. If $r_{\textup{DNN}}$  is smaller than $r_{\textup{OMP}}$, $\Delta_{\textup{DNN}}$ includes more elements in the true support $\Omega$ than $\Delta_{\textup{OMP}}$. Thus, extended support estimate $\Psi=T_{m-1}(f_{\theta}(y))$ is obtained via the DNN-based index selection in step $4$ when  $r_{\textup{DNN}}$  is smaller than $r_{\textup{OMP}}$. Otherwise, OMP-based index selection is used to generate set $\Psi $ as $\Gamma_{m-1}$ in step 6. The third stage  (steps 8--13) generates  theinitial estimate $\check \Omega$ for $\Omega(k)$ in TSN via the same approach from steps $5$--$6$ of Algorithm \ref{alg3} using extended support estimate $\Psi$.

\begin{algorithm} 
\footnotesize
   \caption{TSN($y,\Phi,k, \boldsymbol{\tau} :=\{q,z,\epsilon,\boldsymbol{l},\boldsymbol{g}, t_{\textup{max}}$\})}
\label{alg5}
\begin{algorithmic}[1]
   \Input {$y \in \mathbb{K}^{m}, \Phi \in \mathbb{K}^{m \times n}$, $(k,q,z) \in \mathbb{N}^3$, $(\epsilon,t_{\textup{max}}) \in \mathbb{R}^2$, $\boldsymbol{l}:=(l_i)_{i=1}^{b}$ and $\boldsymbol{g}:=(g_i)_{i=1}^{b}$ given a positive integer $b$ where $l_i \in \mathbb{N}$ and $g_i \in \mathbb{N}$ for $i \in \{1:b\}$ such that $\sum_{i=1}^b l_i\leq k$, a trained DNN-SR $f_{\theta}(\cdot): \mathbb{K}^{m} \rightarrow \mathbb{T}^n$}
   \Initialize $\rho := \min \limits_{x \in \{\bar x \mid p_{x_0}(\bar x)>0\}, i \in \supp(x)}  \frac{|x_i |}{2} $, $(a,o)=(1,0)$, $\boldsymbol{G}_{i} = (\check \Delta,\infty)$ where $\check \Delta$ is the empty index set for $i \in \{0:b\}$
\State $(\check \Omega,e) \gets$ Initialize($y,\Phi,k,\epsilon, \lambda$)
\If{$e=0$} 
\State $\check r \gets \underset{x}{\min}\left \| \Phi_{\check\Omega} x - y\right\| $
\While{$ o \leq  \sum_{i=1}^b l_i$} 
\State $o \gets o+l_a$
\For{$\Delta_i$ in $\boldsymbol{G}_{a-1} \,(=(\Delta_i,\cdot)_{i=1}^{|\boldsymbol{G}_{a-1}|})$} 
   \State $\boldsymbol{S}^i_{1} \gets$ Expand($y,\Phi, \Delta_i ,q $) 
 \State $\boldsymbol{T} \gets  \{\}$
   \For{$j =2$ to $ l_a$}
    \For{$ \Gamma \in \boldsymbol{S}^i_{j-1}$}
    \State $\boldsymbol{T} \gets \boldsymbol{T} \cup$ Expand($y,\Phi, \Gamma,q $)
    \EndFor
\State $\boldsymbol{S}^i_{j} \gets \boldsymbol{T} $ 
    \EndFor
    \State $(e,\boldsymbol{G}_{a},\check \Omega, \check r ) \gets \textup{Prune}(\boldsymbol{S}^i_{l_a},\boldsymbol{G}_{a},\check \Omega, \check r,y,\Phi,k,g_{a},z,\epsilon)$
\If{$e =1$ or the execution  time of TSN $ > t_{\textup{max}}$} 
\State{go to step $23$}
\EndIf
\EndFor
\State $a \gets a+1$
\EndWhile
\EndIf
\State $\bar{x}^{\check \Omega}  \gets (\Phi^{*}_{\check \Omega}\Phi_{\check \Omega}+ \eta^2 D(\gamma)^{-1})^{-1}\Phi^{*}_{\check \Omega} y  $  where the pair $(\gamma,\eta)$ is obtained by $\underset{{{\bar \gamma} \in \mathbb{R}^{m},\bar\eta \in \mathbb{R}}}{\arg\min\limits} L(\Phi_{{\check \Omega}},y)$
\State $\hat \Omega \gets \{ \textup{indices $i$ of $\abs(\bar x_i)$ such that  $\abs(\bar x_i)>\rho$} \}$
\State $(\hat x)^{\hat \Omega}  \gets \underset{x}{\arg\min}\left \| \Phi_{\hat \Omega} x - y\right\|$
\State {\bfseries return} $ (\hat \Omega \subseteq \{1:n\},\hat x \in \mathbb{K}^{n})$
\end{algorithmic}
\end{algorithm}

\subsection{TSN procedure}
\label{tsnd}

The proposed TSN (Algorithm \ref{alg5}) obtains $(\hat x, \hat \Omega)$ as an estimate of $x_0$ and its support $\Omega$ given input tuple ($\Phi,y,k$) and a DNN-SR $f_{\theta}(\cdot)$ trained by Algorithm \ref{alg1}.

First, TSN runs Algorithm \ref{alg4} in step 1 to obtain $k$-support estimate $\check \Omega$ and determine whether its signal error is below threshold $\epsilon$. If the error is over the threshold, i.e., $e=0$, TSN goes to steps 2--22 to update $k$-support estimate $\check \Omega$ through the tree search. Then, steps 23--26 in TSN return $(\hat x, \hat \Omega)$ from $k$-support estimate $\check \Omega$ obtained in steps 1--22.

\begin{figure}
\centering
  \footnotesize
  \subfigure{\includegraphics[width=11cm, height=6.4cm]{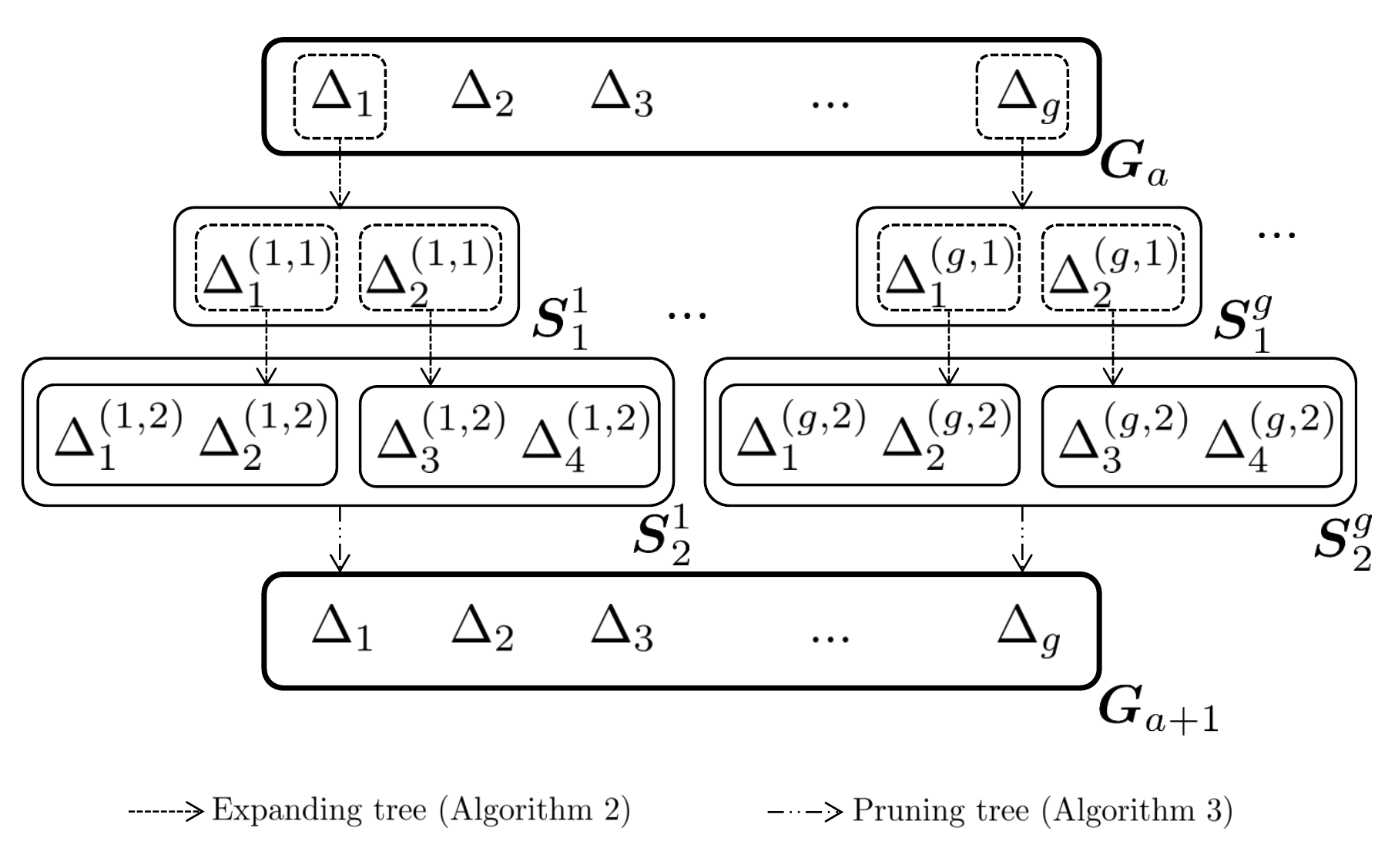}}

{\footnotesize From each index set $\Delta_i$ symbolizing a node (a parent node) in $\boldsymbol{G}_{a}$ for $i \in \{1:g\}$, 
$q^{l_a}$ child nodes, $\Delta^{(i,l_a)}_j$ for $j \in \{1:q^{l_a}\}$ in $\boldsymbol{S}^i_{l_a}$, are obtained such that $l_a$ indices are generated from the trained DNN-SR by recursively calling Algorithm \ref{alg2} and added to $\Delta_i$  to obtain each child node. $G_{a+1}$ is generated by reducing the size of the union of $\boldsymbol{S}^j_{l_a}$ for $j \in \{1:g\}$ as $g$ through Algorithm \ref{alg3}.\par}
\caption{Tree structure in TSN (Algorithm \ref{alg5}) between $G_a$ and $G_{a+1}$ when $(q,l_a,g_a)=(2,2,g)$}
\label{fig1}
\end{figure}

In step 15, $\boldsymbol{G}_{a}$ is a family of index sets generated from $\boldsymbol{G}_{a-1}$ by executing the loop in steps 4--21, where $a$ is a positive integer with initial value $1$ and incremented by $1$ at every iteration, and $\boldsymbol{G}_{0}$ is the empty set. The procedure to generate $\boldsymbol{G}_{a}$ from $\boldsymbol{G}_{a-1}$ is explained in the sequel and illustrated in Figure \ref{fig1}, which mainly consists of expanding and pruning the search tree.

(Steps 7--14: expand) Suppose that family $\boldsymbol{G}_{a-1}:=(\Delta_{i})_{i=1}^{|\boldsymbol{G}_{a-1}|}$ of index sets is available, where each set $\Delta_{i}$ represents a support estimate of size $\sum_{j=1}^{a-1}l_j$ for $i \in \{1:|\boldsymbol{G}_{a-1}|\}$. Algorithm \ref{alg2} is executed by taking each index set $\Delta_{i}$ as its input and returns a family $\boldsymbol{S}^i_{1}:=(\Lambda_{j})_{j=1}^{|\boldsymbol{S}^i_1|}$ of index sets in step 7, where each set $\Lambda_{l}$ for $l \in \{1:|\boldsymbol{S}^i_1|\}$ is a support estimate of size $1+|\Delta_{i}|$, which is an extension of $\Delta_{i}$ by adding one index. Hence, multiple child nodes $(\Lambda_{l})_{l \in \{1:|\boldsymbol{S}^i_{1}|\}}$ are generated from each parent node $\Delta_{i}$ to expand the search tree. By repeating the same procedure $l_a-1$ times in steps 9--14, TSN expands the tree such that family $\boldsymbol{S}^i_{l_a}$ of sets having $\sum_{i=1}^{a}l_i$ indices is obtained from each index set $\Delta_{i}$ in $\boldsymbol{G}_{a-1}$. 

(Step 15: prune) 
Once each family $\boldsymbol{S}^i_{l_a}$ of index sets is obtained for $i \in \{1:|\boldsymbol{G}_{a-1}|\}$ after step 14, Algorithm \ref{alg3} is executed in step 15 to update $\boldsymbol{G}_{a}$ by selecting $g_a$ index sets with smallest signal errors among index sets in $\boldsymbol{S}^i_{l_a}$ and $\boldsymbol{G}_{a}$.
Note that Algorithm \ref{alg3} (steps 6--7 in Algorithm \ref{alg3})  in step 15 generates pairs of a temporary $k$-support estimate and its signal error  from each index set in $\boldsymbol{S}^i_{l_a}$ and selects one $k$-support estimate $\dot \Omega$ with the smallest signal error among the pairs if $g_a$ is larger than 1.\footnote{$k$-support estimate $\dot \Omega$ is obtained in step 20  in Algorithm \ref{alg3} if $g_a$ equal to 1.} That is, set $\dot \Omega$ represents a $k$-support estimate obtained from the nodes existing at the tree depth $a$. Then,  Algorithm \ref{alg3} takes $\check \Omega$, which is the $k$-support estimate generated from nodes at the tree depth smaller than $a$, as its input and updates $\check \Omega$ as $\bar \Omega$  if the signal error $\check r$ corresponding to $\check \Omega$ is larger than that obtained from $\bar \Omega$.  If the signal error generated from $k$-support estimate $\check \Omega$ is below threshold $\epsilon$ ($e=0$), TSN terminates the tree search and goes to step 23 for estimating $x_0$ and $\Omega$ from $\check \Omega$; set $\check \Omega$ in step 23 indicates the final $k$-support estimate via the tree search in TSN.

Steps $23$--$25$ determine the final estimate $\hat \Omega$ of true support $\Omega$ from $k$-support estimate $\check \Omega$ by using the following two-stage process. In the first stage, signal estimate $\bar x:=(\bar x_1,...,\bar x_n)$, whose nonzero elements are supported on $\check \Omega$, is generated through the ridge regression shown in (\ref{sble}). Then, in the second stage, support $\Omega$ is estimated as $\hat \Omega$ by selecting indices $i$ of $\bar x_i$ whose absolute values $\abs(\bar x_i)$ are larger than a threshold $\rho$. We set this threshold to $v/2 $, where $v$ is the minimum among the absolute values of a possible signal vector.\footnote{If the sparsity $|\Omega|$ is given to TSN, steps $23$--$24$ in TSN can be omitted by setting the TSN input $k$ to $|\Omega|$ and $\hat \Omega$ to $\check \Omega$.} Then, estimated signal vector $\hat x$ is generated in step 25 by the least-squares method from support estimate $\hat \Omega$, and provided along with $\hat \Omega$ as the final TSN output.

\section{Proof of Lemma \ref{lem1}}
\label{apena}
From the assumptions that $|\Omega|<m$ and every $m$ columns in $\Phi$ exhibit full rank, the condition $P^{\perp}_{\mathcal{R}(\Phi_{\Gamma})}y \in \mathcal{R}(\Phi_{\Omega \setminus \Gamma})$ holds given any index set $\Gamma \subseteq \{1:n\}$.
Suppose that $P^{\perp}_{\mathcal{R}(\Phi_{\Gamma})}y$ does not belong to the following subspace $E$ ($P^{\perp}_{\mathcal{R}(\Phi_{\Gamma})}y \notin E$)
\begin{align}
E:=\cup_{J \subseteq \{1:n\} \textup{ s.t. } |J \cap (\Omega \setminus \Gamma)|<|\Omega \setminus \Gamma| \textup{ and }  |J|<m } \,\mathcal{R}(\Phi_{J}).
\end{align}
Then, $P^{\perp}_{\mathcal{R}(\Phi_{\Gamma})}y$ belongs to $\mathcal{R}(\Phi_{\Omega \setminus \Gamma}) \setminus E$. Given that (\ref{abc1}) implies that $P^{\perp}_{\mathcal{R}(\Phi_{\Gamma})}y \in \mathcal{R}(\Phi_{D})$, $P^{\perp}_{\mathcal{R}(\Phi_{\Gamma})}y \in \mathcal{R}(\Phi_{D}) \cap (\mathcal{R}(\Phi_{\Omega \setminus \Gamma}) \setminus E)$ holds for any index set $D$ satisfying (\ref{abc1}). Note that the condition $P^{\perp}_{\mathcal{R}(\Phi_{\Gamma})}y \in \mathcal{R}(\Phi_{D}) \cap (\mathcal{R}(\Phi_{\Omega \setminus \Gamma}) \setminus E)$ implies that ${D} \supseteq \Omega \setminus \Gamma$. Thus, ${D} \supseteq \Omega \setminus \Gamma$ holds   for any index set $D$ satisfying (\ref{abc1})  if $P^{\perp}_{\mathcal{R}(\Phi_{\Gamma})}y \notin E$.  Given that from the assumption for $\Phi$, the rank of $\mathcal{R}(\Phi_{\Omega \setminus \Gamma})$ is strictly larger than that of $\mathcal{R}(\Phi_{J}) \cap \mathcal{R}(\Phi_{\Omega \setminus \Gamma})$ for any index set $J \subseteq \{1:n\}$ such that  $|J \cap (\Omega \setminus \Gamma)|<|\Omega \setminus \Gamma|$ and $ |J|<m$, the event region satisfying $P^{\perp}_{\mathcal{R}(\Phi_{\Gamma})}y \in E$ has Lebesgue measure zero on the range space $\mathcal{R}(\Phi_{\Omega \setminus \Gamma})$ so that  the condition $P^{\perp}_{\mathcal{R}(\Phi_{\Gamma})}y \notin E$ is satisfied almost surely.

\section{Performance comparison given complex-valued measurements}
\label{apenb}
In this section, the performance of TSN is compared to other existing SR  algorithms shown in Section \ref{rvm} given complex-valued  measurements.  Let $\mathcal{CN}(a,b)$ denote the complex Gaussian distribution whose real and imaginary parts have mean $a$ and variance $b$, respectively. Real and imaginary parts of each nonzero elements in $x_0$ are independently and uniformly sampled from $-1$ to $1$, excluding the interval from $-$0.1 to 0.1, respectively. The measurement noise $w$ follows $\mathcal{CN}(0,\sigma^2_w)$ with $\sigma_w$ dependent on the given SNR. Algorithm \ref{alg1} whose input $(k_1,k_2)$ is set to $(1,15)$ is used to train the GFLSTM-based network $f_\theta(\cdot)$. The sparsity $s$ of $x_0$ is varied from $0$ to $12$, which is given to SR algorithms except for TSN. The input $k$ of TSN is fixed to $12$. The other parameters for simulation setting are set equal to those in Section \ref{se}.

\begin{figure}
\begin{center}
\subfigure[$\mathbb{P}( \hat x  = x_0 )$]{\includegraphics[width=5.0cm, height=5.0cm]{cpxstgaussian_noiseless_m20n100_s}}
\subfigure[$\mathbb{E}(\left \| x_0 - \hat x\right \| /\left \| x_0 \right \| )$]{\includegraphics[width=5.0cm, height=4.85cm]{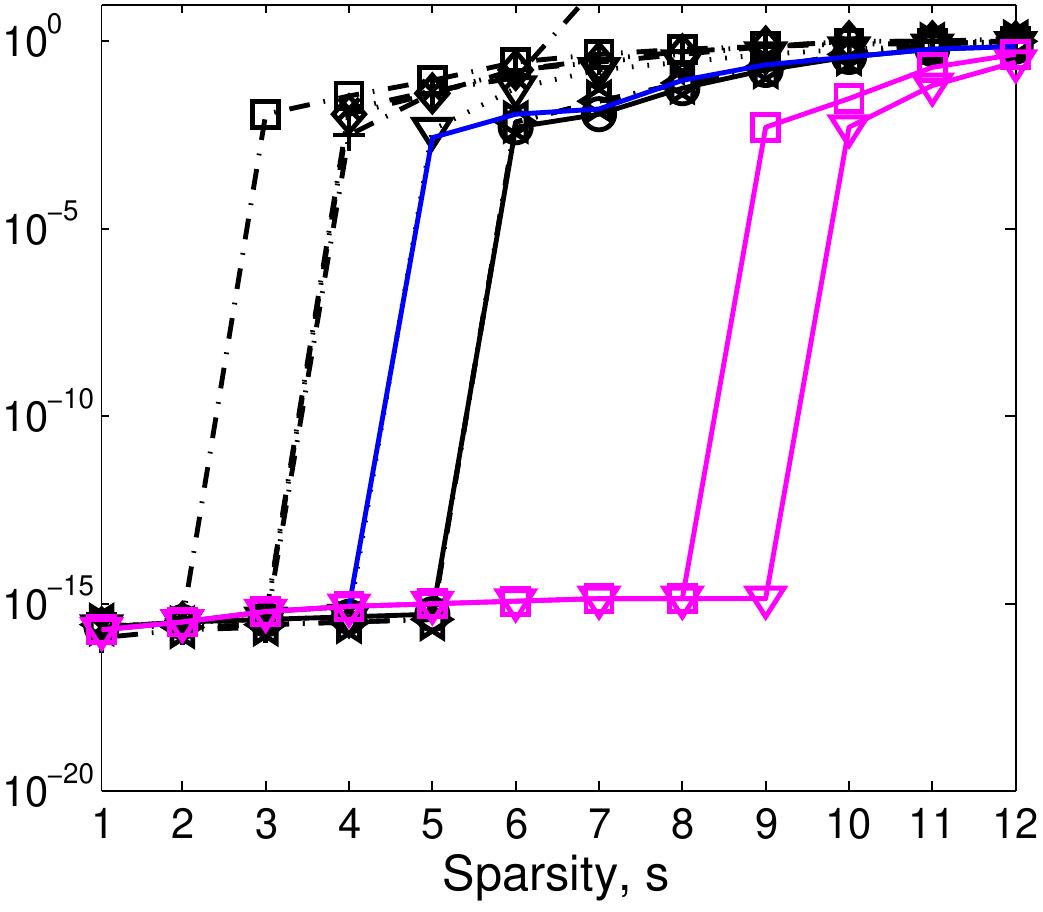}}
\subfigure[Execution time (seconds)]{\includegraphics[width=5.0cm, height=5.0cm]{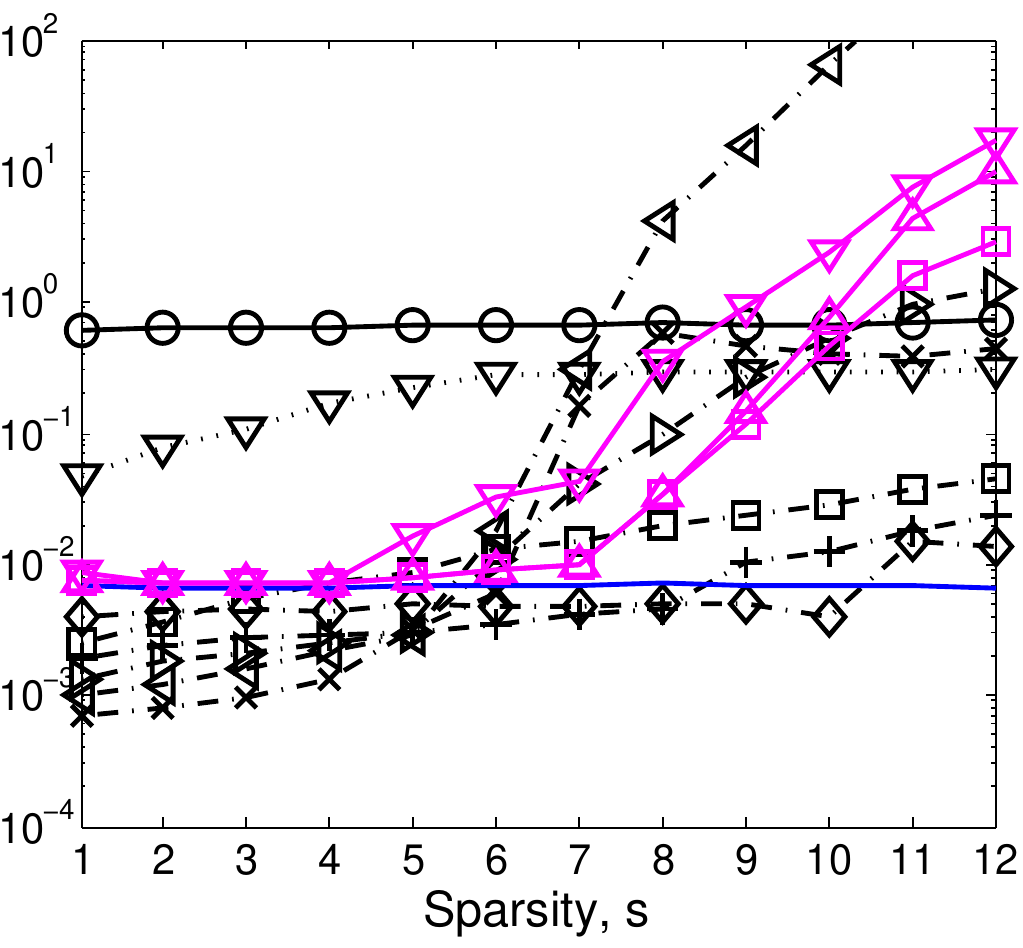}}
\subfigure[$\mathbb{P}(\left \| \Phi \hat x - \Phi x_0 \right \| \leq \left \| w \right \|)$]{\includegraphics[width=5.0cm, height=5.0cm]{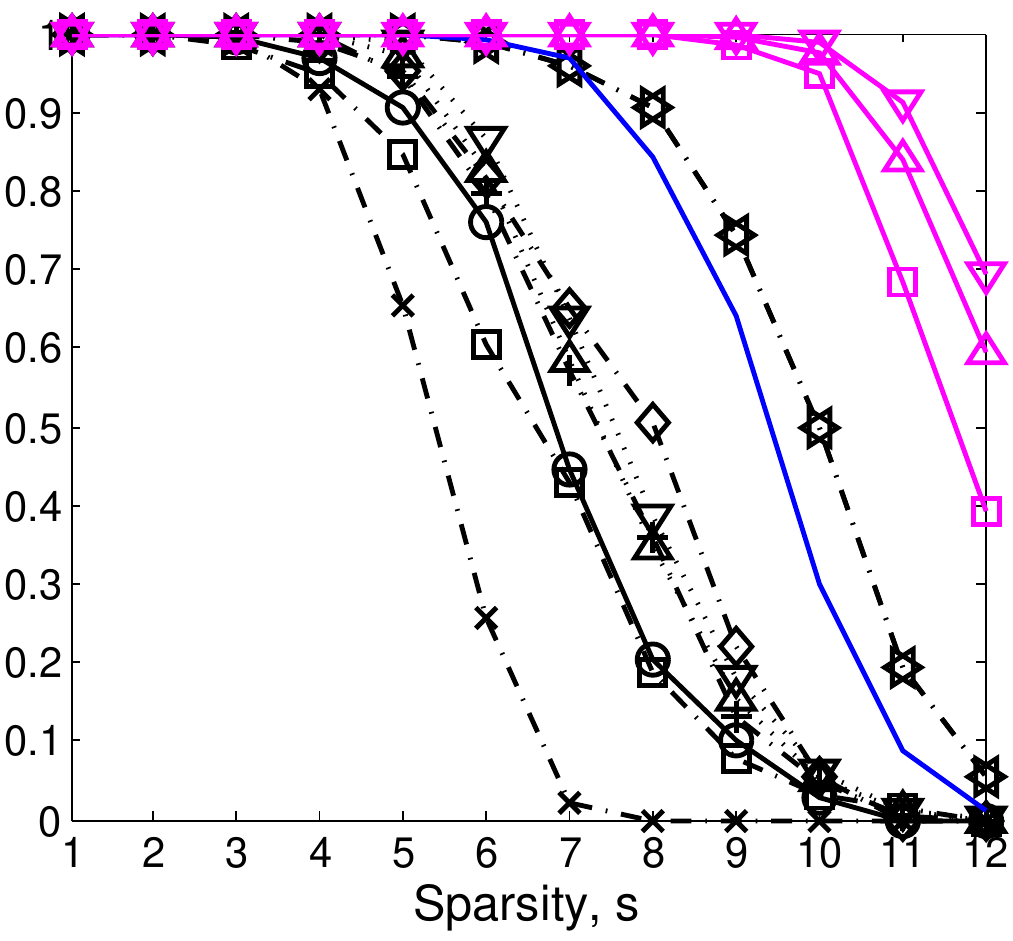}}
\subfigure[$\mathbb{E}(\left \| x_0 - \hat x\right \| /\left \| x_0 \right \| )$]{\includegraphics[width=5.0cm, height=4.85cm]{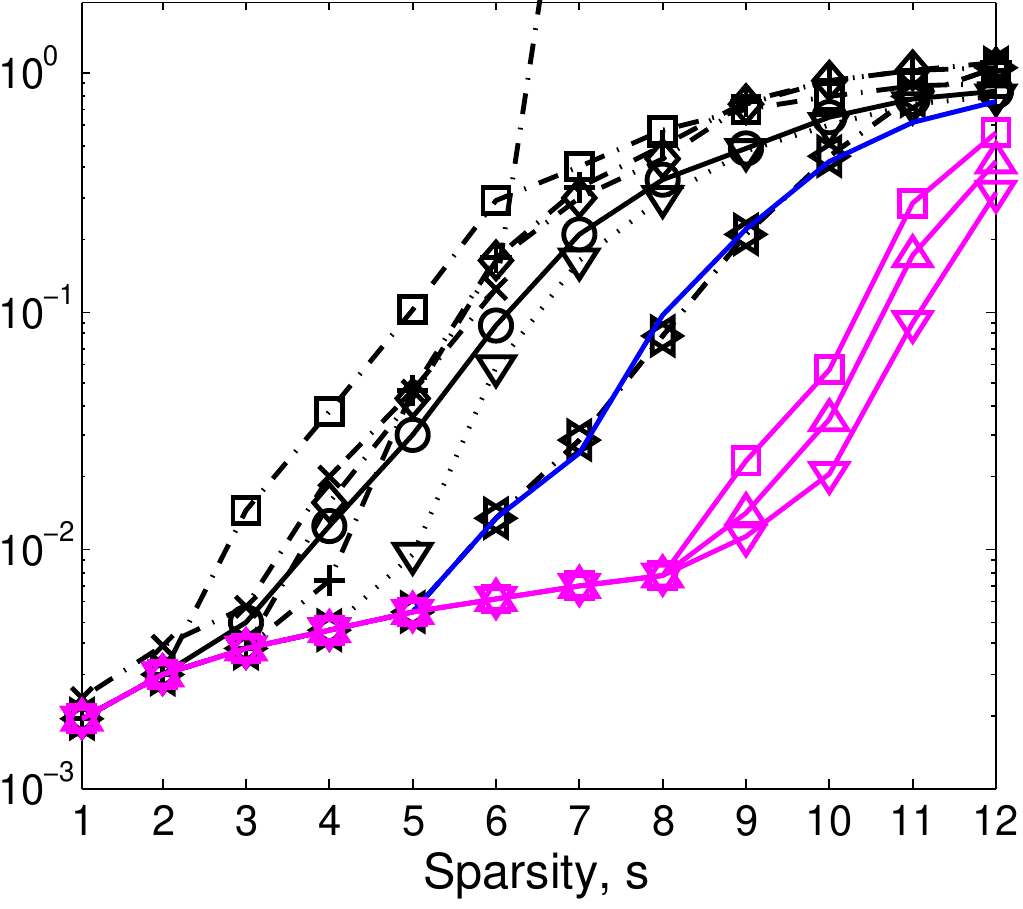}}
\subfigure[Execution time (seconds)]{\includegraphics[width=5.0cm, height=5.0cm]{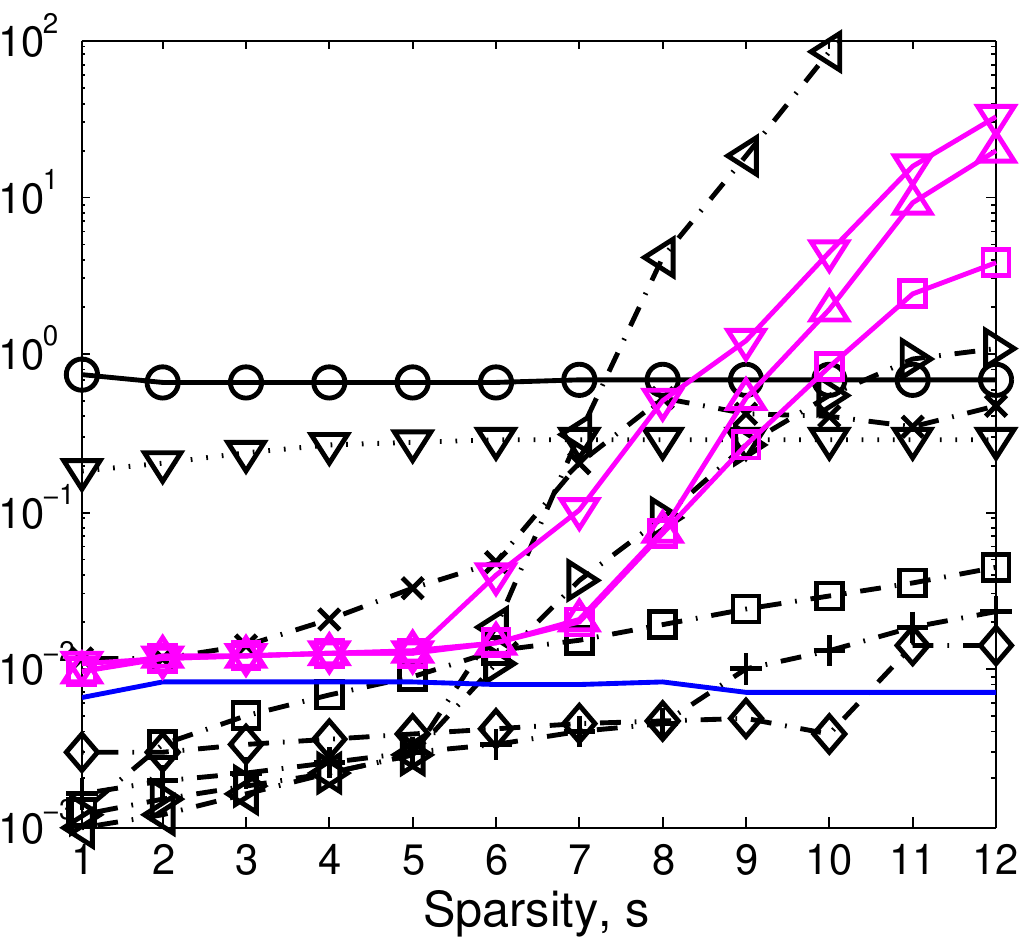}}
\subfigure[$\mathbb{P}(\left \| \Phi \hat x - \Phi x_0 \right \| \leq \left \| w \right \|)$]{\includegraphics[width=5.0cm, height=5.0cm]{cpxstgaussian_5db_m20n100_i}}
\subfigure[$\mathbb{E}(\left \| x_0 - \hat x\right \| /\left \| x_0 \right \| )$]{\includegraphics[width=5.0cm, height=5.0cm]{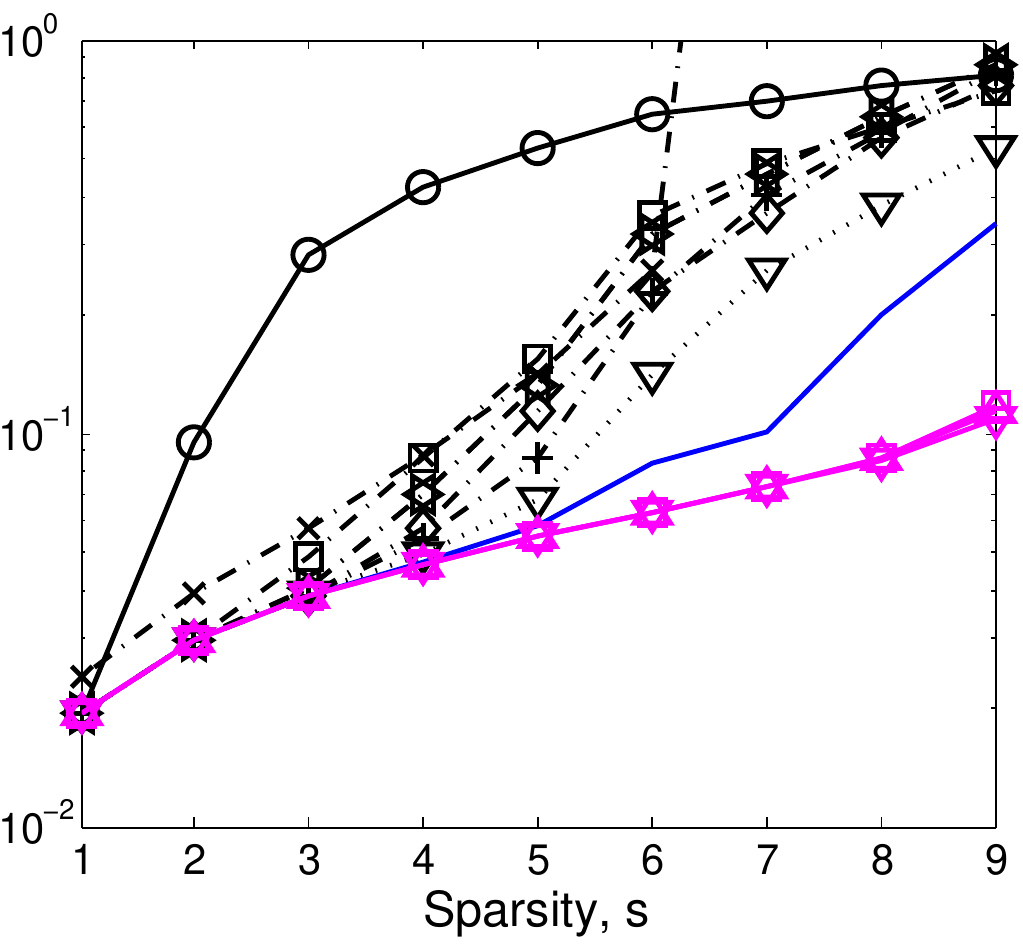}}
\subfigure[Execution time (seconds)]{\includegraphics[width=5.0cm, height=5.0cm]{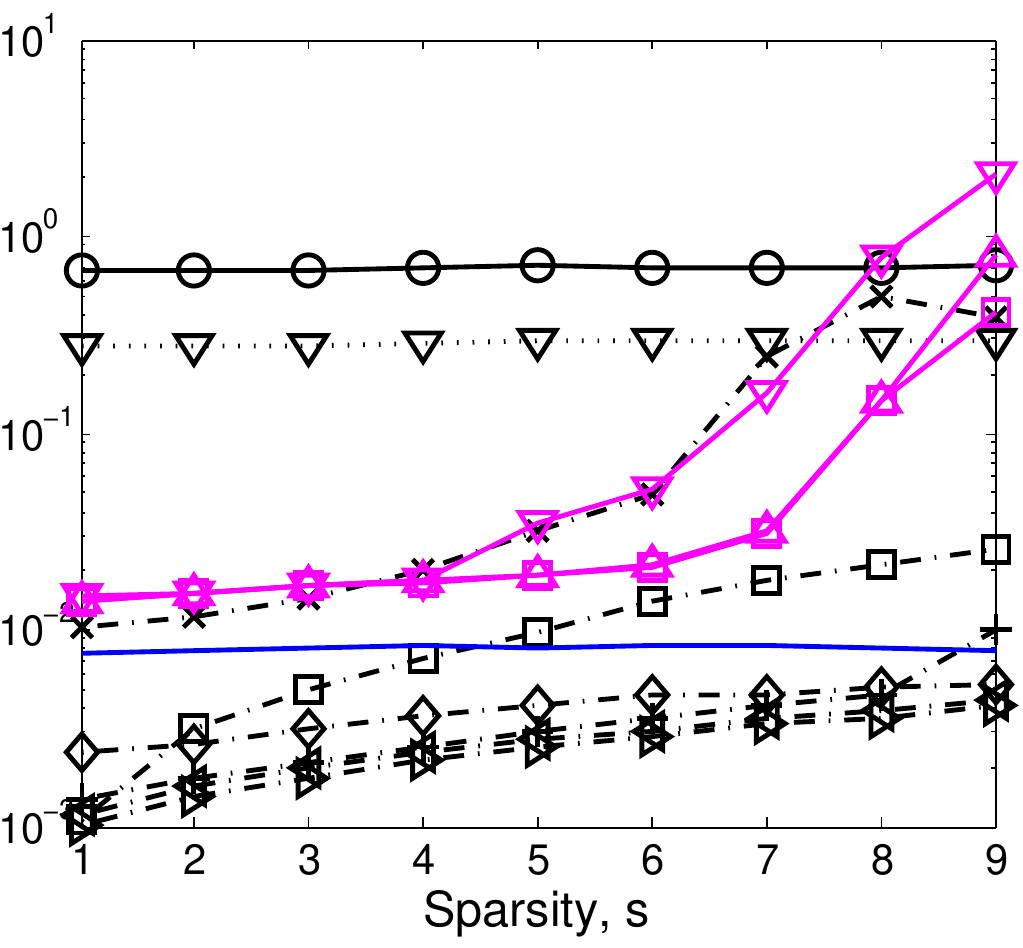}}
\caption{Performance comparison for complex-valued signal recovery when the sensing matrix $\Phi$ is set to the complex Gaussian matrix ((a)-(c): noiseless case,  (d)-(f): noisy case (SNR = $25$ dB),  (g)-(i): noisy case (SNR = $5$ dB)) }
\label{cpxfig}
\end{center}
\end{figure}

\begin{figure}
\begin{center}
\subfigure[$\mathbb{P}( \hat x  = x_0 )$]{\includegraphics[width=5.0cm, height=5.0cm]{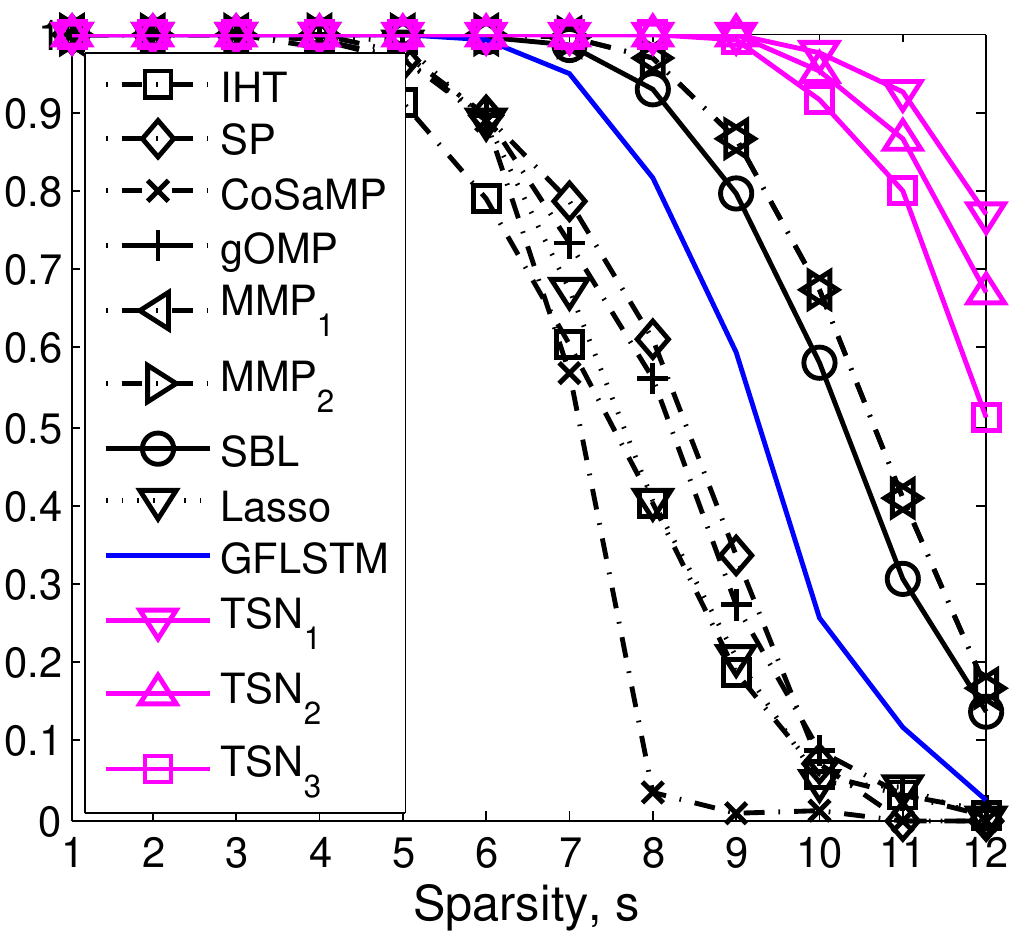}}
\subfigure[$\mathbb{E}(\left \| x_0 - \hat x\right \| /\left \| x_0 \right \| )$]{\includegraphics[width=5.0cm, height=4.81cm]{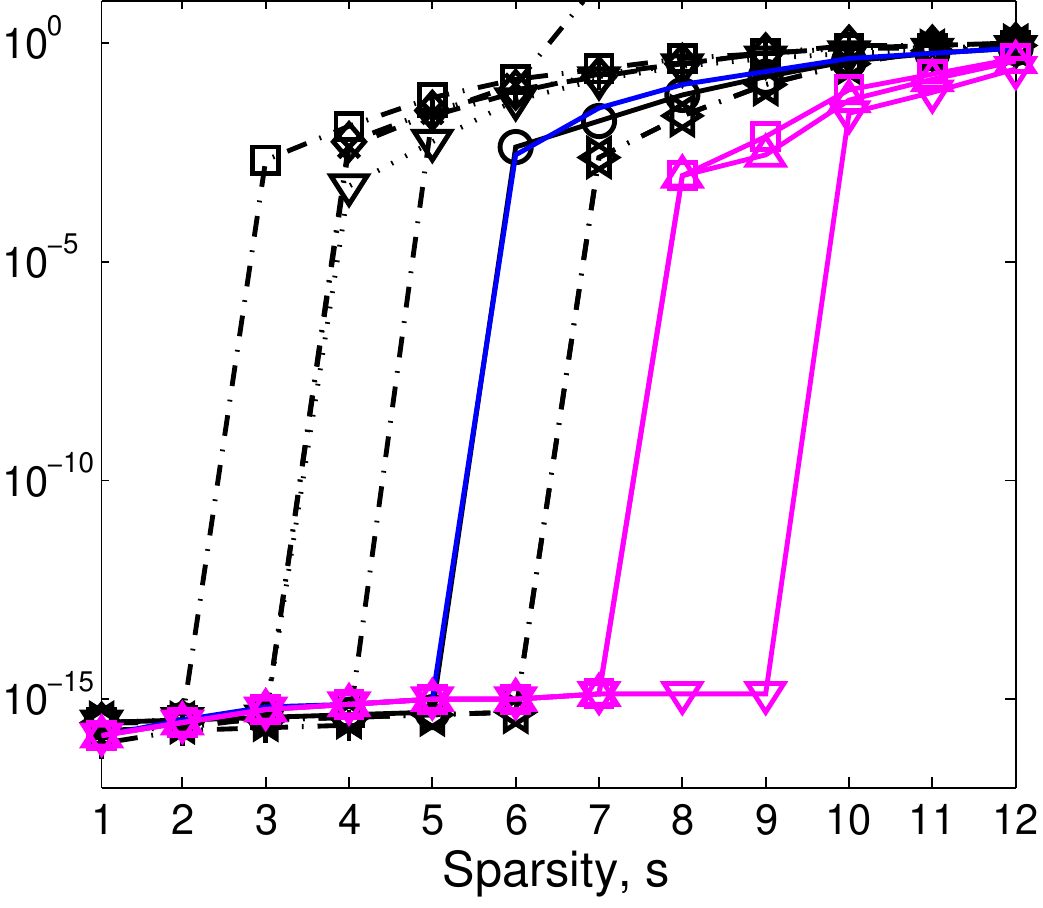}}
\subfigure[Execution time (seconds)]{\includegraphics[width=5.0cm, height=5.0cm]{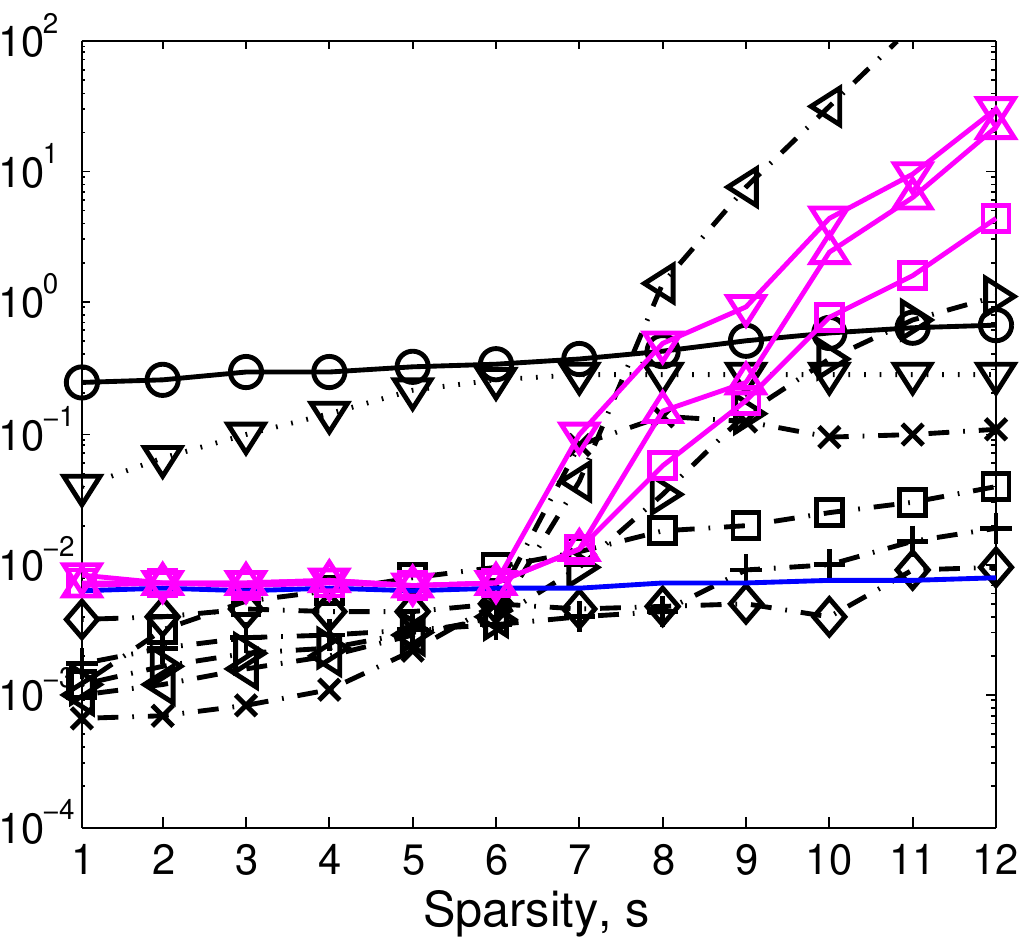}}
\subfigure[$\mathbb{P}(\left \| \Phi \hat x - \Phi x_0 \right \| \leq \left \| w \right \|)$]{\includegraphics[width=5.0cm, height=5.0cm]{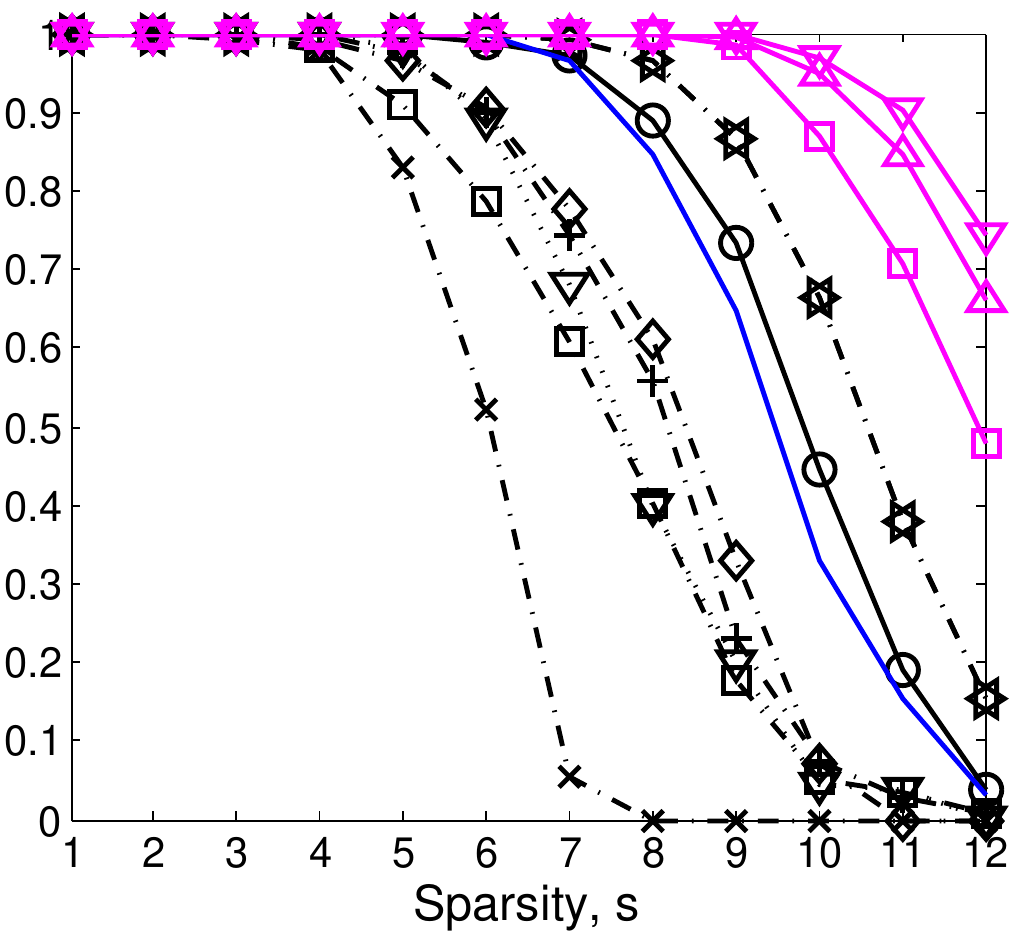}}
\subfigure[$\mathbb{E}(\left \| x_0 - \hat x\right \| /\left \| x_0 \right \| )$]{\includegraphics[width=5.0cm, height=4.83cm]{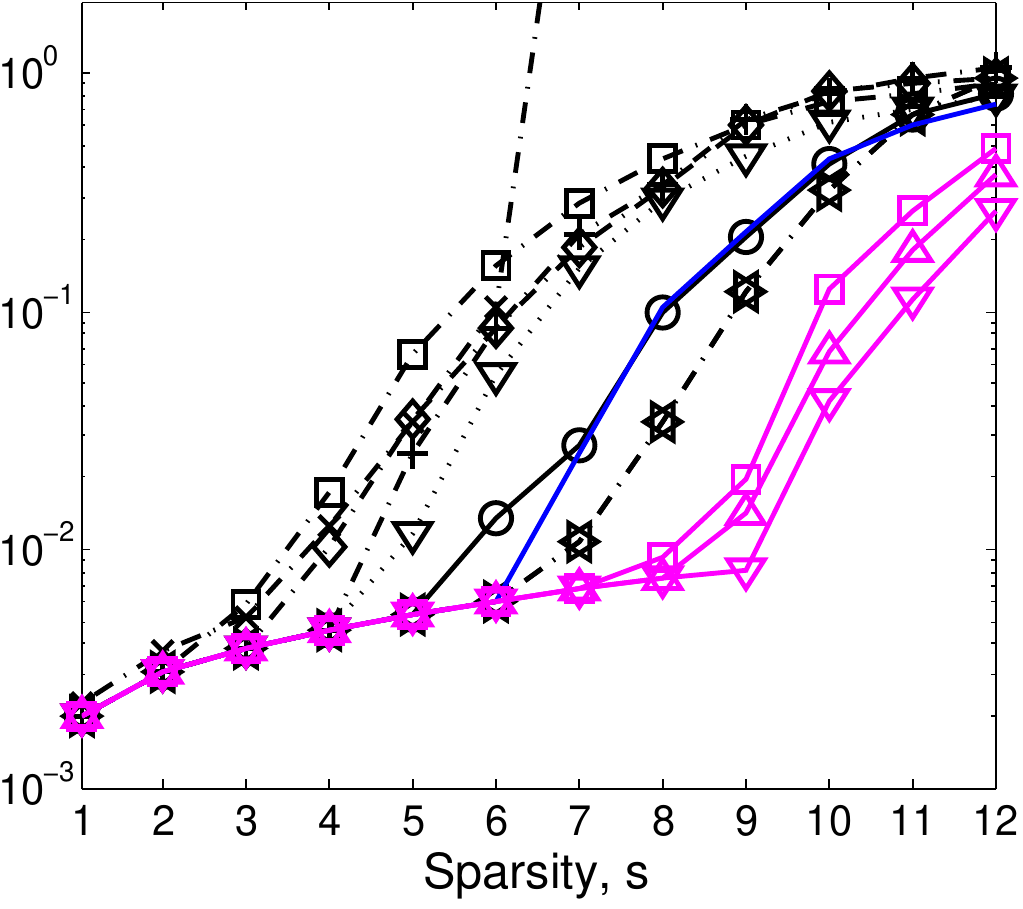}}
\subfigure[Execution time (seconds)]{\includegraphics[width=5.0cm, height=5.0cm]{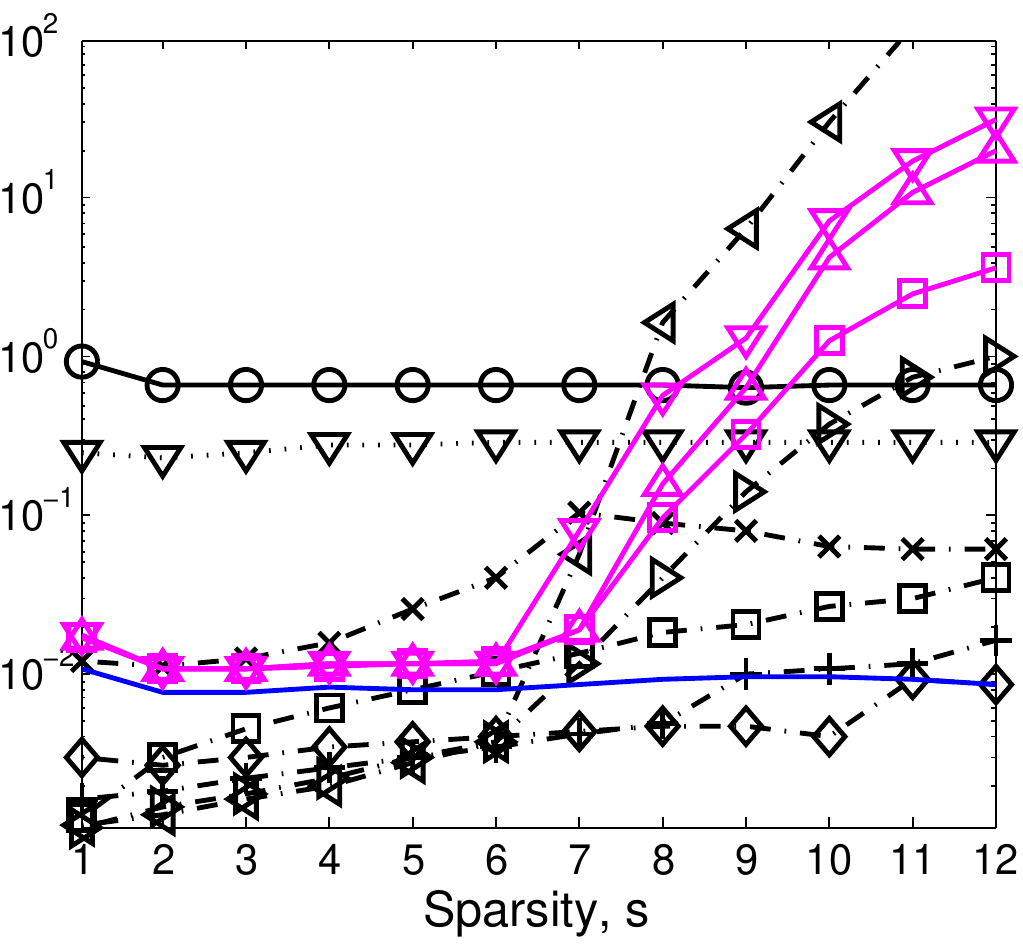}}
\subfigure[$\mathbb{P}(\left \| \Phi \hat x - \Phi x_0 \right \| \leq \left \| w \right \|)$]{\includegraphics[width=5.0cm, height=5.0cm]{cpxfourier_5db_m20n100_i}}
\subfigure[$\mathbb{E}(\left \| x_0 - \hat x\right \| /\left \| x_0 \right \| )$]{\includegraphics[width=5.0cm, height=4.81cm]{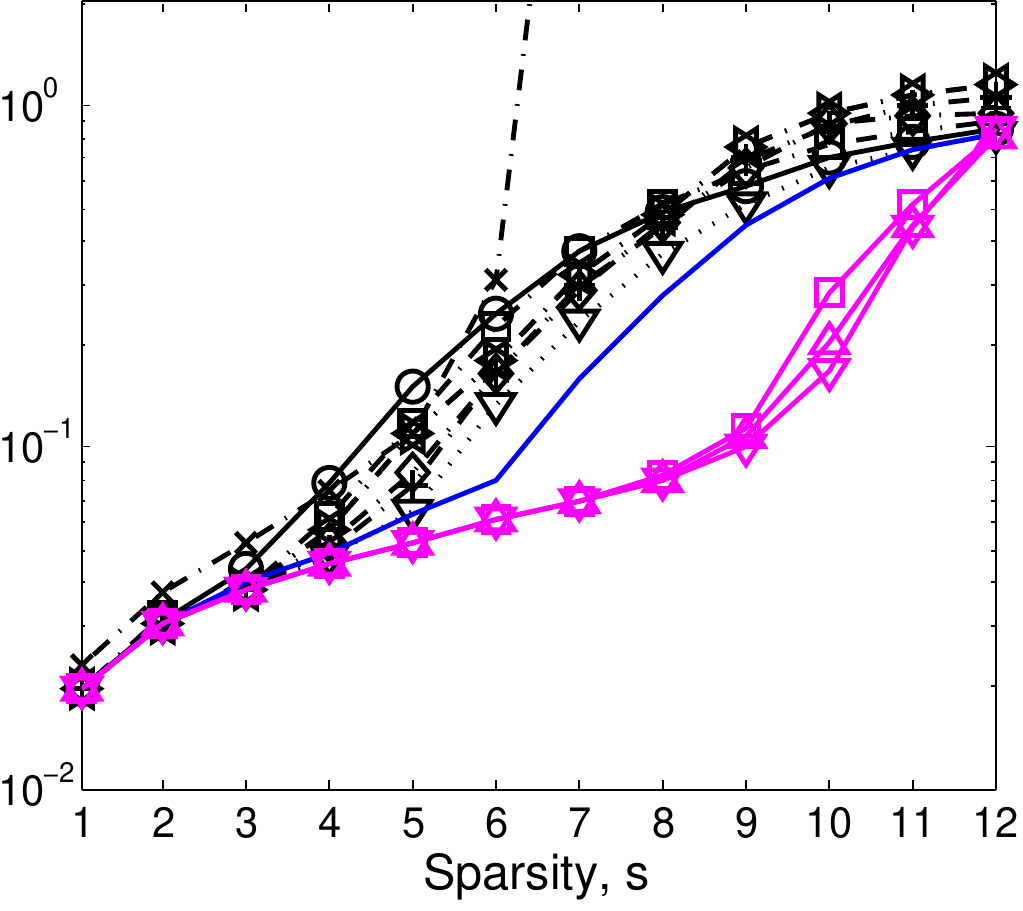}}
\subfigure[Execution time (seconds)]{\includegraphics[width=5.0cm, height=5.0cm]{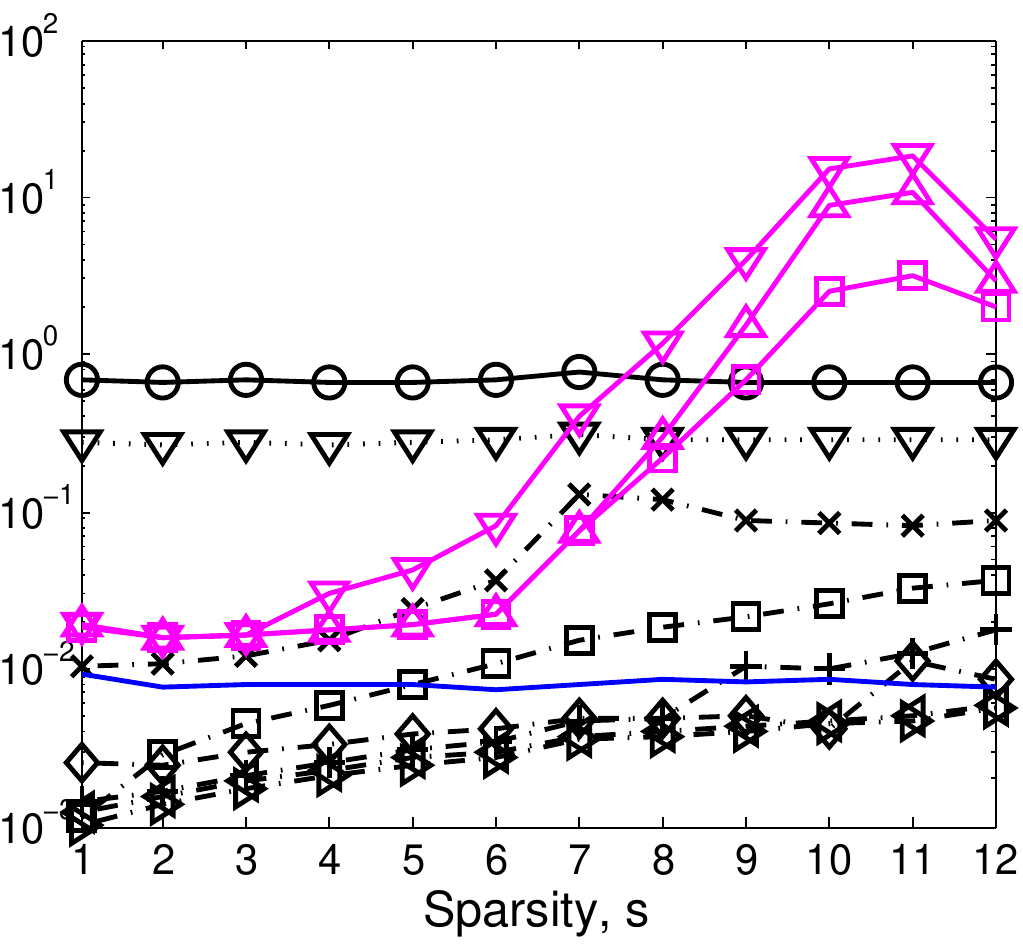}}
\caption{Performance comparison for complex-valued signal recovery when the sensing matrix $\Phi$ is set to the partial DFT matrix ((a)-(c): noiseless case,  (d)-(f): noisy case (SNR = $25$ dB),  (g)-(i): noisy case (SNR = $5$ dB)) }
\label{dftfig}
\end{center}
\end{figure}

\begin{figure}
\begin{center}
\subfigure[$\mathbb{P}( \hat x  = x_0 )$]{\includegraphics[width=5.0cm, height=5.0cm]{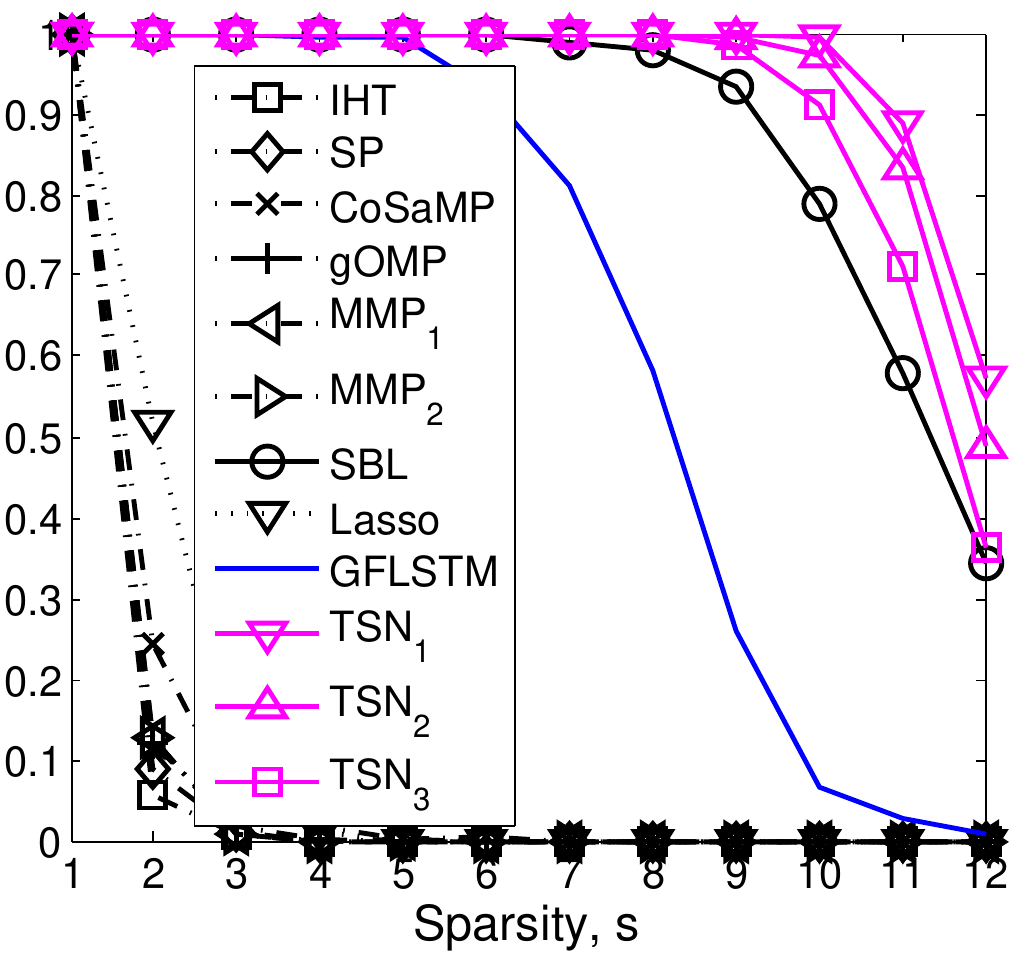}}
\subfigure[$\mathbb{E}(\left \| x_0 - \hat x\right \| /\left \| x_0 \right \| )$]{\includegraphics[width=5.0cm, height=4.83cm]{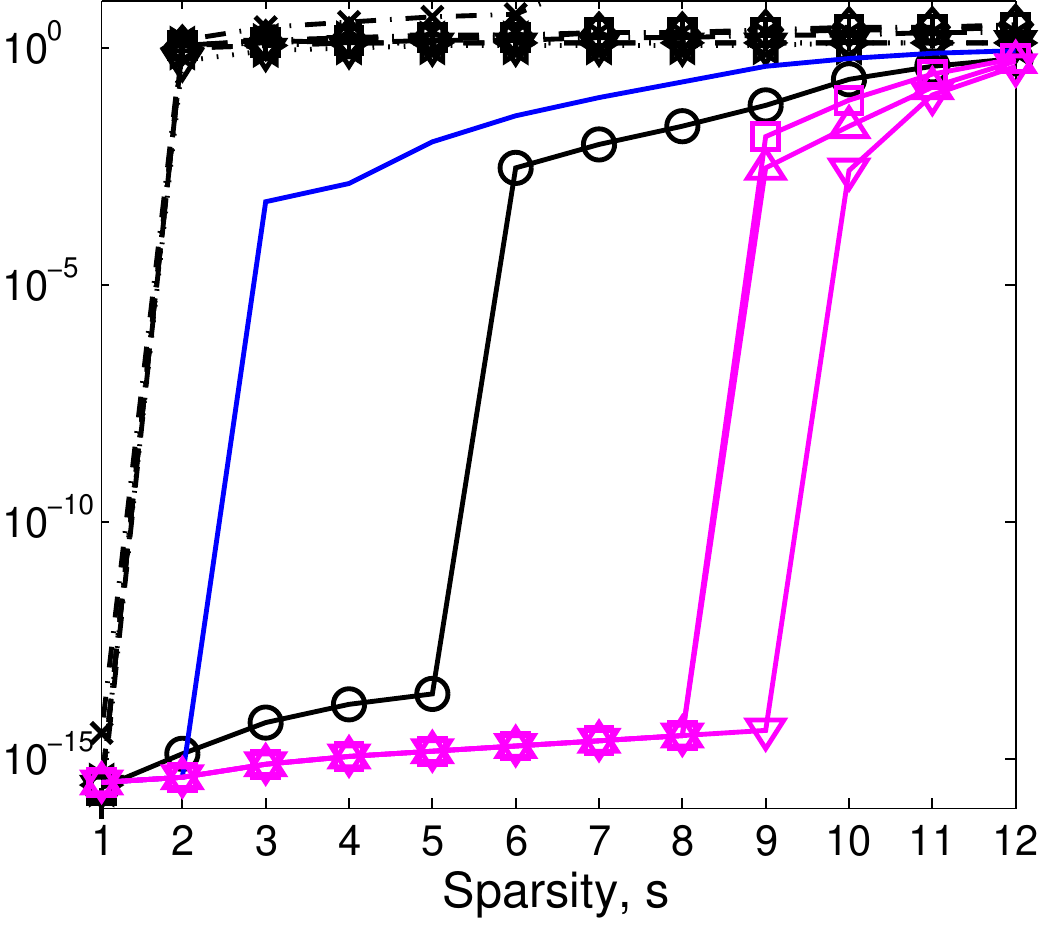}}
\subfigure[Execution time (seconds)]{\includegraphics[width=5.0cm, height=5.0cm]{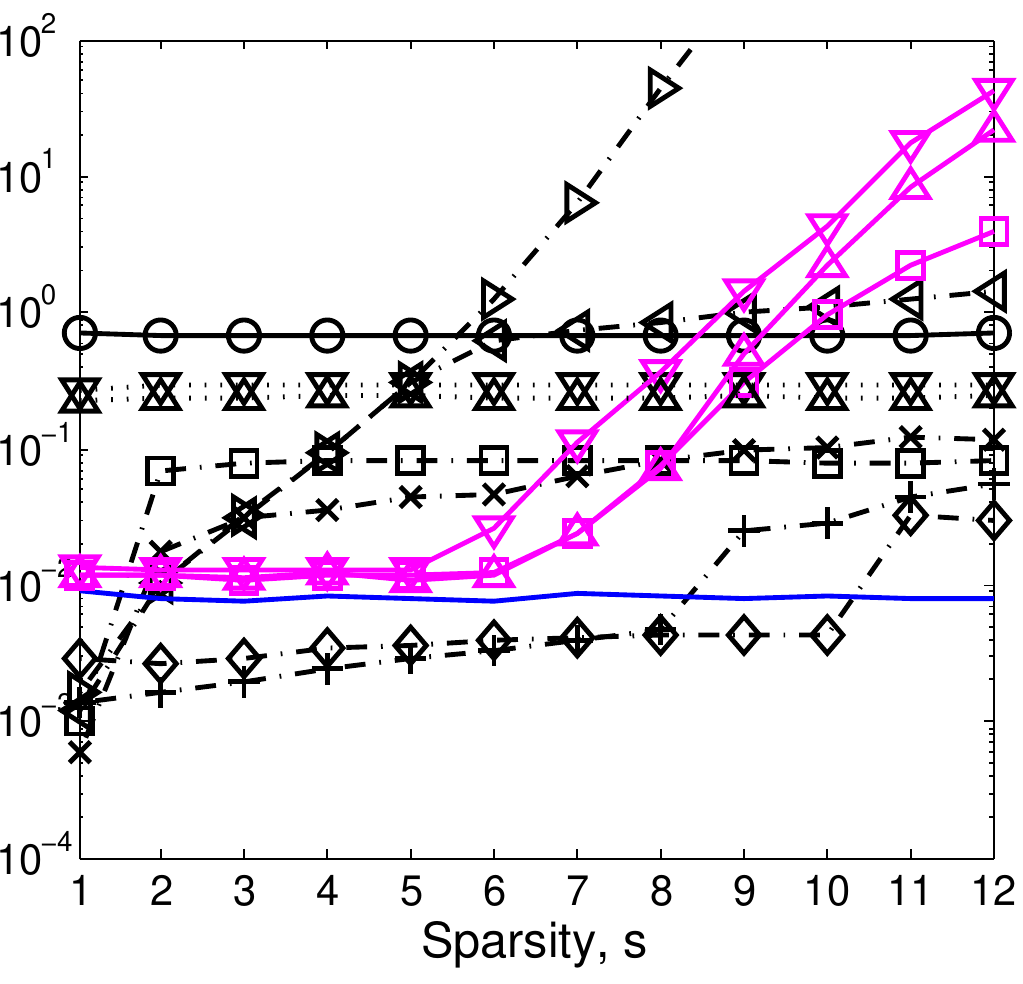}}
\subfigure[$\mathbb{P}(\left \| \Phi \hat x - \Phi x_0 \right \| \leq \left \| w \right \|)$]{\includegraphics[width=5.0cm, height=5.0cm]{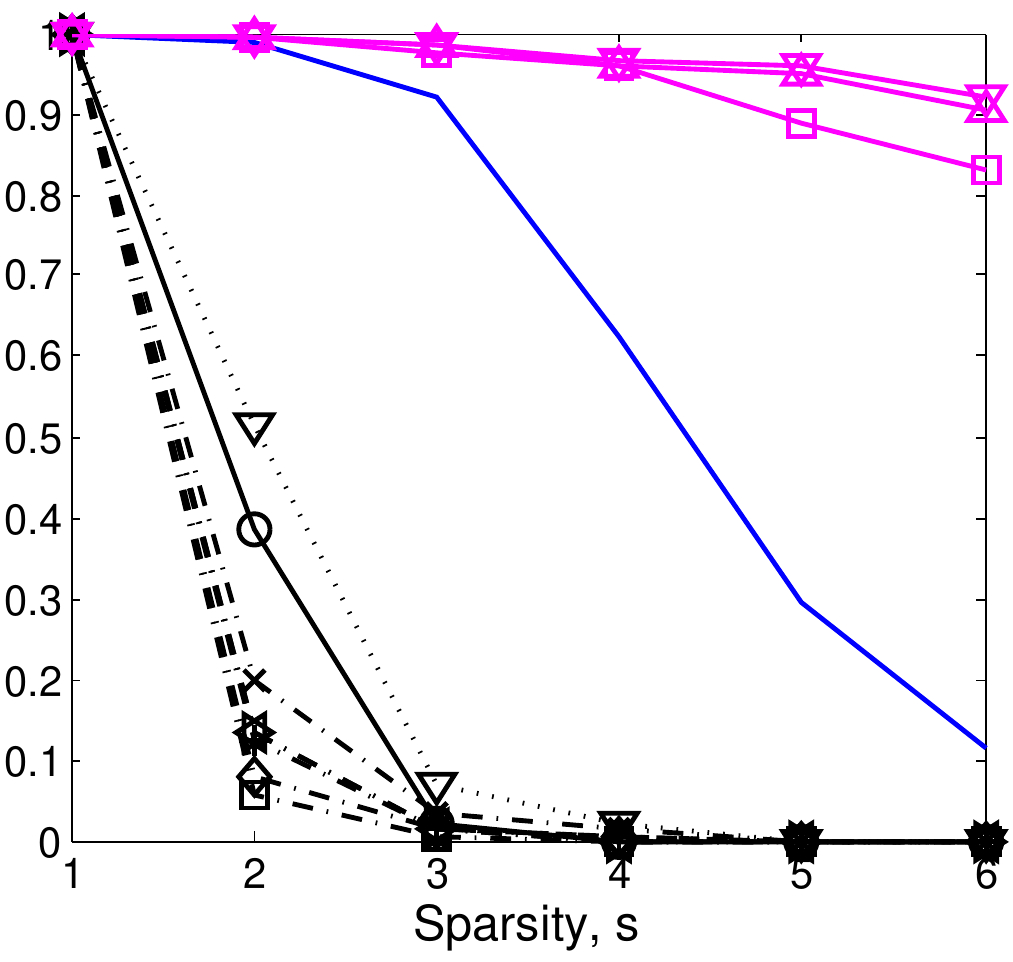}}
\subfigure[$\mathbb{E}(\left \| x_0 - \hat x\right \| /\left \| x_0 \right \| )$]{\includegraphics[width=5.0cm, height=5.0cm]{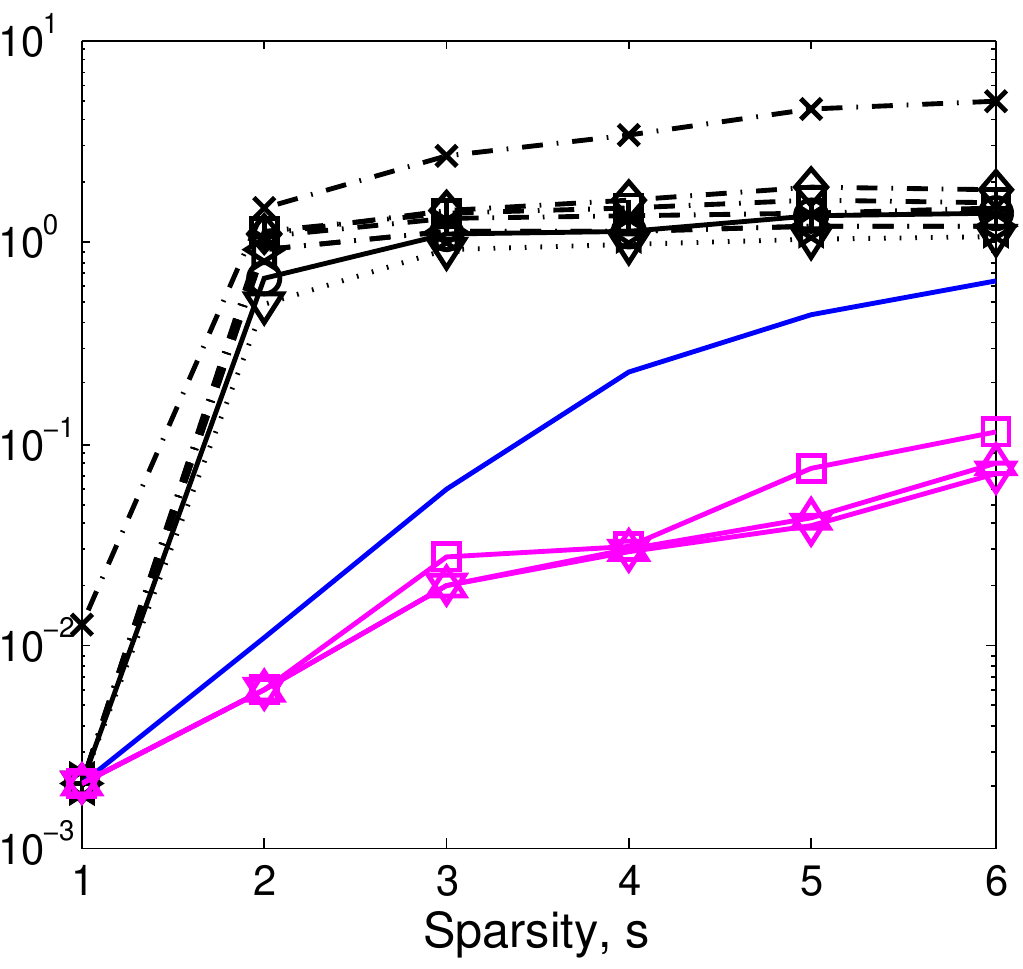}}
\subfigure[Execution time (seconds)]{\includegraphics[width=5.0cm, height=5.0cm]{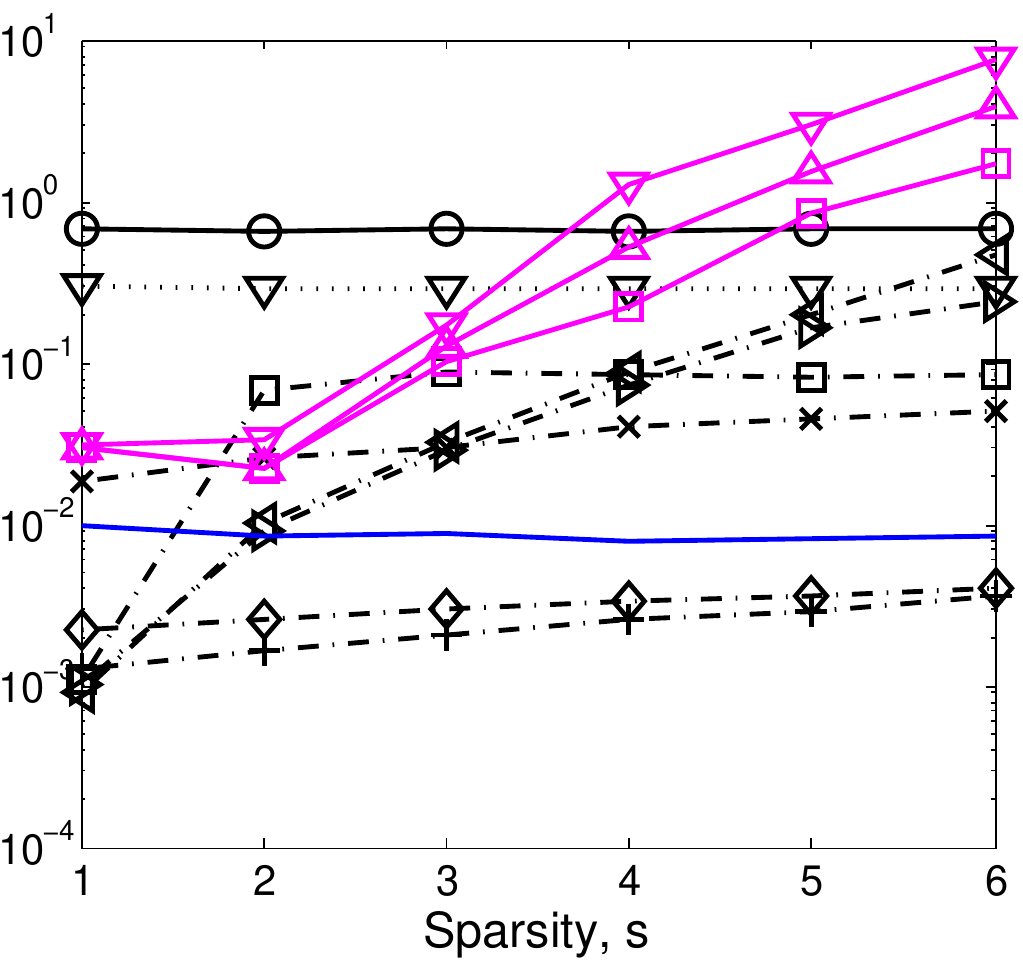}}
\subfigure[$\mathbb{P}(\left \| \Phi \hat x - \Phi x_0 \right \| \leq \left \| w \right \|)$]{\includegraphics[width=5.0cm, height=5.0cm]{corcpxstgaussian_5db_m20n100_i}}
\subfigure[$\mathbb{E}(\left \| x_0 - \hat x\right \| /\left \| x_0 \right \| )$]{\includegraphics[width=5.0cm, height=5.0cm]{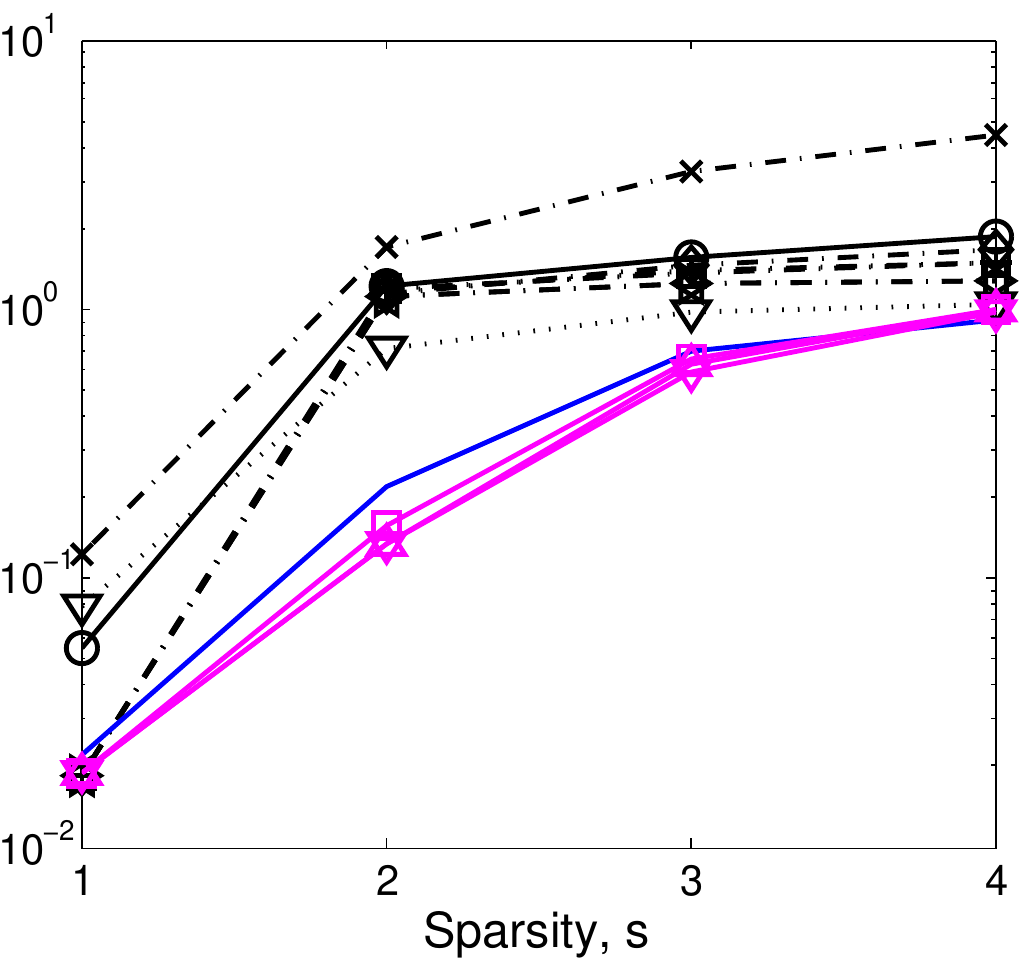}}
\subfigure[Execution time (seconds)]{\includegraphics[width=5.0cm, height=5.0cm]{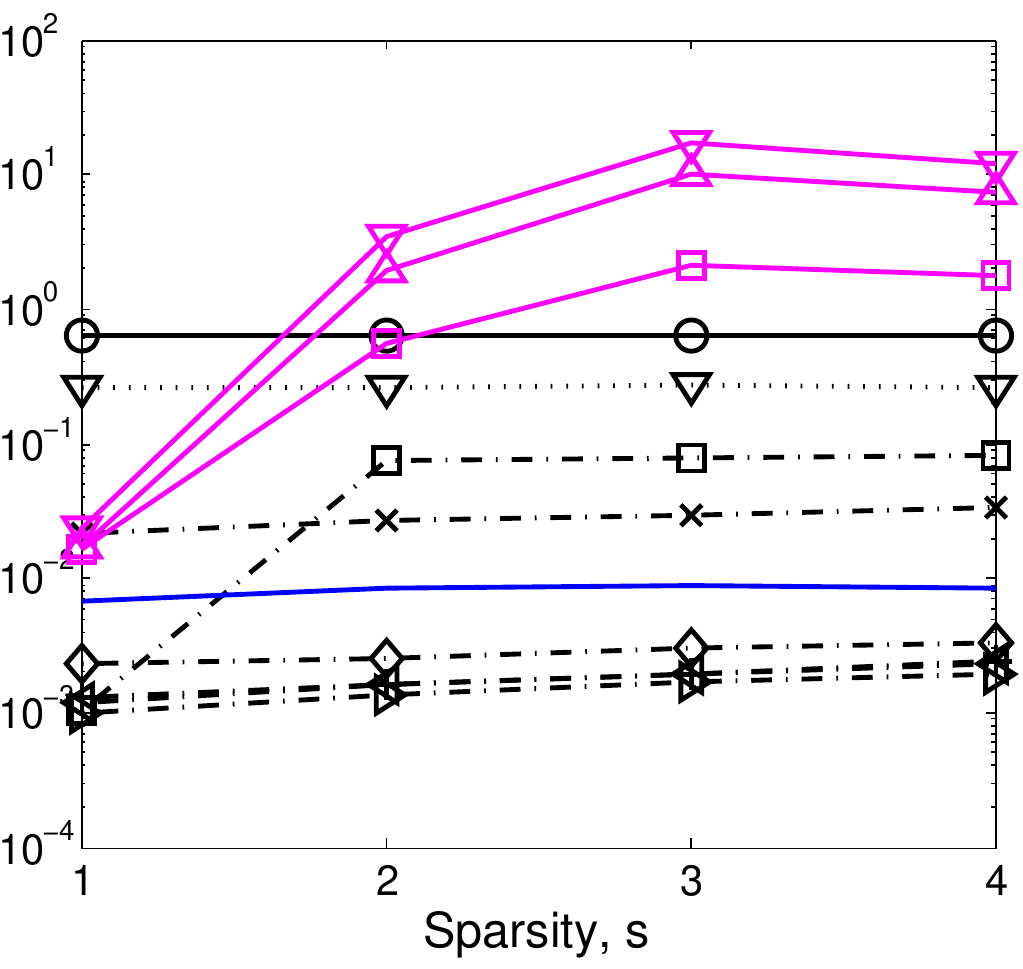}}
\caption{Performance comparison for complex-valued signal recovery when the sensing matrix $\Phi$ is set to the matrix with correlated columns ((a)-(c): noiseless case,  (d)-(f): noisy case (SNR = $25$ dB),  (g)-(i): noisy case (SNR = $5$ dB)) }
\label{corcpxfig}
\end{center}
\end{figure}

To demonstrate the superiority of TSN over other algorithms for the case of using various types of the sensing matrix $ \Phi$, we have set $ \Phi $ in one of the following three matrices: a complex Gaussian matrix, a partial DFT matrix, and a complex matrix with highly correlated columns \cite{he2017bayesian,xin2016maximal}. These matrices are generated according to the following rules and the columns of $\Phi$ are $l_2$-normalized.
\begin{itemize}
\item (Complex Gaussian matrix)  Each element of $\Phi \in \mathbb{C}^{m \times n}$ follows $\mathcal{CN}(0,1)$.
\item (Partial DFT matrix) $m$ rows are randomly selected from the DFT matrix of size $n \times n$.
\item (Matrix with correlated columns) $\Phi = \sum_{z=1}^{n}[ \frac{1}{z^2} p^{1}_z (q^{1}_z)^{\top} + \sqrt{-1}\cdot (\frac{1}{z^2} p^{2}_z (q^{2}_z)^{\top})]$ where $\sqrt{-1}$ is the imaginary unit and each element in $p^{i}_z \in \mathbb{R}^m$ ($q^{i}_z \in \mathbb{R}^n$) is drawn independently from a standard Gaussian distribution $\mathcal{N}(0,1)$ for $i \in \{1:2\}$ and $z \in \{1:n\}$.
\end{itemize}
Figures \ref{cpxfig}, \ref{dftfig}, and \ref{corcpxfig} show the performance comparison of each algorithm when $ \Phi $ is set to the complex Gaussian matrix, the partial DFT matrix, and the complex matrix with correlated columns, respectively.
It is observed from the results that the performance of SR algorithms shows the same trend as that in Figure \ref{realfig} in Section \ref{ne}; TSN shows better performance than existing SR algorithms in both noiseless and noisy cases. This indicates that TSN can verify to be applied to both real or complex signal restoration. In particular, it is observed in Figures \ref{cpxfig}(a), \ref{dftfig}(a), and \ref{corcpxfig}(a) that TSN almost achieves the ideal limit of sparsity $(s=10)$ for the uniform recovery  of $x_0$ with execution time of about 1 second, regardless of the sensing matrix type.

\begin{figure}

\begin{center}
\subfigure[$\mathbb{P}( \hat x  = x_0 )$]{\includegraphics[width=5.0cm, height=4.0cm]{realstgaussian_pos_noiseless_m20n100_s}}
\subfigure[$\mathbb{E}(\left \| x_0 - \hat x\right \| /\left \| x_0 \right \| )$]{\includegraphics[width=5.0cm, height=3.87cm]{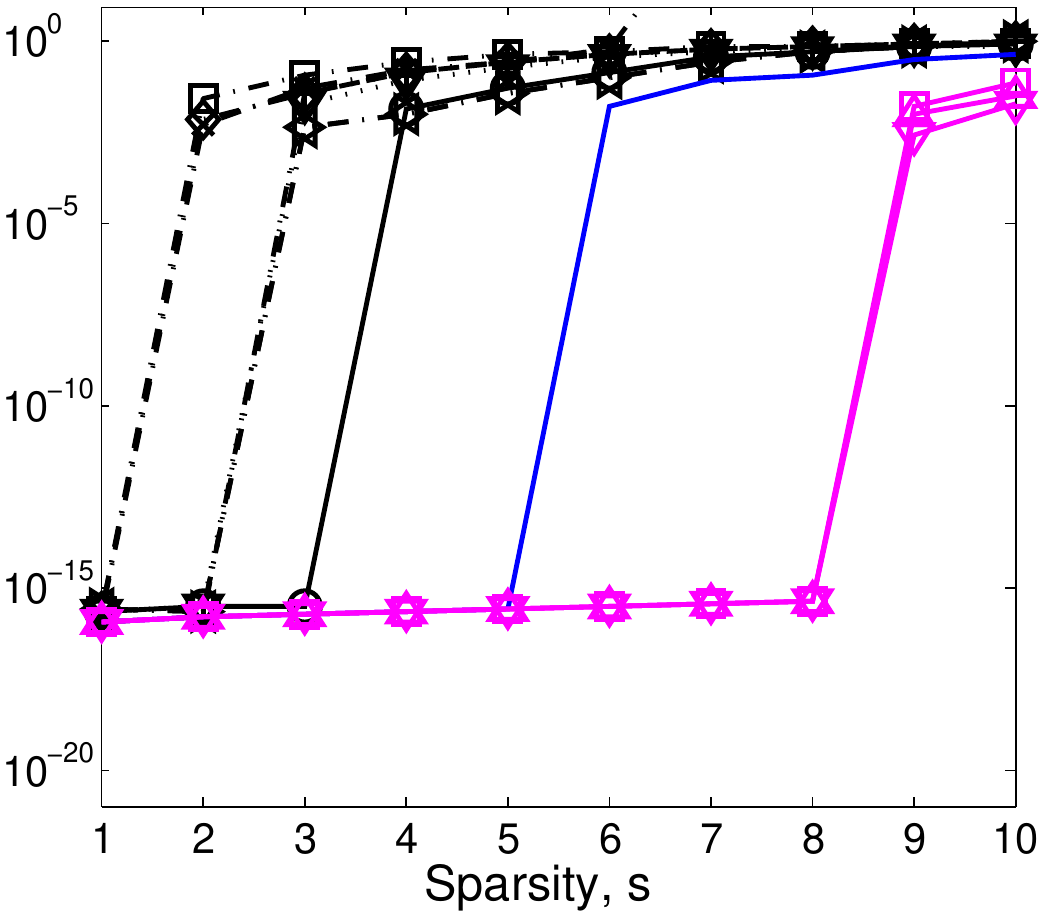}}
\subfigure[Execution time (seconds)]{\includegraphics[width=5.0cm, height=4.0cm]{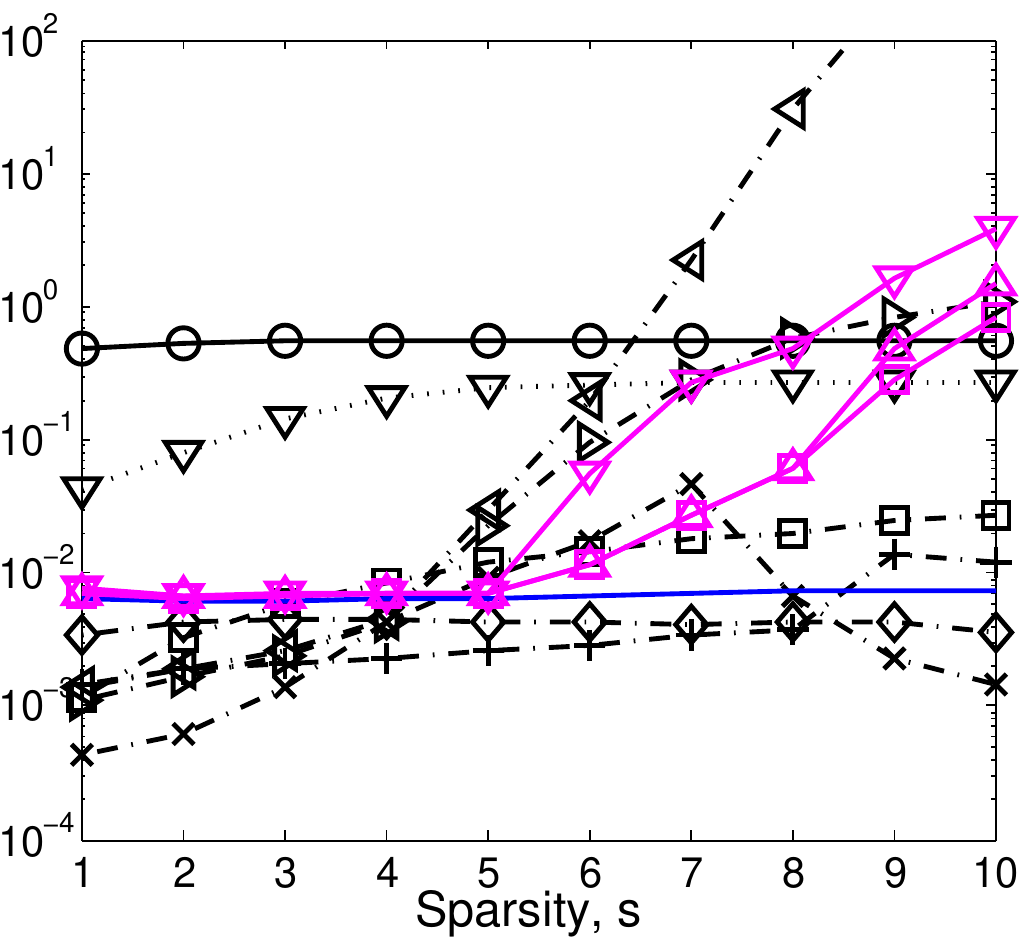}}

\subfigure[$\mathbb{P}( \hat x  = x_0 )$]{\includegraphics[width=5.0cm, height=4.0cm]{cpxstgaussian_pos_noiseless_m20n100_s}}
\subfigure[$\mathbb{E}(\left \| x_0 - \hat x\right \| /\left \| x_0 \right \| )$]{\includegraphics[width=5.0cm, height=4.0cm]{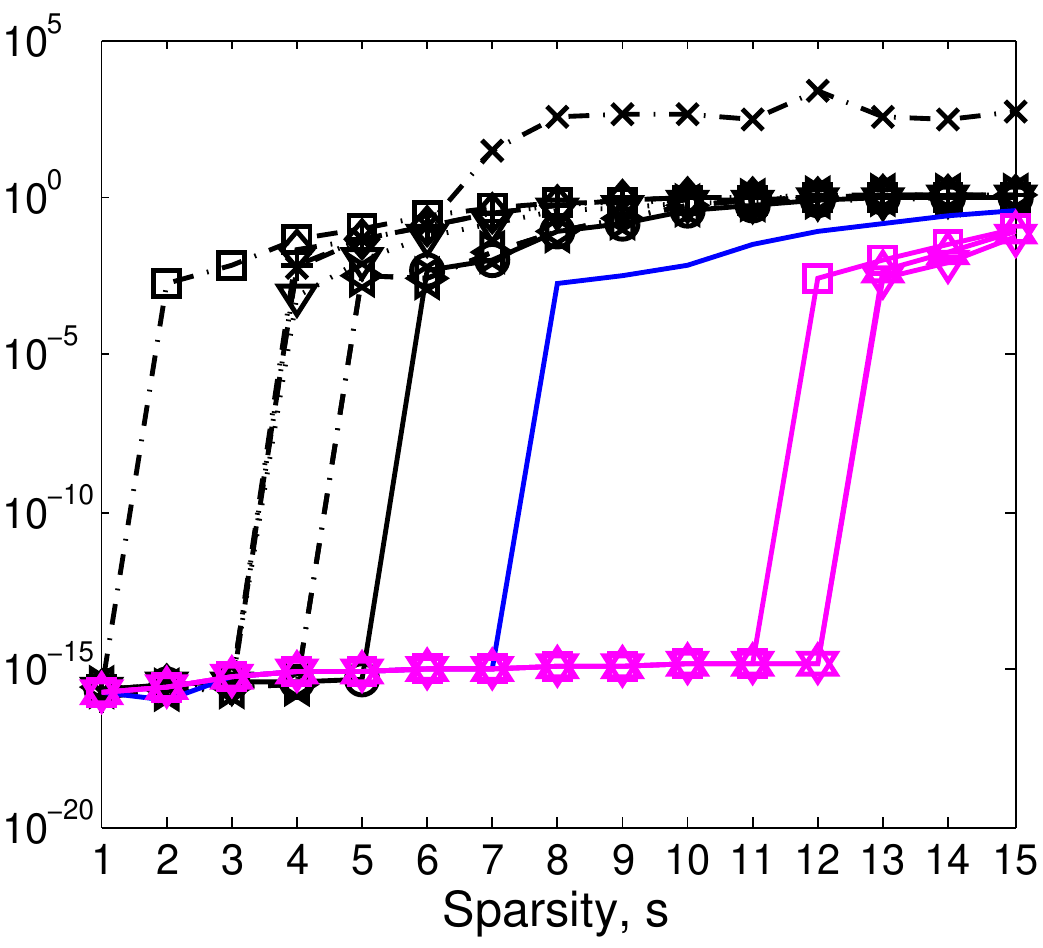}}
\subfigure[Execution time (seconds)]{\includegraphics[width=5.0cm, height=4.0cm]{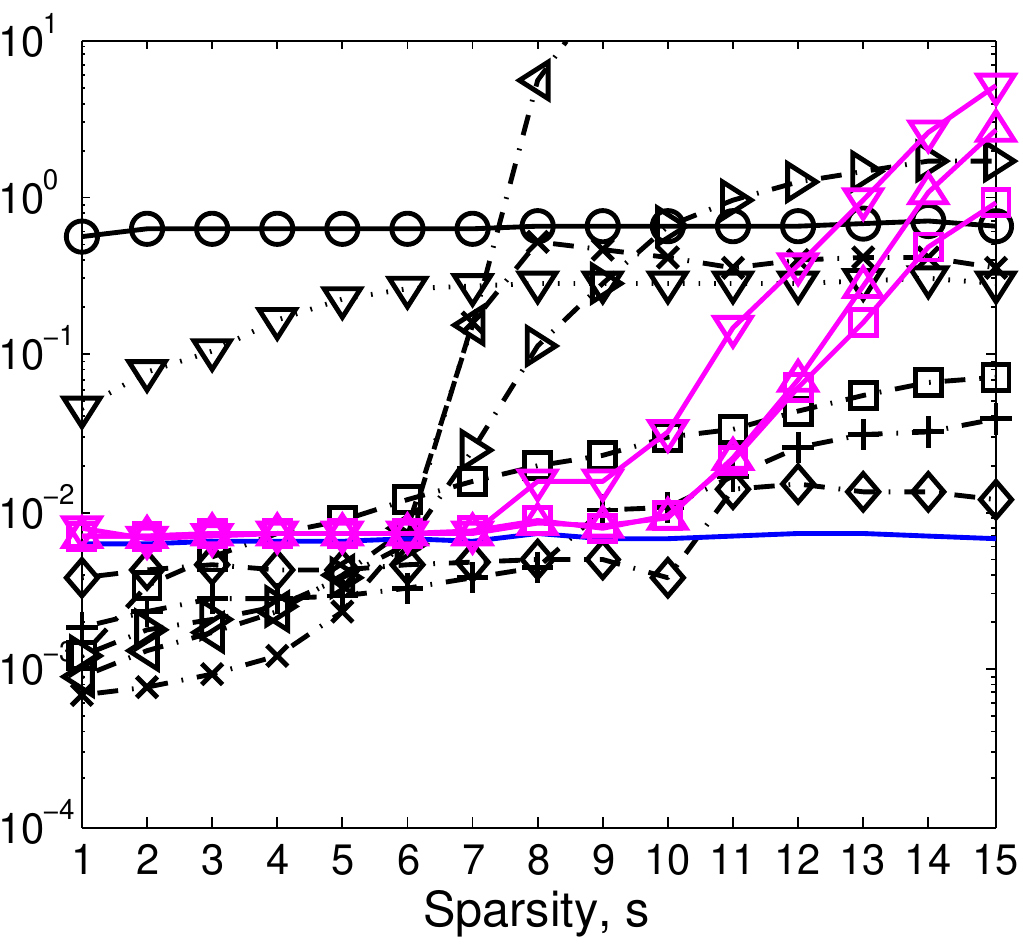}}

\subfigure[$\mathbb{P}( \hat x  = x_0 )$]{\includegraphics[width=5.0cm, height=4.0cm]{cpxfourier_pos_noiseless_m20n100_s}}
\subfigure[$\mathbb{E}(\left \| x_0 - \hat x\right \| /\left \| x_0 \right \| )$]{\includegraphics[width=5.0cm, height=3.87cm]{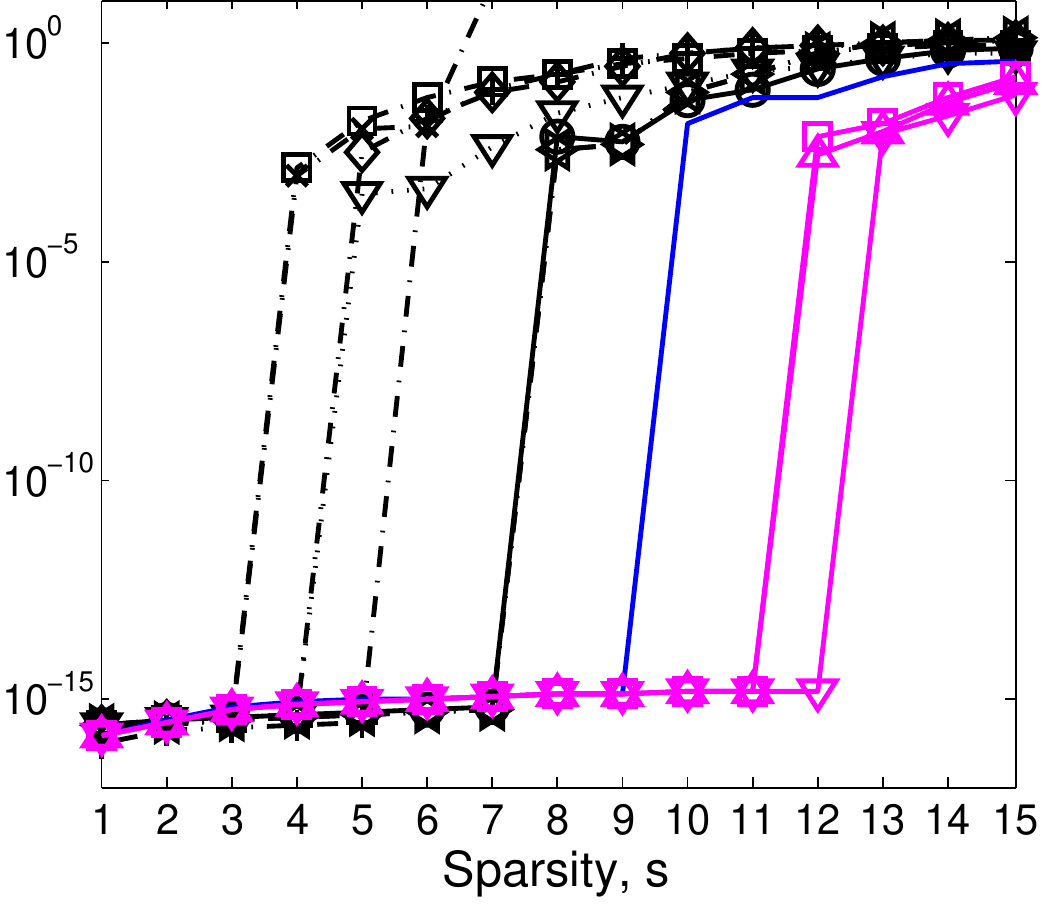}}
\subfigure[Execution time (seconds)]{\includegraphics[width=5.0cm, height=4.0cm]{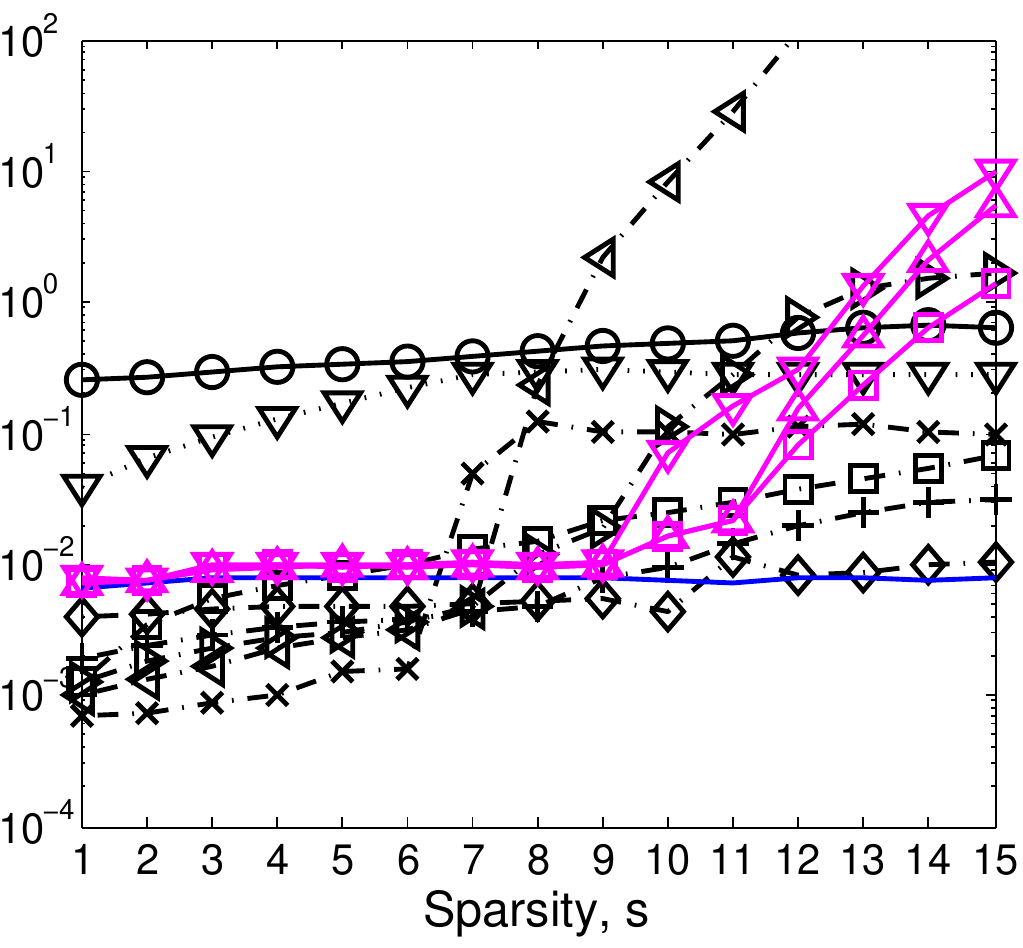}}

\subfigure[$\mathbb{P}( \hat x  = x_0 )$]{\includegraphics[width=5.0cm, height=4.0cm]{corcpxstgaussian_pos_noiseless_m20n100_s}}
\subfigure[$\mathbb{E}(\left \| x_0 - \hat x\right \| /\left \| x_0 \right \| )$]{\includegraphics[width=5.0cm, height=4.0cm]{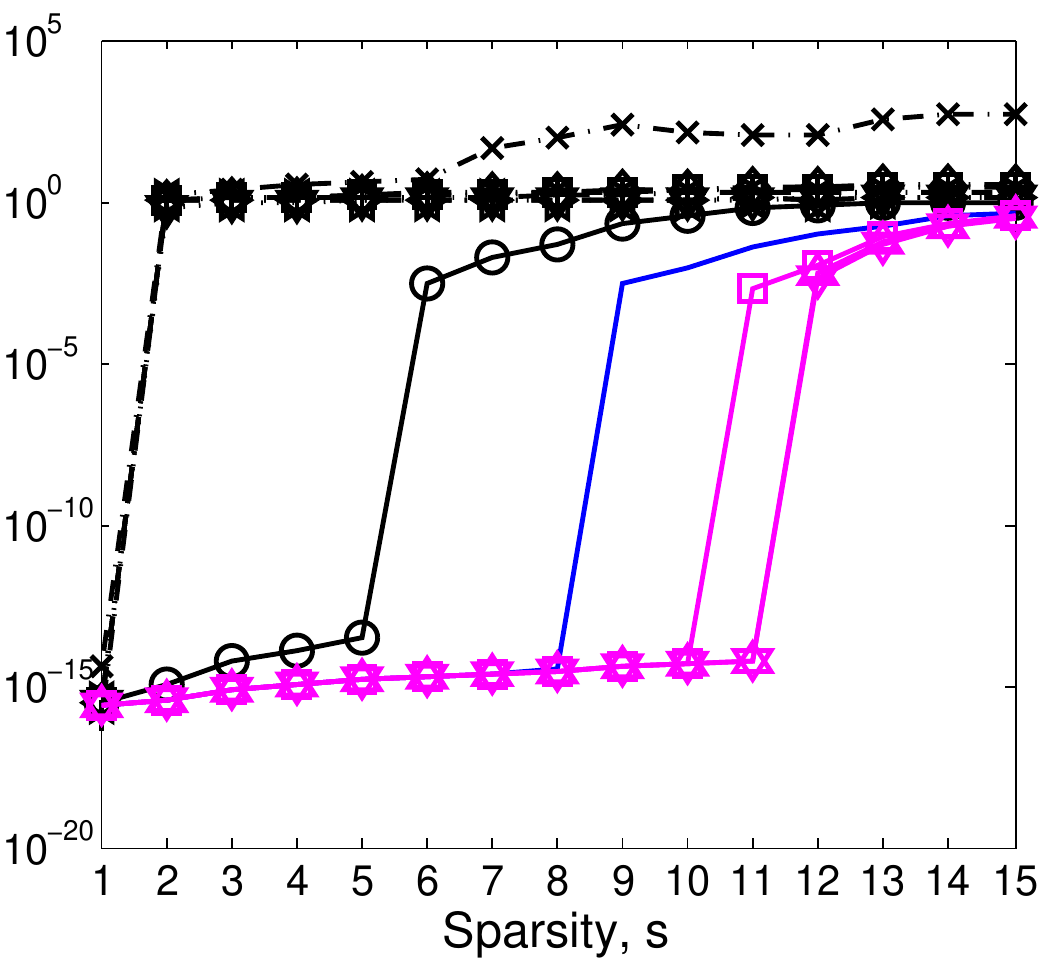}}
\subfigure[Execution time (seconds)]{\includegraphics[width=5.0cm, height=4.0cm]{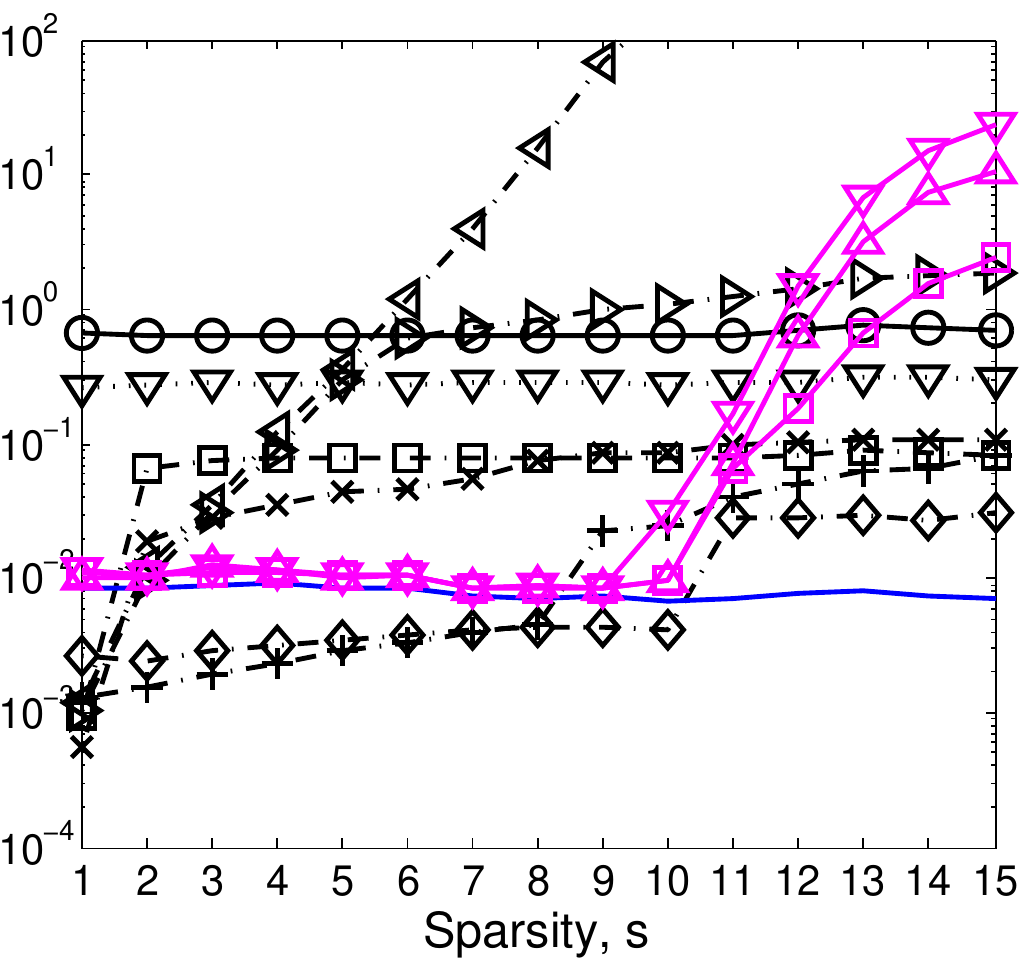}}

\caption{Performance comparison to recover a non-negative (structured) signal $x_0$ in the noiseless case ((a)-(c): the real-valued Gaussian sensing matrix,  (d)-(f): the complex-valued Gaussian sensing matrix,  (g)-(i): the partial DFT sensing matrix,  (j)-(l): the sensing matrix with correlated columns)}
\label{posfig}
\end{center}
\end{figure}

Furthermore, TSN can be applied to recover structured sparse signals and show better performance than existing SR algorithms. To provide an example, we consider the case when a non-negative constraint is added to the signal distributions used to plot Figures \ref{realfig}, \ref{cpxfig}, \ref{dftfig}, and \ref{corcpxfig} such that the real and imaginary parts of nonzero elements in $x_0$ are uniformly and independently generated from $0.1$ to $1$. The sparsity $s$ of $x_0$ is varied from $0$ to $15$ and given to SR algorithms except for TSN whose input $k$ is set to $10$ in the real-valued case or $15$ in the complex-valued case. The other setting parameters are the same as those used to plot the figures. Figure \ref{posfig} shows the performance evaluation of each algorithm in this experimental environment. It is observed from Figure \ref{posfig} that TSN enhances  the performance of its target DNN-SR, i.e., GFLSTM, and significantly outperforms existing SR algorithms given real-valued or complex-valued measurements.  In particular, it is observed that TSN achieves the ideal limit of sparsity $(s=10)$ for the uniform recovery of $x_0$ with execution time smaller than $0.1$ second in the complex-valued case. It is also observed in the real-valued case (Figures \ref{posfig}(a)-(c)) that the signal recovery rate of TSN is larger than 90 percent with running time about 1 second when the sparsity $s$ is set to its ideal limit $10$.
Figures \ref{posfig}(b) and \ref{posfig}(e) show that the maximum sparsity $s$ for the uniform recovery of $x_0$ using TSN is larger than two times that using SBL when $\Phi$ is set to real-valued or complex-valued Gaussian matrix.

\section{Scalability}
\label{scal_apx}
We set the hidden unit size of GFLSTM to 2000, number $N_{\textup{max}}$ of path candidates in MMP to 20000, and input parameters $(q,\boldsymbol{l},\boldsymbol{g},t_{\textup{max}})$ of TSN, i.e., Algorithm \ref{alg5},  to (m, (2,1,2,1,2,1,2,1), (60,1,60,1,60,1,60,1), 10). We used Algorithm \ref{alg1} whose input $(k_1,k_2)$ is set to $(1,m/2)$ to train GFLSTM and used the trained GFLSTM as the DNN $f_{\theta}(\cdot)$ in TSN. For other parameters, we used the same settings shown in Figure \ref{realfig}(a). Then, we selected the four highest performing algorithms, i.e., TSN, MMP, SBL, and Lasso shown in Table \ref{table0} and performed an additional evaluation test for $m>40$ and $n=10^4$.\footnote{We only tested until $ (m,n) \leq(1500,10^4)$ due to insufficient memory for MMP.} The results under these settings are listed in Table \ref{table_apx_scal}, where TSN outperforms the other algorithms and has a lower complexity than MMP and SBL. Hence, TSN scales well at least for the case where $m$ is less than 1500. However, as $m$ increases, we observed that the ratio of performance versus complexity of TSN tends to decrease relative to other SR algorithms.

\begin{table}
\centering
\footnotesize
\caption{Maximum sparsity $s_{0.95}$ of each evaluated algorithm with exact support recovery rate above 95\% and execution time of each algorithm for sparsity $|\Omega|$ being $s_{0.95}$}
\begin{tabular}{|m{0.1cm}|m{0.1cm}|m{0.1cm}|m{0.1cm}|m{0.1cm}|m{0.1cm}|m{0.1cm}|m{0.1cm}|m{0.1cm}|m{0.1cm}|m{0.1cm}|m{0.1cm}|m{0.1cm}|}
  \hline
  \multicolumn{1}{|c|}{Alg $\setminus$ $(m/n)$} 
& \multicolumn{1}{c|}{$500/10^4$}  & \multicolumn{1}{c|}{$1000/10^4$} & \multicolumn{1}{c|}{$1500/10^4$} \\\hline 
   \multicolumn{1}{|c|}{TSN} 
& \multicolumn{1}{l|}{$\boldsymbol{90}\,(6.4)$}& \multicolumn{1}{l|}{$\boldsymbol{217}\,(18.9)$} & \multicolumn{1}{l|}{$\boldsymbol{351}\,(29.8)$}   \\\cline{1-4}   
   \multicolumn{1}{|c|}{SBL} 
& \multicolumn{1}{l|}{${18}\,(128.3)$}& \multicolumn{1}{l|}{$57\,(220)$} & \multicolumn{1}{l|}{${280}\,(373.5)$}   \\\cline{1-4}
   \multicolumn{1}{|c|}{MMP} 
& \multicolumn{1}{l|}{${75}\,(21.8)$}& \multicolumn{1}{l|}{$166\,(28.4)$} & \multicolumn{1}{l|}{${239}\,(453.4)$}   \\\cline{1-4}
   \multicolumn{1}{|c|}{Lasso}  
& \multicolumn{1}{l|}{$65\,(7.2)$}& \multicolumn{1}{l|}{$215\,(19.6)$} & \multicolumn{1}{l|}{${275}\,(24.3)$} \\\cline{1-4}
\end{tabular}
\label{table_apx_scal}
\end{table}

\section{Application: NOMA}
\label{app_noma}

\begin{algorithm}
  \caption{NOMA system}
\footnotesize
    \begin{algorithmic}[1]
    \Input{Set $\Omega \subseteq \{1:n\}$ of $s$ indices for truly active users, set $\Phi:=[\phi_1,...,\phi_n] \in \mathbb{C}^{m \times n}$ of spreading sequences (columns) assigned to each user in the cell}
    \State  $z_m \in \mathbb{Z}_2^{k \times B} \gets$ generate $s$ message blocks assigned to the corresponding $s$ active users, where $B$ is the block length. \Comment{Message block generation}
    \State  $z_e= \textup{Enc}(z_m) \in \mathbb{Z}_2^{k \times \lceil B /  r_c\rceil } \gets$ encode generated message block  $z_m$, where $\textup{Enc}(\cdot)$ is the encoder and $r_c$ is the coding rate. \Comment{Encoding}
    \State  $X:=[x_1,...,x_{\lceil B / (r_m \cdot r_c) \rceil}]= \textup{Mod}(z_e) \in \mathbb{C}^{k \times \lceil B / (r_m \cdot r_c) \rceil} \gets$ modulate encoded block  $z_e$, where $\textup{Mod}(\cdot)$ is the modulator and $r_m$ is the number of bits per symbol which the modulator retrieves. \Comment{Modulation}
 \State Obtain measurements $y_i = \Phi_{\Omega} x_i + w_i$ for $i \in \{1:\lceil B / (r_m \cdot r_c) \rceil \}$, where $\Omega$ is the support, i.e., a set of $s$ incides for the $s$ active users and $w_i$ is a noise vector. \Comment{AWGN channel}
    \State  Estimate $x_i$ and $\Omega$ as $\bar x_i$ and $\bar \Omega_i$, respectively, by using target SR algorithm given $y_i$ and $\Phi$ for $i \in \{1:\lceil B / (r_m \cdot r_c) \rceil \}$.
        \For{$i \in \{1:\lceil B / (r_m \cdot r_c) \rceil \}$}
        \State $E(u)\gets \sum\limits_{i \in \{1:\lceil B / (r_m \cdot r_c) \rceil}1( u \in \hat \Omega_i)$        \EndFor
    \State $\hat \Omega \gets $ Estimate the support by selecting the $s$-largest indices in $E$ \Comment{Support estimation}
    \State $\hat X = \underset{X }{\arg\min}{\left \| Y - \Phi_{\hat \Omega} X \right \|} \in \mathbb{C}^{k \times \lceil B / (r_m \cdot r_c) \rceil} \gets $  Recover the signals for active users as matrix $\hat X \in \mathbb{C}^{k \times \lceil B / (r_m \cdot r_c) \rceil}$

        \State $\hat z_e = \textup{Dmd}(\hat X) \in \mathbb{Z}_2^{k \times \lceil B/r_c \rceil }$, where $\textup{Dmd}(\cdot)$ denotes the demodulator  \Comment{Demodulation}
    \State  $\hat z_m = \textup{Dec}(\hat z_e) \in \mathbb{Z}_2^{k \times B}$, where $\textup{Dec}(\cdot)$ denotes the decoder  \Comment{Decoding}
    \State  {\bfseries return} {$\hat \Omega,\hat z_m $} \Comment{Return the estimated index set $\hat \Omega$ and message blocks $\hat z_m$ for $s$ active users}
  \end{algorithmic}
\label{noma_alg}
\end{algorithm}
NOMA has been widely adopted into upcoming 5G wireless networks to enhance spectral efficiency.
We propose Algorithm \ref{noma_alg} to detail NOMA. We set number $n$ of users in the cell and number $m$ of measurements to $100$ and $20$, respectively. The spreading sequence of each user corresponds to each column of $20 \times 100$ partial DFT matrix $\Phi$. In addition, we set number $s$ of active users in the cell to $10$.  Support set $\Omega$ of size $s$ was uniformly sampled from $\{1:n\}$, with each element representing an index for each of the $s$ active users. We randomly generated set $z_m$ of $s$ message blocks, each consisting of $4$ bits and assigned to each active user. We used Bose-–Chaudhuri-–Hocquenghem code with codeword length $7$ and message length $4$ for the encoding and decoding steps 2 and 12 at coding rate $r_c$ of $4/7$. Then, we applied quadrature phase shift keying for the modulation and demodulation steps 3 and 11, retrieving $r_m=2$ bits per symbol. In step 4, we used the AWGN channel by sampling each element of noise vector $w_i$ from a Gaussian distribution with zero mean and variance determined by the target SNR. 

The goal of NOMA is to recover user index set $\Omega$ and set $z_m$ of message blocks for the $s$ active users. In step 5, we run a target SR algorithm to obtain support candidate $\bar \Omega_i$ from each measurement vector $y_i$. Then, the final estimate $\hat \Omega$ for support $\Omega$ is obtained in step 9 by selecting the $s$ most frequent indices in candidates $(\bar \Omega_i)_{i=1}^{\lceil B / (r_m \cdot r_c) \rceil }$. From estimate $\hat \Omega$, signal matrix $X$ is estimated as $\hat X$ using least-squares regression in step 10. Finally, set $z_m$ of $s$ message blocks is reconstructed as $\hat z_m$ during demodulation and decoding steps 11 and 12 by using estimate $\hat X$. To evaluate the proposed TSN, we used TSN$_3$ to plot Figure \ref{dftfig}. The DNN, i.e., GFLSTM, used in TSN was trained as described in Section \ref{apenb} and the other setting parameters are the same as those used to plot Figure \ref{dftfig}.

\section{Application: image reconstruction}
\label{app_ir}

\begin{figure}
\begin{center}
\subfigure[\scriptsize Original image]{\includegraphics[width=3.4cm, height=3cm]{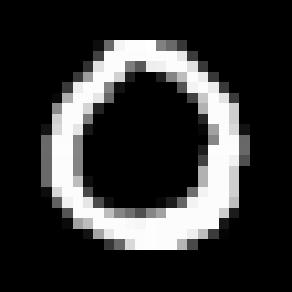}}
\subfigure[\scriptsize TSN (output)]{\includegraphics[width=3.4cm, height=3cm]{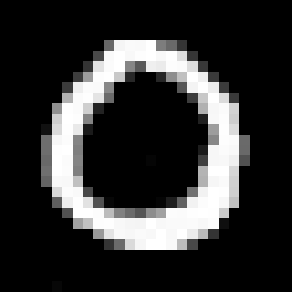}}
\subfigure[\scriptsize TSN (difference)]{\includegraphics[width=3.4cm, height=3cm]{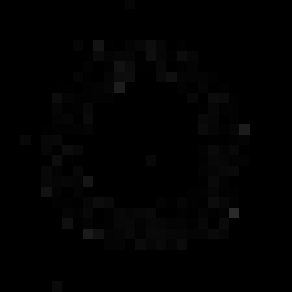}}
\subfigure[\scriptsize GFLSTM (output)]{\includegraphics[width=3.4cm, height=3cm]{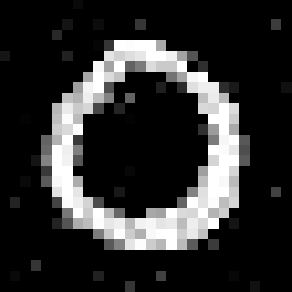}}
\subfigure[\scriptsize GFLSTM (difference)]{\includegraphics[width=3.4cm, height=3cm]{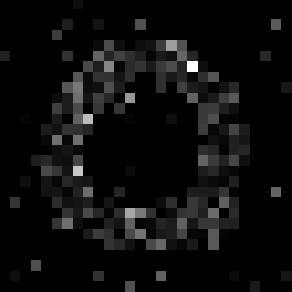}}
\subfigure[\scriptsize SBL (output)]{\includegraphics[width=3.4cm, height=3cm]{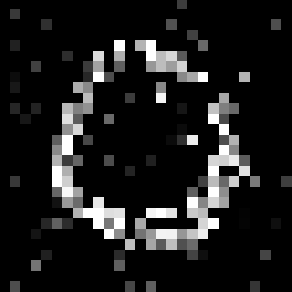}}
\subfigure[\scriptsize SBL (difference)]{\includegraphics[width=3.4cm, height=3cm]{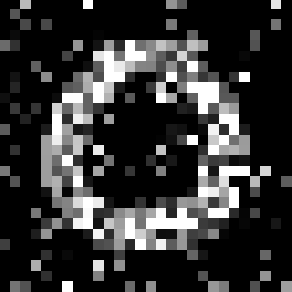}}
\subfigure[\scriptsize MMP (output)]{\includegraphics[width=3.4cm, height=3cm]{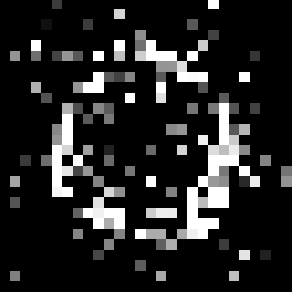}}
\subfigure[\scriptsize MMP (difference)]{\includegraphics[width=3.4cm, height=3cm]{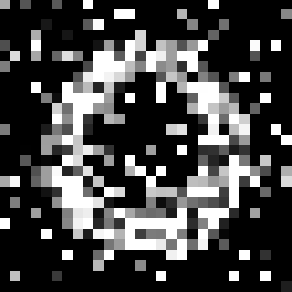}}
\subfigure[\scriptsize Lasso (output)]{\includegraphics[width=3.4cm, height=3cm]{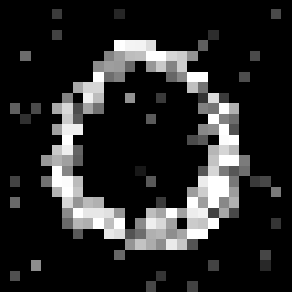}}
\subfigure[\scriptsize Lasso (difference)]{\includegraphics[width=3.4cm, height=3cm]{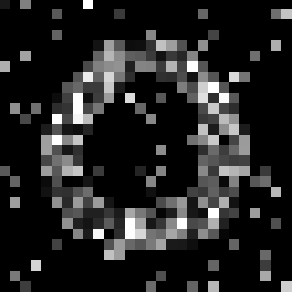}}
\subfigure[\scriptsize IHT (output)]{\includegraphics[width=3.4cm, height=3cm]{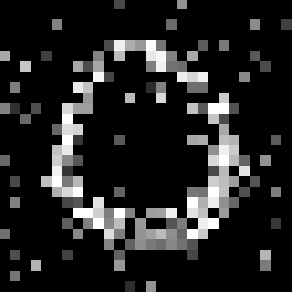}}
\subfigure[\scriptsize IHT (difference)]{\includegraphics[width=3.4cm, height=3cm]{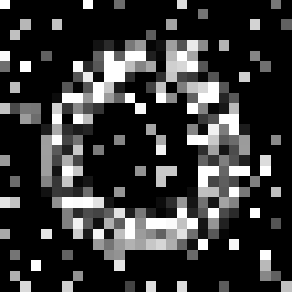}}
\subfigure[\scriptsize SP (output)]{\includegraphics[width=3.4cm, height=3cm]{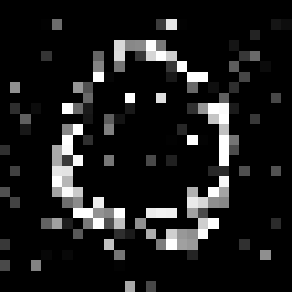}}
\subfigure[\scriptsize SP (difference)]{\includegraphics[width=3.4cm, height=3cm]{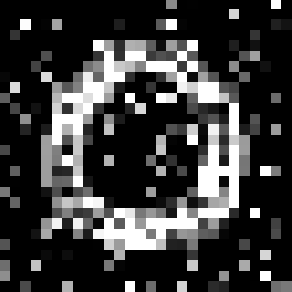}}
\subfigure[\scriptsize CoSaMP (output)]{\includegraphics[width=3.4cm, height=3cm]{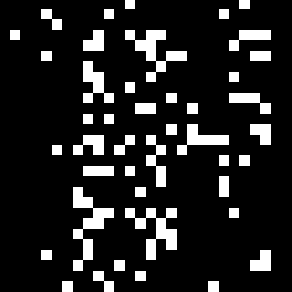}}
\subfigure[\scriptsize CoSaMP (difference)]{\includegraphics[width=3.4cm, height=3cm]{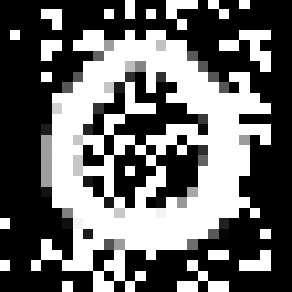}}
\subfigure[LVAMP (output)]{\includegraphics[width=3.4cm, height=3cm]{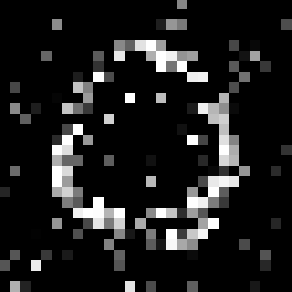}}
\subfigure[LVAMP (difference)]{\includegraphics[width=3.4cm, height=3cm]{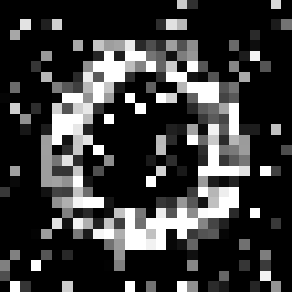}}
\caption{Example for reconstructing a MNIST image}
\label{mnist_ex1}
\end{center}
\end{figure}

\begin{figure}
\begin{center}
\subfigure[\scriptsize Original image]{\includegraphics[width=3.4cm, height=3cm]{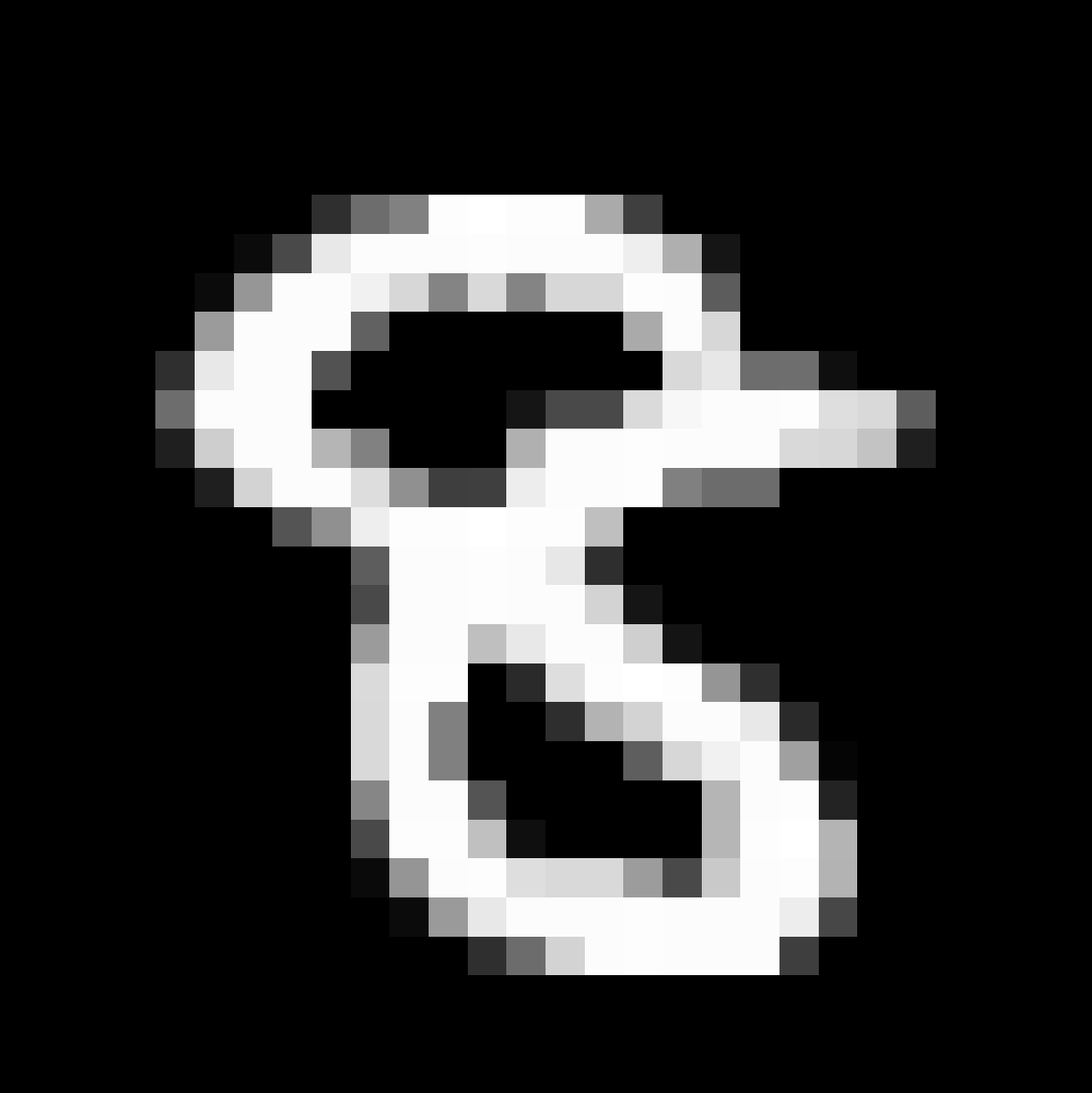}}
\subfigure[\scriptsize TSN (output)]{\includegraphics[width=3.4cm, height=3cm]{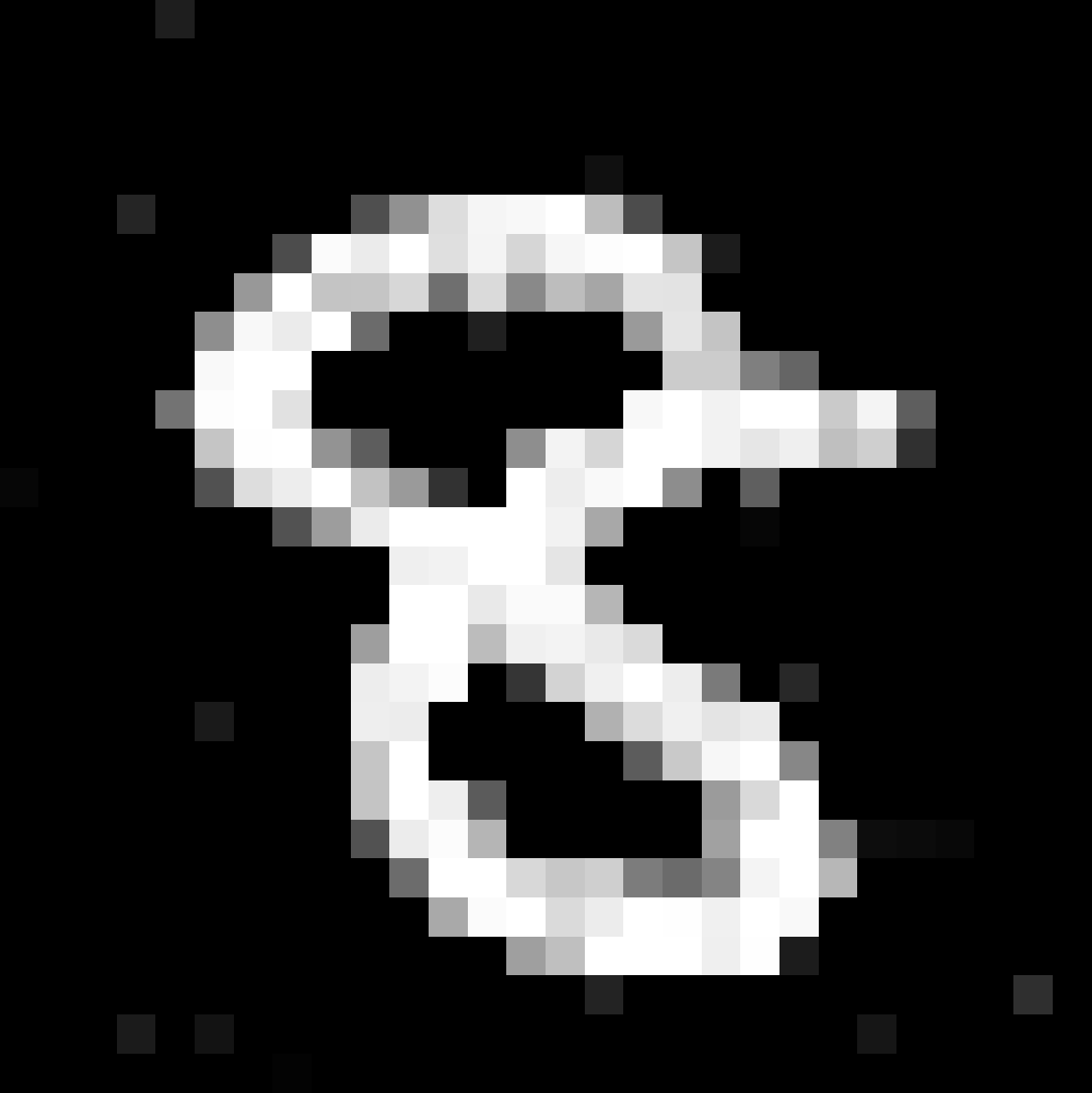}}
\subfigure[\scriptsize TSN (difference)]{\includegraphics[width=3.4cm, height=3cm]{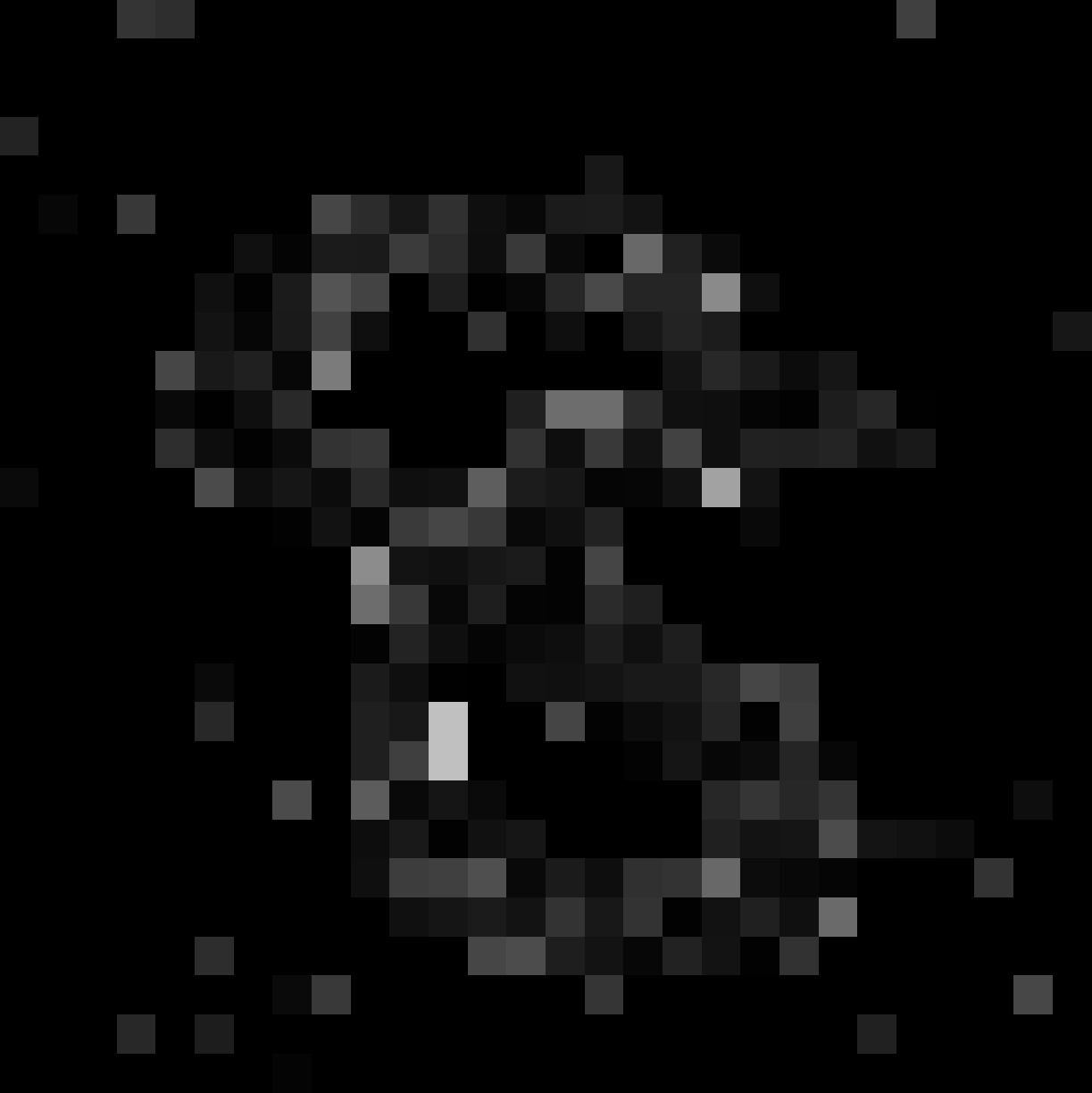}}
\subfigure[\scriptsize GFLSTM (output)]{\includegraphics[width=3.4cm, height=3cm]{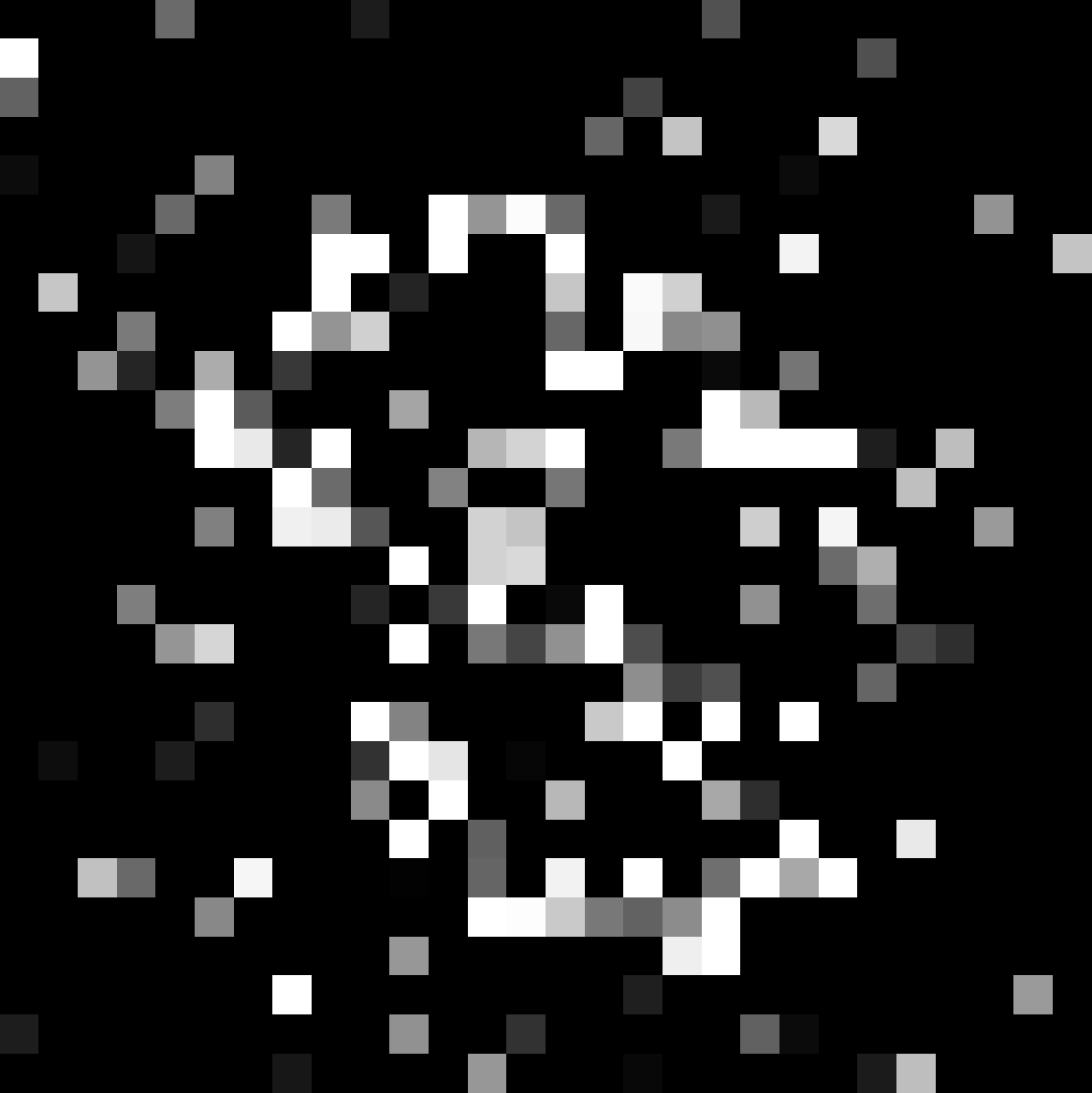}}
\subfigure[\scriptsize GFLSTM (difference)]{\includegraphics[width=3.4cm, height=3cm]{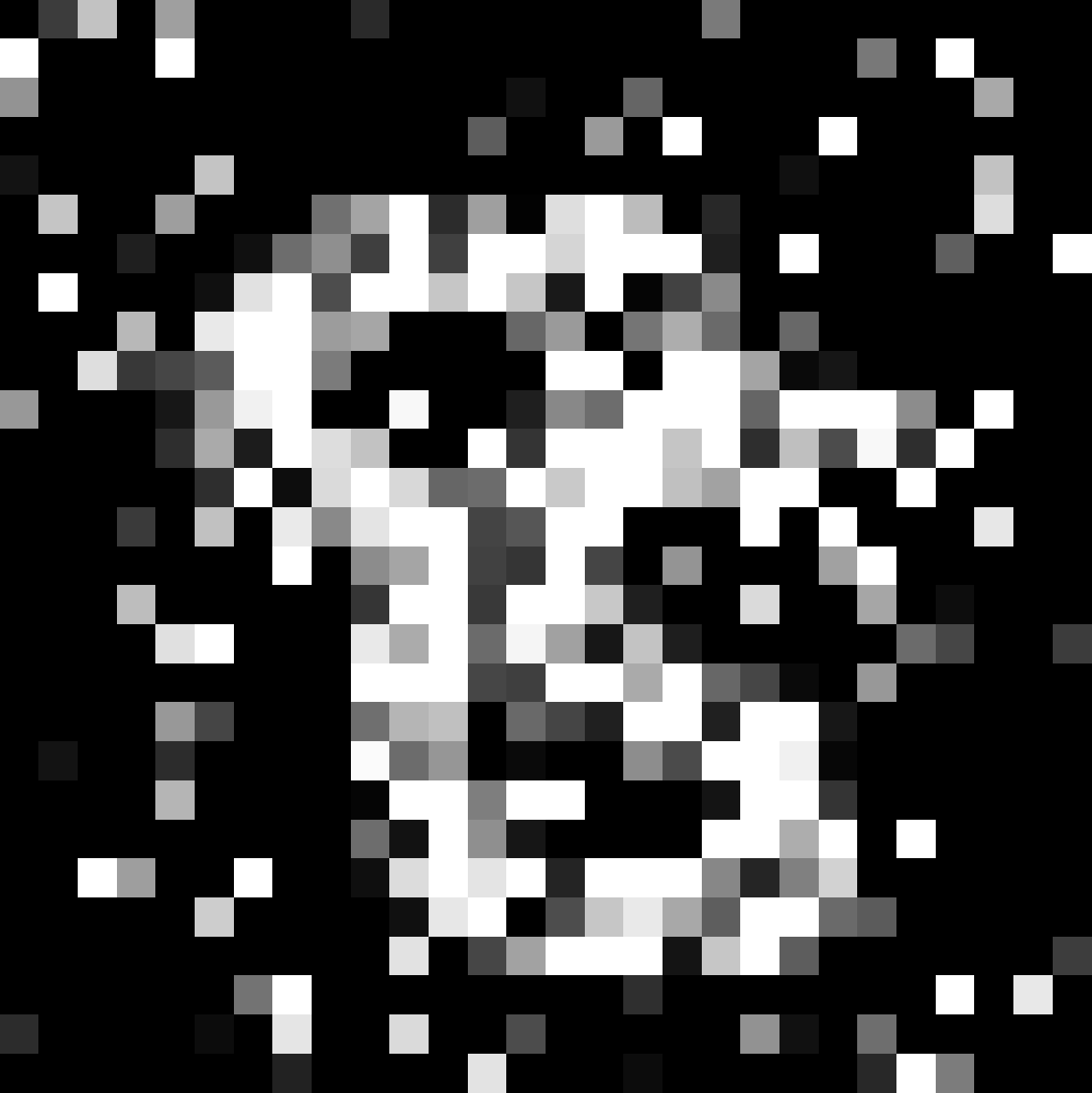}}
\subfigure[\scriptsize SBL (output)]{\includegraphics[width=3.4cm, height=3cm]{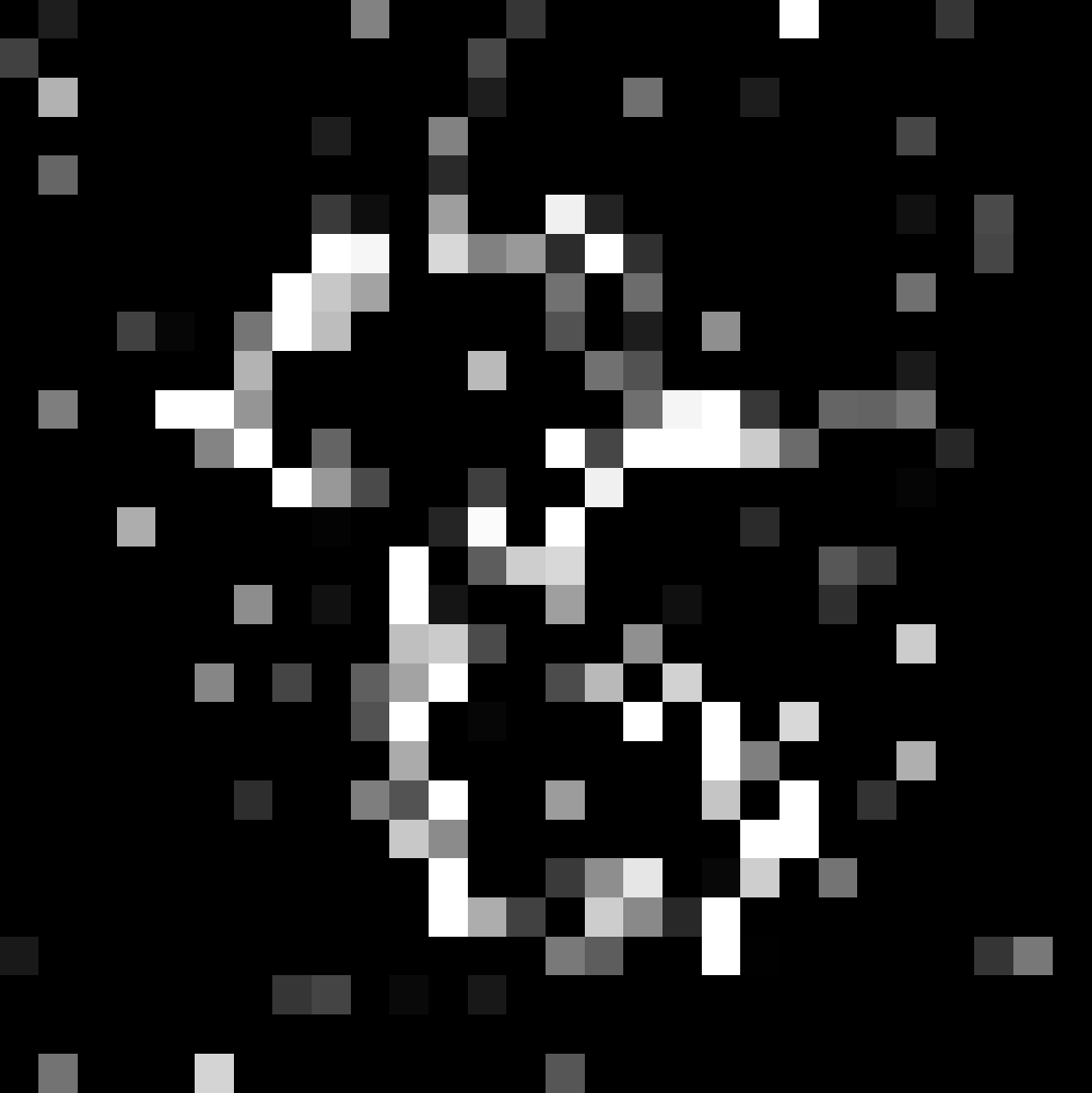}}
\subfigure[\scriptsize SBL (difference)]{\includegraphics[width=3.4cm, height=3cm]{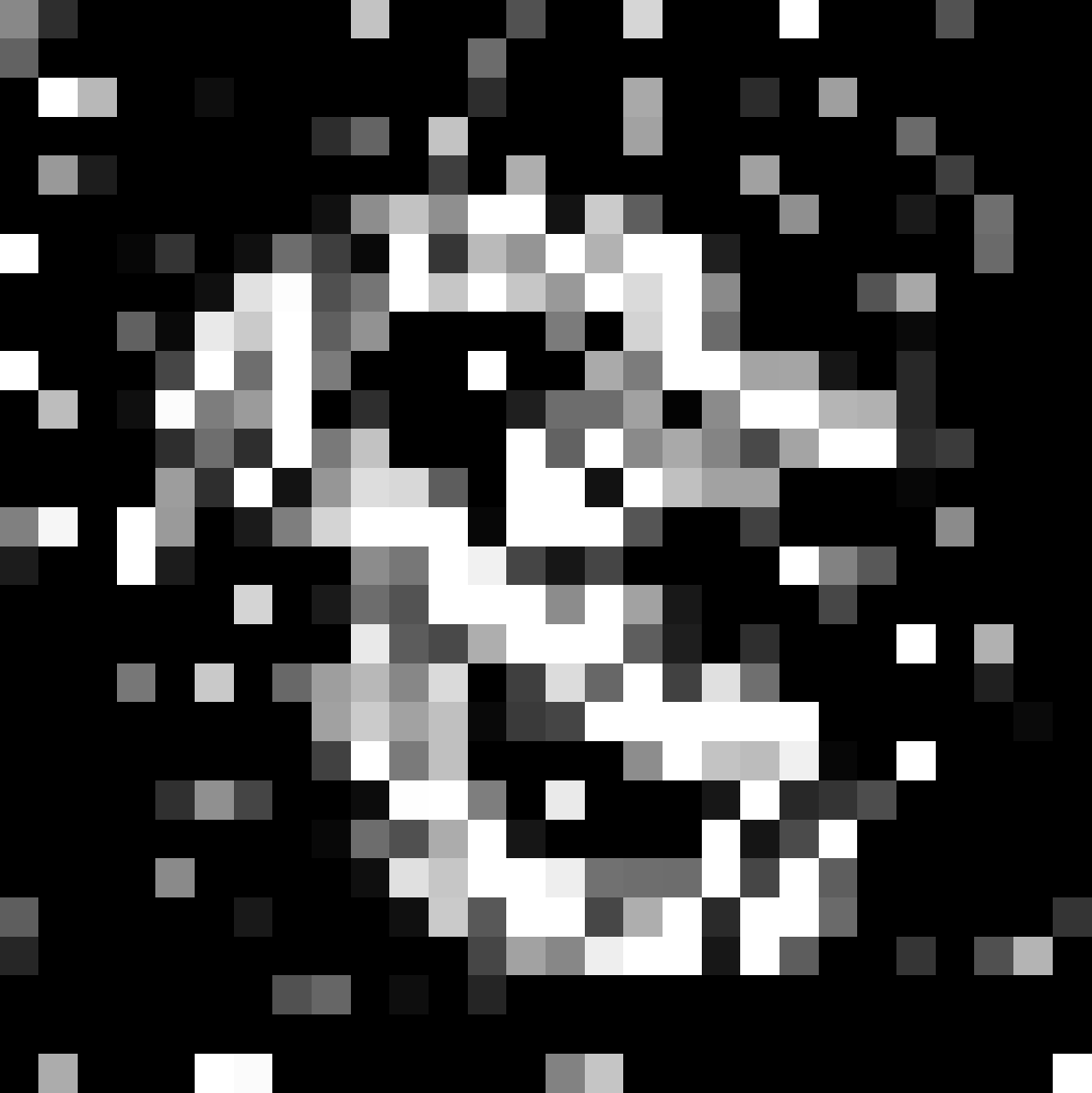}}
\subfigure[\scriptsize MMP (output)]{\includegraphics[width=3.4cm, height=3cm]{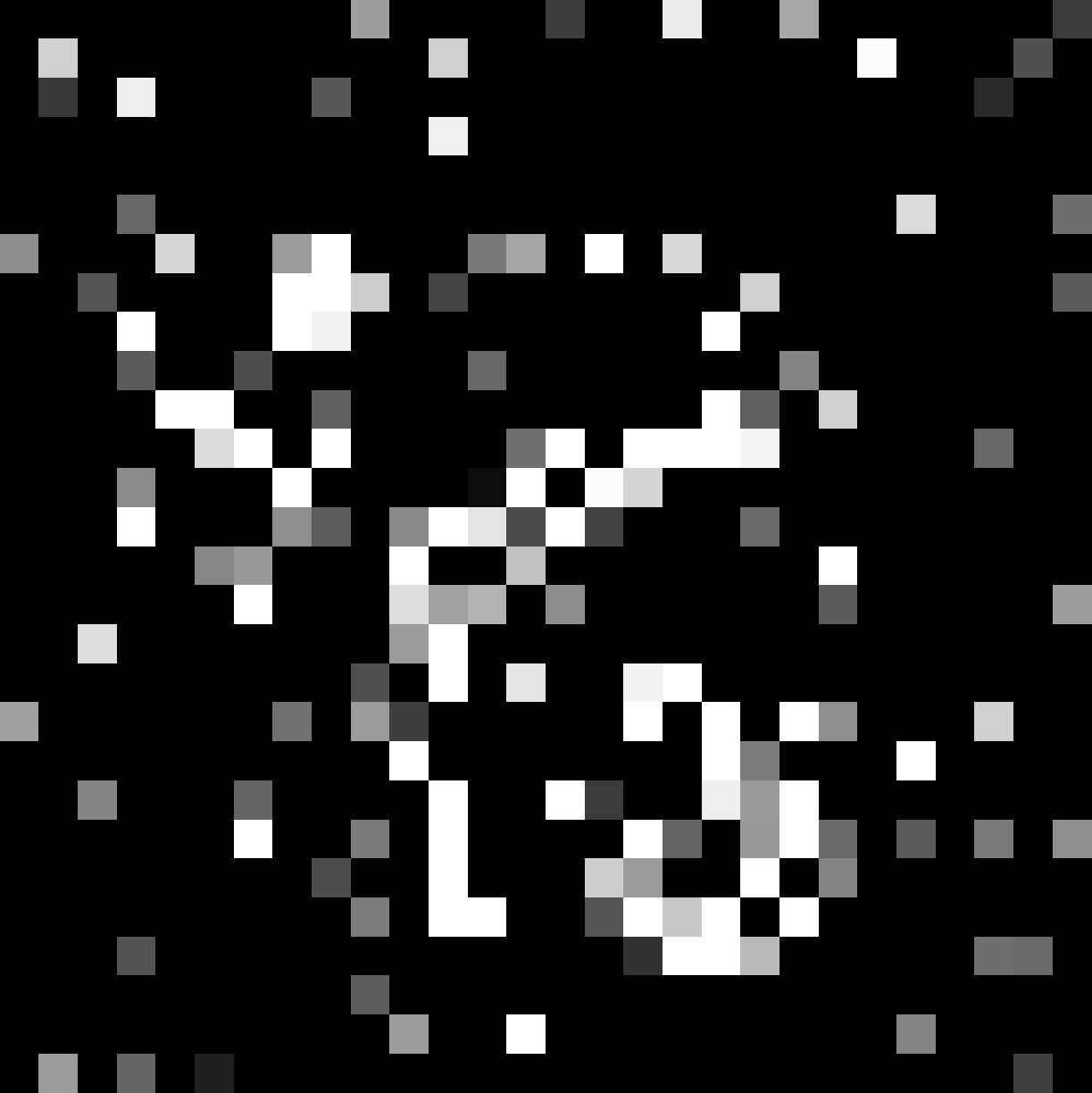}}
\subfigure[\scriptsize MMP (difference)]{\includegraphics[width=3.4cm, height=3cm]{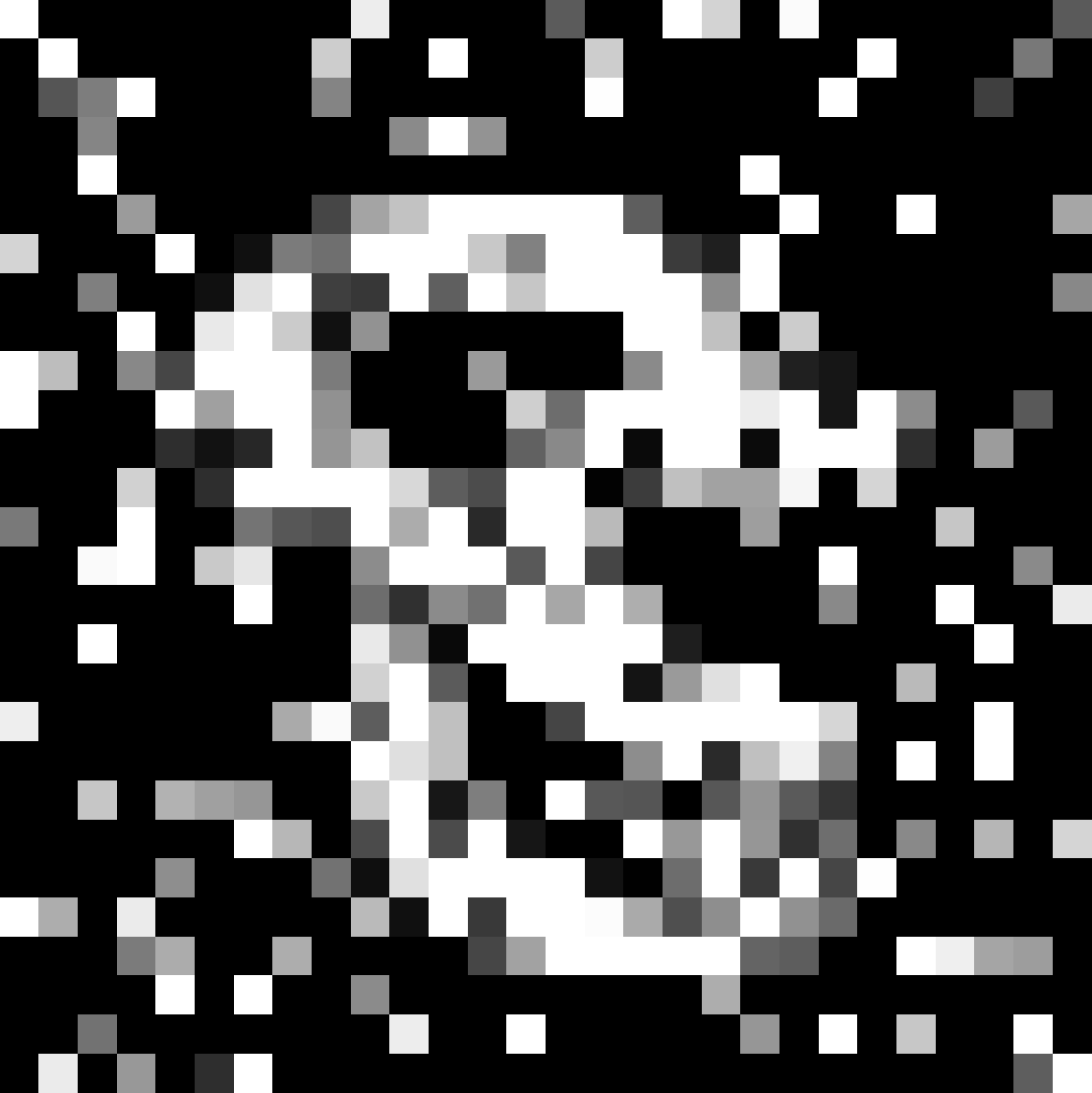}}
\subfigure[\scriptsize Lasso (output)]{\includegraphics[width=3.4cm, height=3cm]{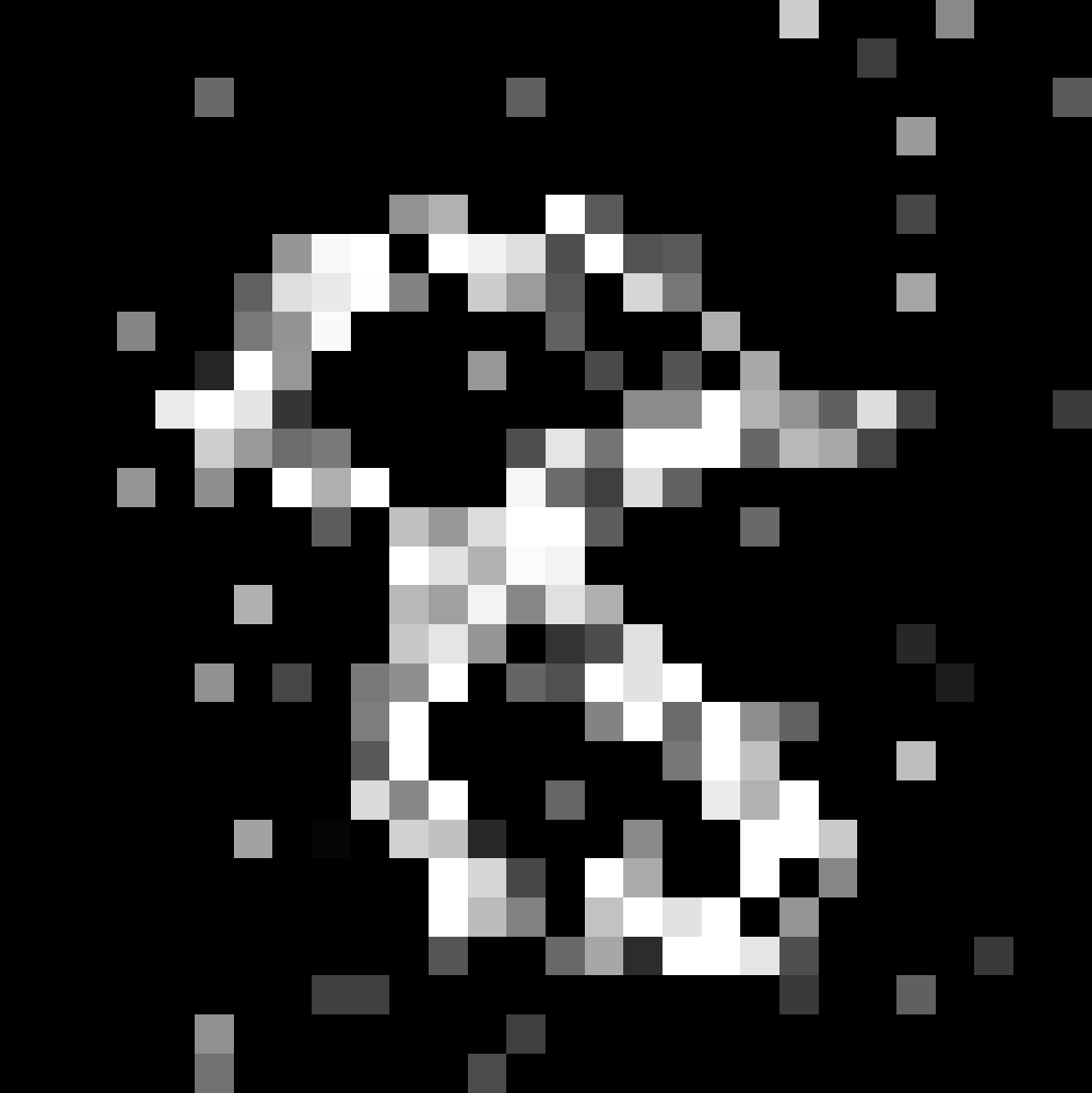}}
\subfigure[\scriptsize Lasso (difference)]{\includegraphics[width=3.4cm, height=3cm]{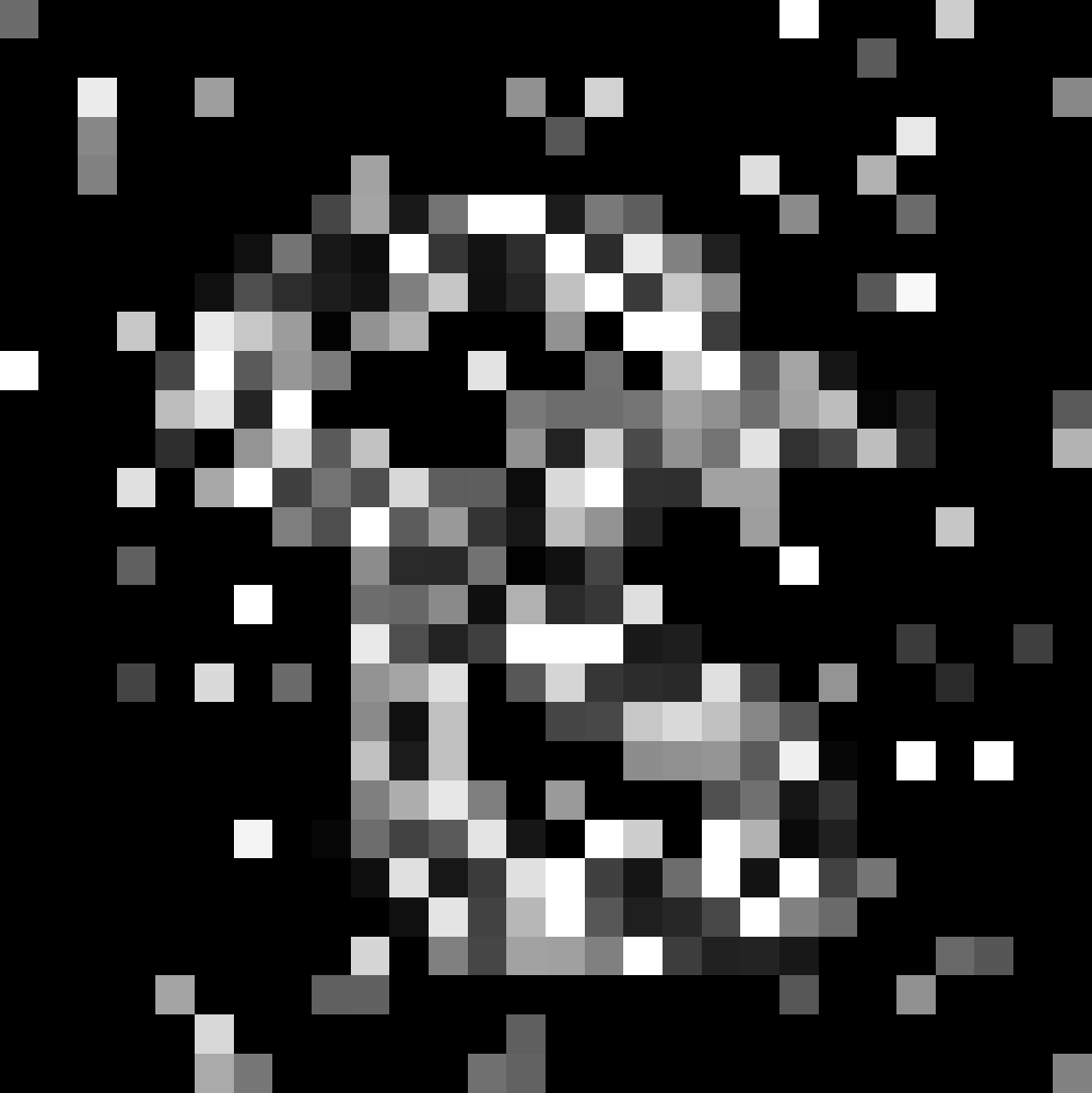}}
\subfigure[\scriptsize IHT (output)]{\includegraphics[width=3.4cm, height=3cm]{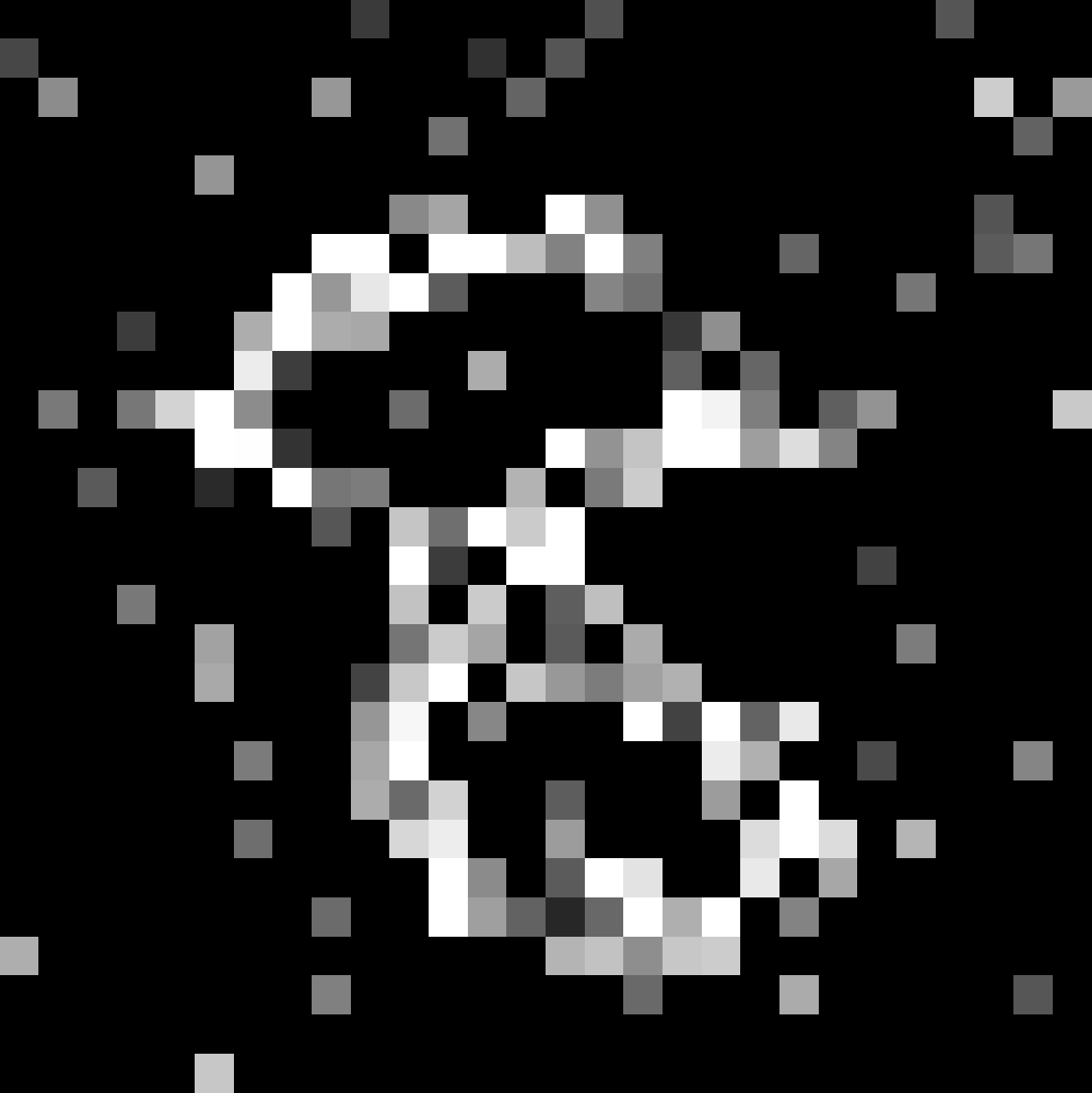}}
\subfigure[\scriptsize IHT (difference)]{\includegraphics[width=3.4cm, height=3cm]{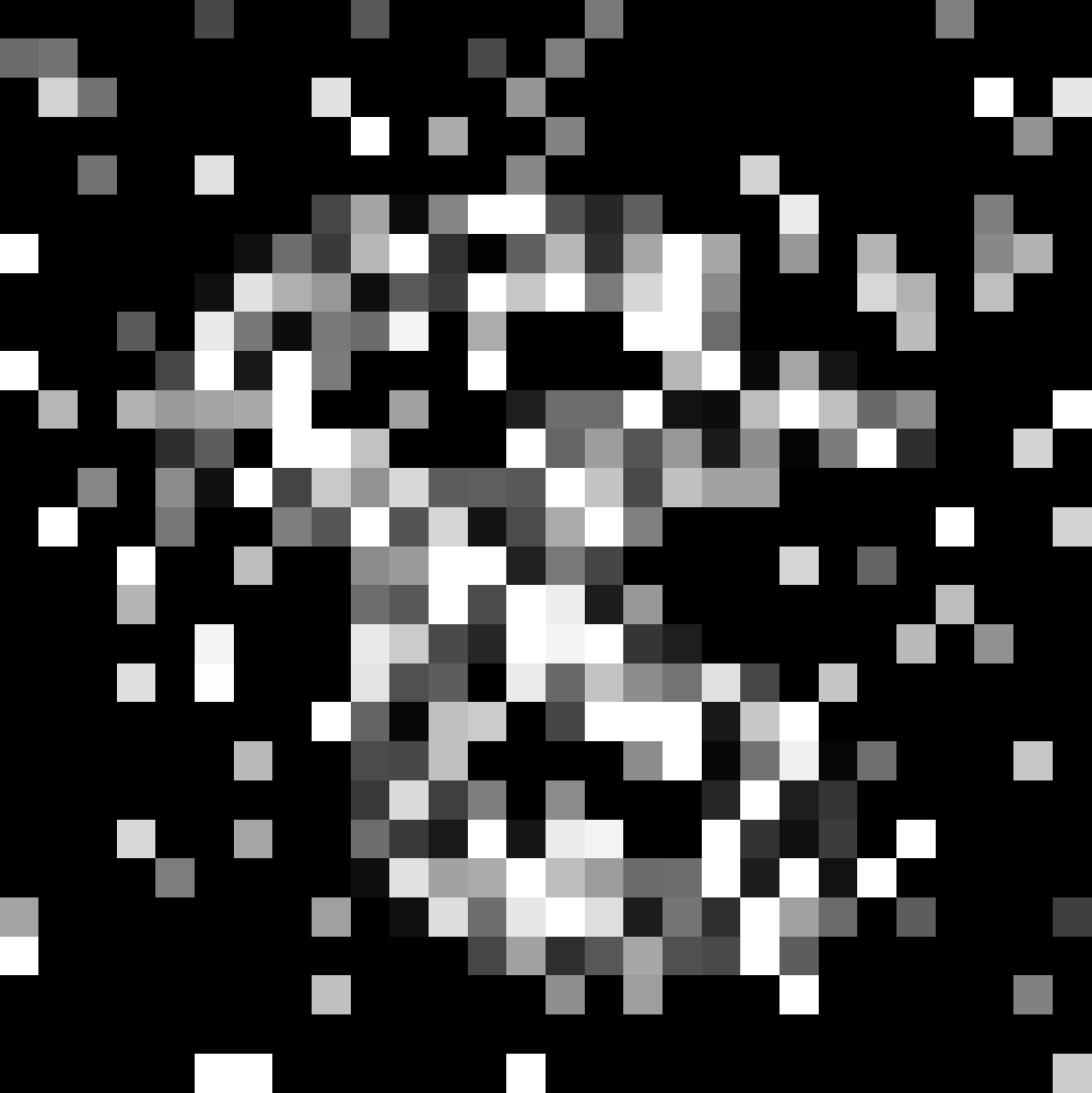}}
\subfigure[\scriptsize SP (output)]{\includegraphics[width=3.4cm, height=3cm]{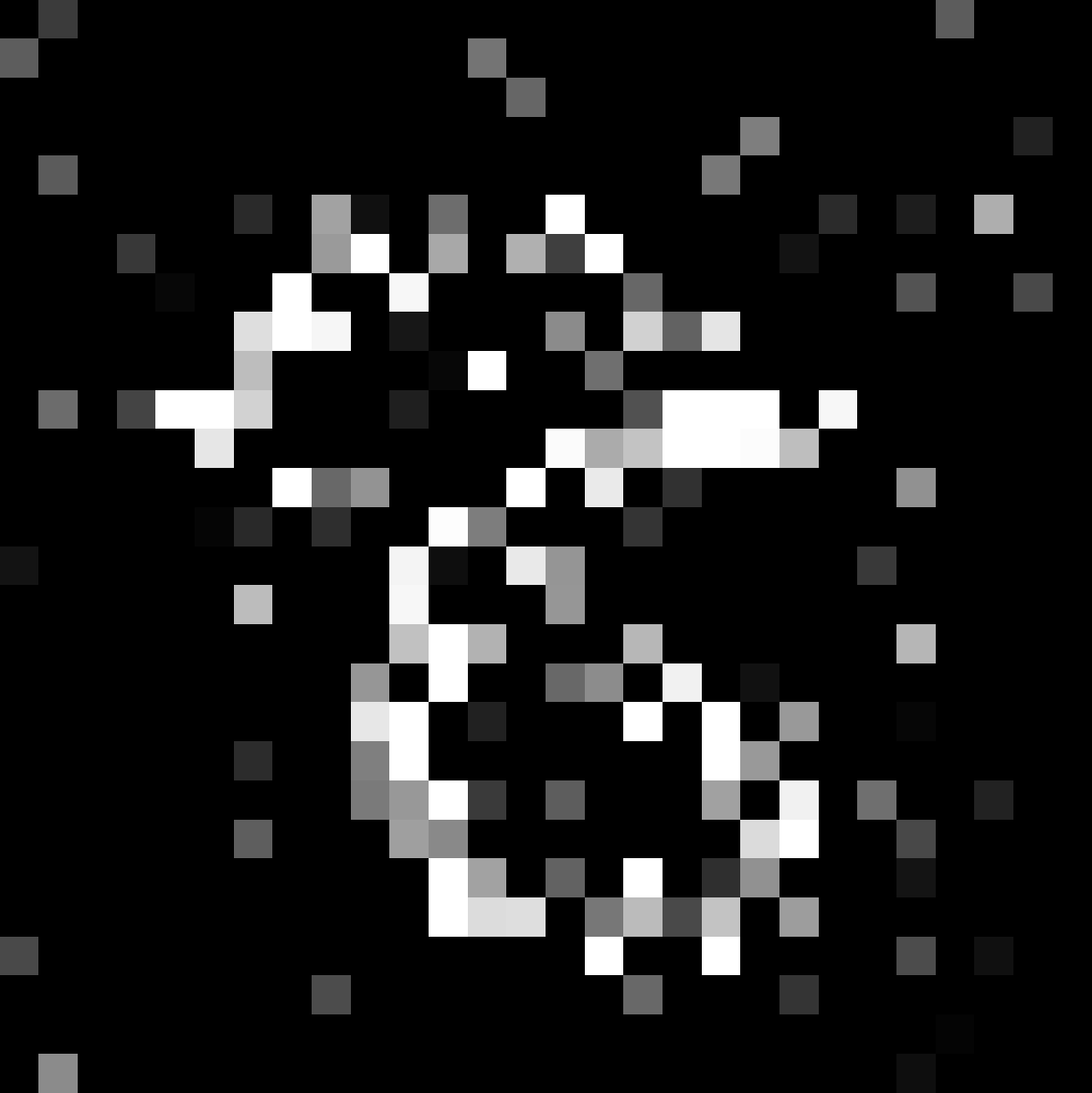}}
\subfigure[\scriptsize SP (difference)]{\includegraphics[width=3.4cm, height=3cm]{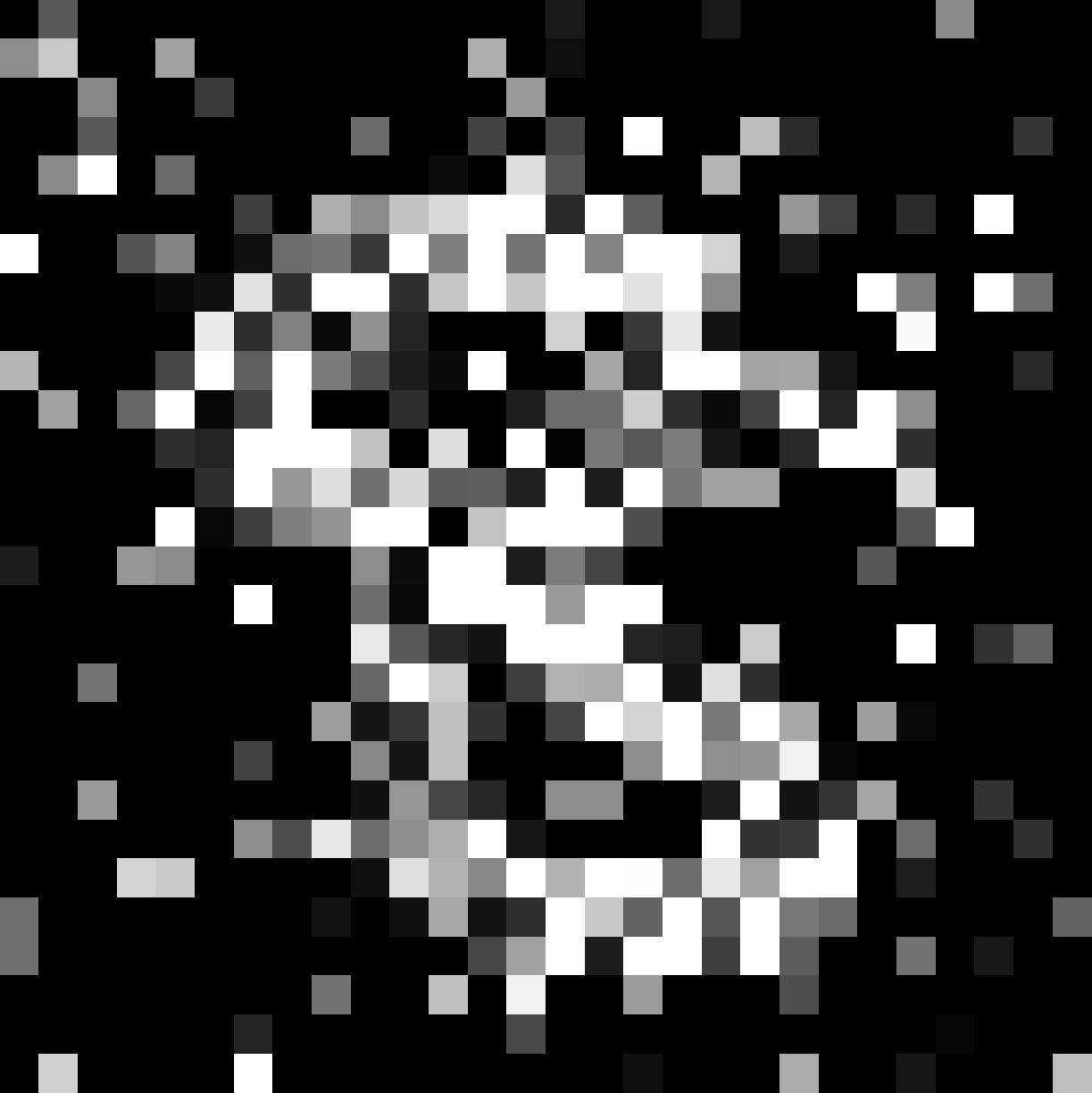}}
\subfigure[\scriptsize CoSaMP (output)]{\includegraphics[width=3.4cm, height=3cm]{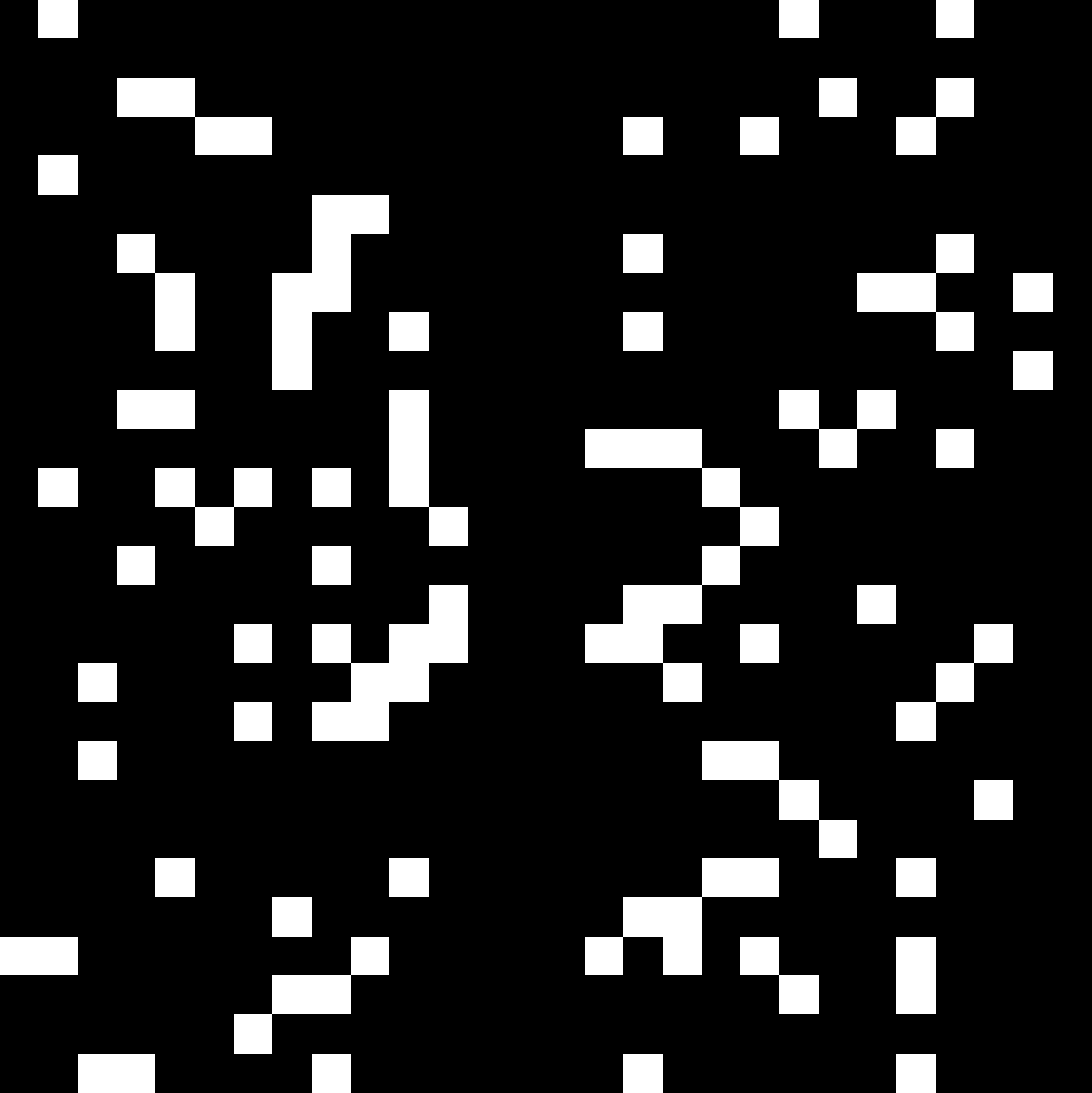}}
\subfigure[\scriptsize CoSaMP (difference)]{\includegraphics[width=3.4cm, height=3cm]{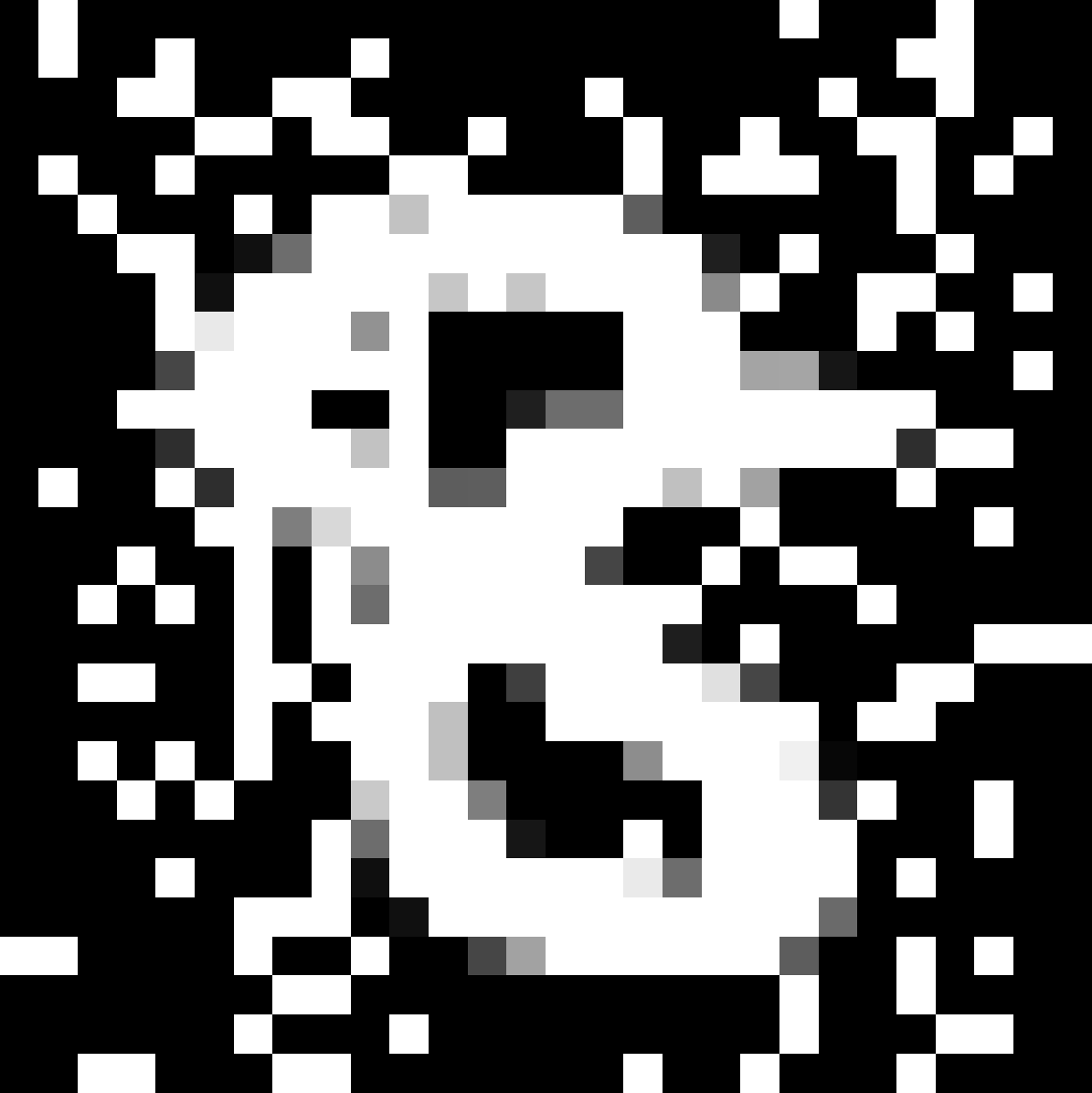}}
\subfigure[LVAMP (output)]{\includegraphics[width=3.4cm, height=3cm]{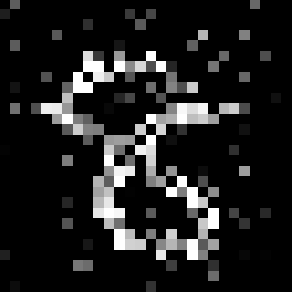}}
\subfigure[LVAMP (difference)]{\includegraphics[width=3.4cm, height=3cm]{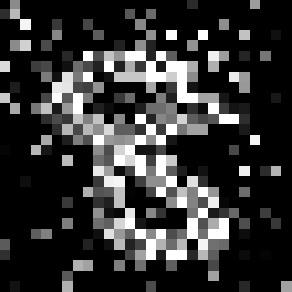}}
\caption{Example for reconstructing a MNIST image}
\label{mnist_ex2}
\end{center}
\end{figure}

\begin{figure}
\begin{center}
\subfigure[\scriptsize Original image]{\includegraphics[width=3.4cm, height=3cm]{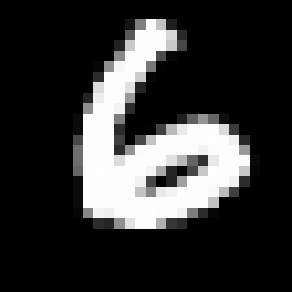}}
\subfigure[\scriptsize TSN (output)]{\includegraphics[width=3.4cm, height=3cm]{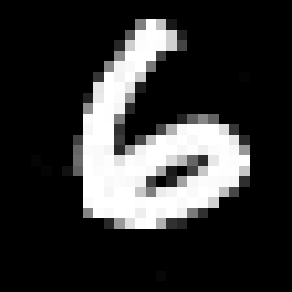}}
\subfigure[\scriptsize TSN (difference)]{\includegraphics[width=3.4cm, height=3cm]{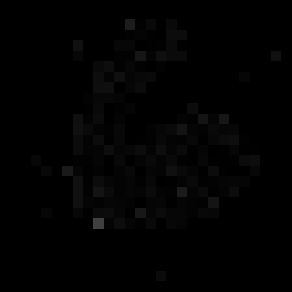}}
\subfigure[\scriptsize GFLSTM (output)]{\includegraphics[width=3.4cm, height=3cm]{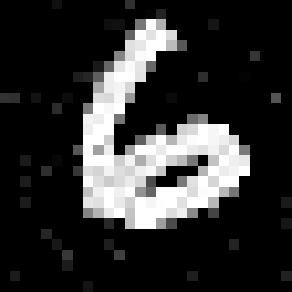}}
\subfigure[\scriptsize GFLSTM (difference)]{\includegraphics[width=3.4cm, height=3cm]{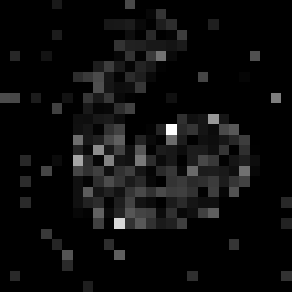}}
\subfigure[\scriptsize SBL (output)]{\includegraphics[width=3.4cm, height=3cm]{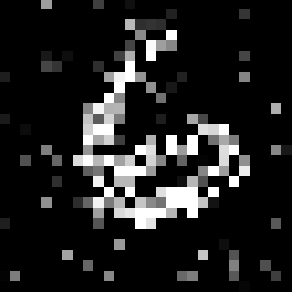}}
\subfigure[\scriptsize SBL (difference)]{\includegraphics[width=3.4cm, height=3cm]{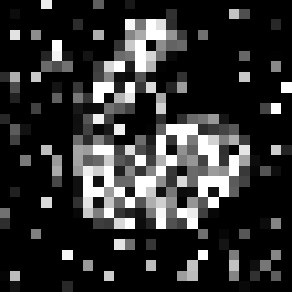}}
\subfigure[\scriptsize MMP (output)]{\includegraphics[width=3.4cm, height=3cm]{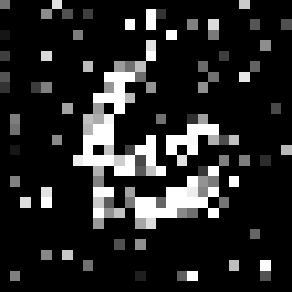}}
\subfigure[\scriptsize MMP (difference)]{\includegraphics[width=3.4cm, height=3cm]{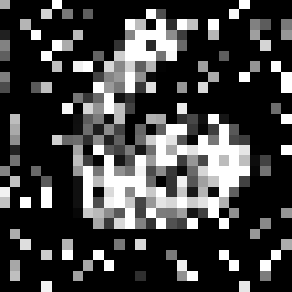}}
\subfigure[\scriptsize Lasso (output)]{\includegraphics[width=3.4cm, height=3cm]{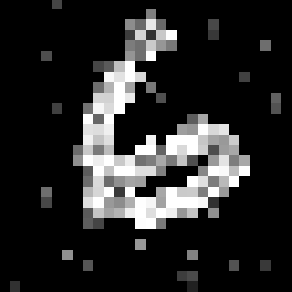}}
\subfigure[\scriptsize Lasso (difference)]{\includegraphics[width=3.4cm, height=3cm]{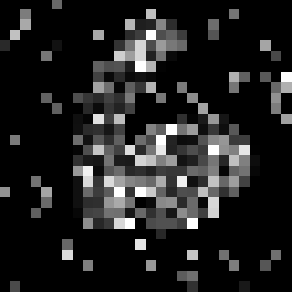}}
\subfigure[\scriptsize IHT (output)]{\includegraphics[width=3.4cm, height=3cm]{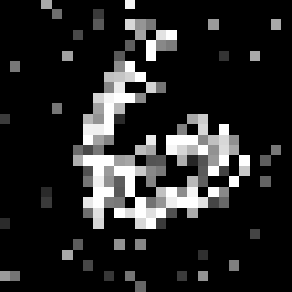}}
\subfigure[\scriptsize IHT (difference)]{\includegraphics[width=3.4cm, height=3cm]{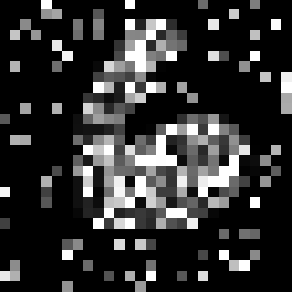}}
\subfigure[\scriptsize SP (output)]{\includegraphics[width=3.4cm, height=3cm]{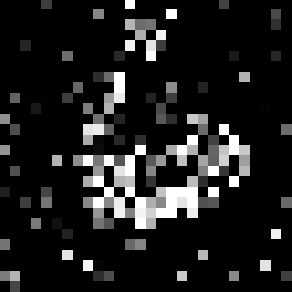}}
\subfigure[\scriptsize SP (difference)]{\includegraphics[width=3.4cm, height=3cm]{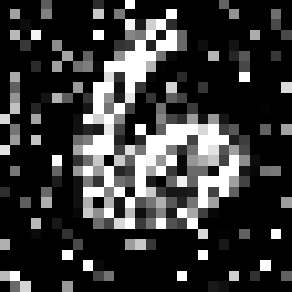}}
\subfigure[\scriptsize CoSaMP (output)]{\includegraphics[width=3.4cm, height=3cm]{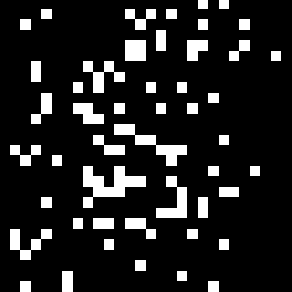}}
\subfigure[\scriptsize CoSaMP (difference)]{\includegraphics[width=3.4cm, height=3cm]{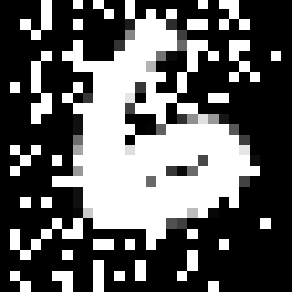}}
\subfigure[LVAMP (output)]{\includegraphics[width=3.4cm, height=3cm]{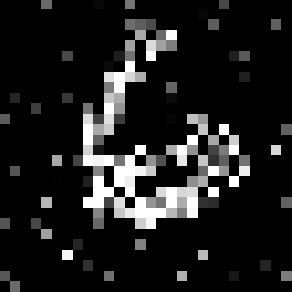}}
\subfigure[LVAMP (difference)]{\includegraphics[width=3.4cm, height=3cm]{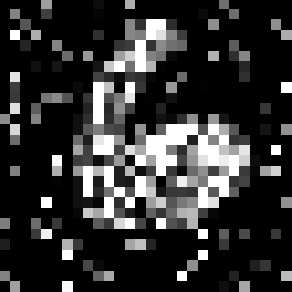}}
\caption{Example for reconstructing a MNIST image}
\label{mnist_ex3}
\end{center}
\end{figure}

\begin{figure}
\begin{center}
\subfigure[\scriptsize Original image]{\includegraphics[width=3.4cm, height=3cm]{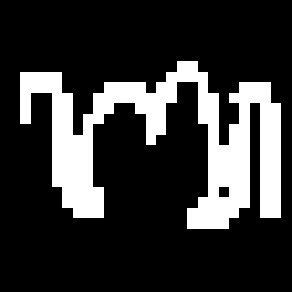}}
\subfigure[\scriptsize TSN (output)]{\includegraphics[width=3.4cm, height=3cm]{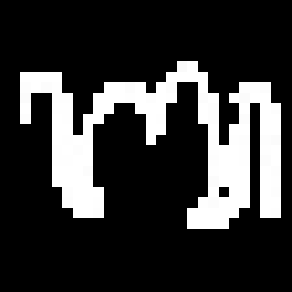}}
\subfigure[\scriptsize TSN (difference)]{\includegraphics[width=3.4cm, height=3cm]{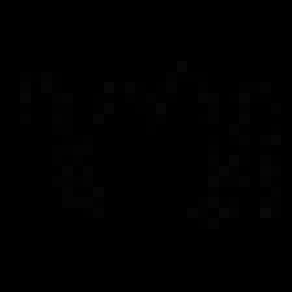}}
\subfigure[\scriptsize GFLSTM (output)]{\includegraphics[width=3.4cm, height=3cm]{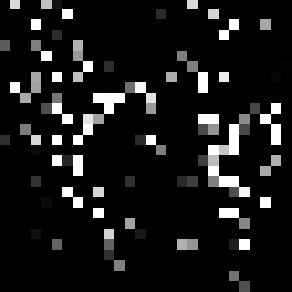}}
\subfigure[\scriptsize GFLSTM (difference)]{\includegraphics[width=3.4cm, height=3cm]{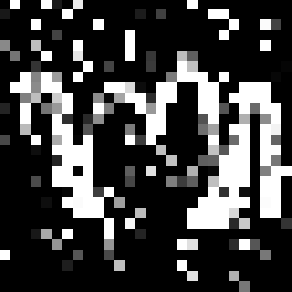}}
\subfigure[\scriptsize SBL (output)]{\includegraphics[width=3.4cm, height=3cm]{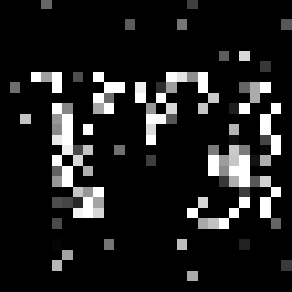}}
\subfigure[\scriptsize SBL (difference)]{\includegraphics[width=3.4cm, height=3cm]{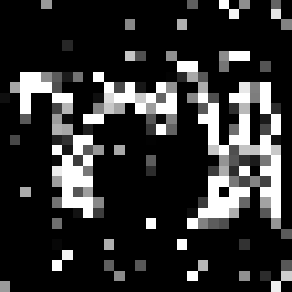}}
\subfigure[\scriptsize MMP (output)]{\includegraphics[width=3.4cm, height=3cm]{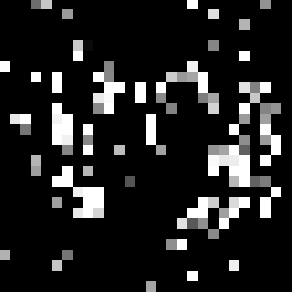}}
\subfigure[\scriptsize MMP (difference)]{\includegraphics[width=3.4cm, height=3cm]{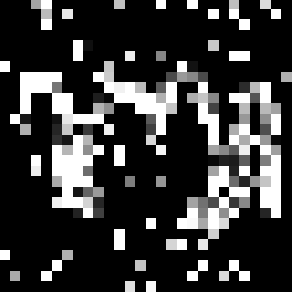}}
\subfigure[\scriptsize Lasso (output)]{\includegraphics[width=3.4cm, height=3cm]{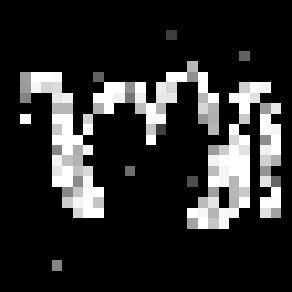}}
\subfigure[\scriptsize Lasso (difference)]{\includegraphics[width=3.4cm, height=3cm]{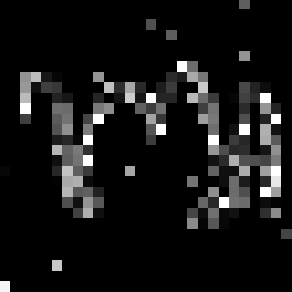}}
\subfigure[\scriptsize IHT (output)]{\includegraphics[width=3.4cm, height=3cm]{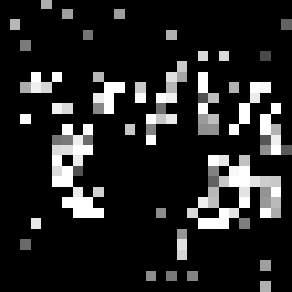}}
\subfigure[\scriptsize IHT (difference)]{\includegraphics[width=3.4cm, height=3cm]{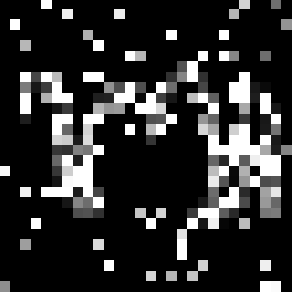}}
\subfigure[\scriptsize SP (output)]{\includegraphics[width=3.4cm, height=3cm]{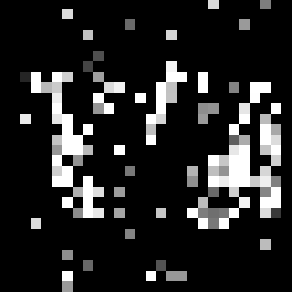}}
\subfigure[\scriptsize SP (difference)]{\includegraphics[width=3.4cm, height=3cm]{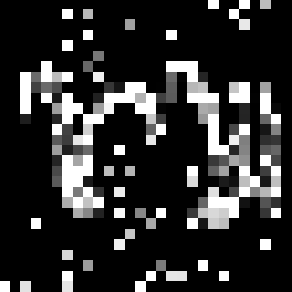}}
\subfigure[\scriptsize CoSaMP (output)]{\includegraphics[width=3.4cm, height=3cm]{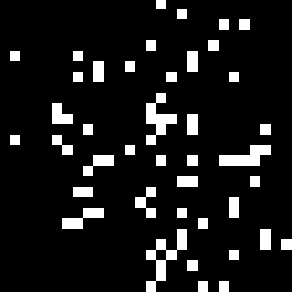}}
\subfigure[\scriptsize CoSaMP (difference)]{\includegraphics[width=3.4cm, height=3cm]{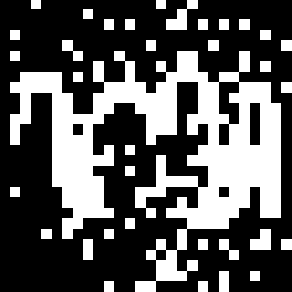}}
\subfigure[LVAMP (output)]{\includegraphics[width=3.4cm, height=3cm]{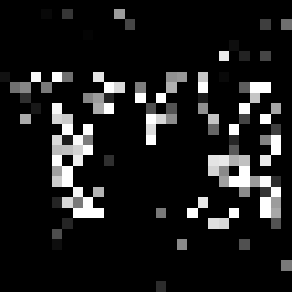}}
\subfigure[LVAMP (difference)]{\includegraphics[width=3.4cm, height=3cm]{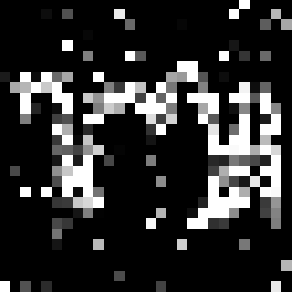}}
\caption{Example for reconstructing an OMNIGLOT image}
\label{omni_ex1}
\end{center}
\end{figure}

\begin{figure}
\begin{center}
\subfigure[\scriptsize Original image]{\includegraphics[width=3.4cm, height=3cm]{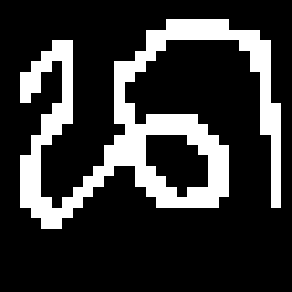}}
\subfigure[\scriptsize TSN (output)]{\includegraphics[width=3.4cm, height=3cm]{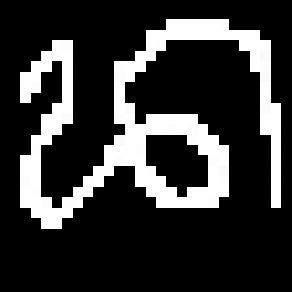}}
\subfigure[\scriptsize TSN (difference)]{\includegraphics[width=3.4cm, height=3cm]{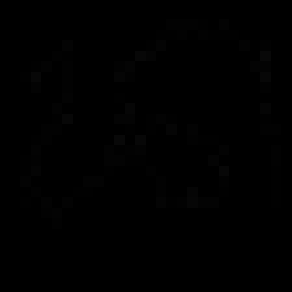}}
\subfigure[\scriptsize GFLSTM (output)]{\includegraphics[width=3.4cm, height=3cm]{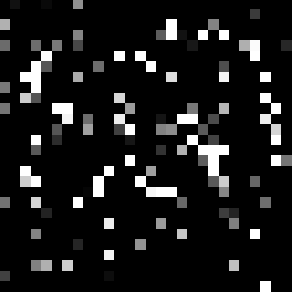}}
\subfigure[\scriptsize GFLSTM (difference)]{\includegraphics[width=3.4cm, height=3cm]{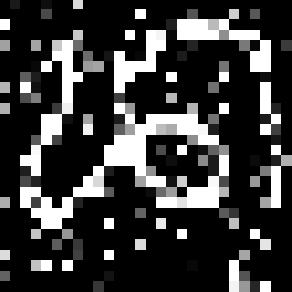}}
\subfigure[\scriptsize SBL (output)]{\includegraphics[width=3.4cm, height=3cm]{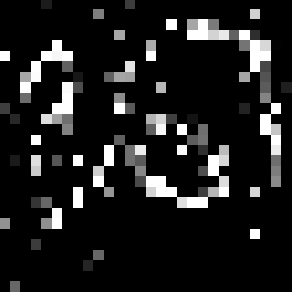}}
\subfigure[\scriptsize SBL (difference)]{\includegraphics[width=3.4cm, height=3cm]{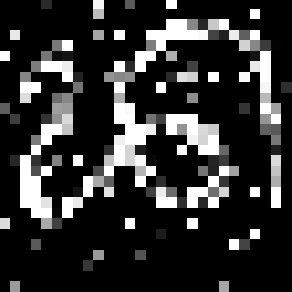}}
\subfigure[\scriptsize MMP (output)]{\includegraphics[width=3.4cm, height=3cm]{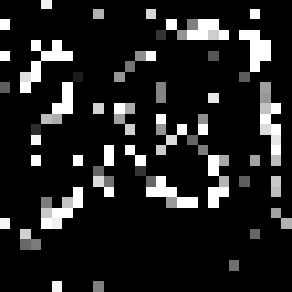}}
\subfigure[\scriptsize MMP (difference)]{\includegraphics[width=3.4cm, height=3cm]{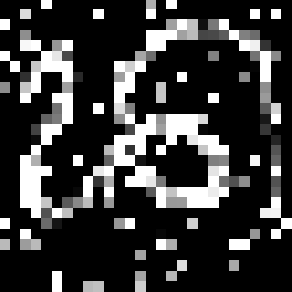}}
\subfigure[\scriptsize Lasso (output)]{\includegraphics[width=3.4cm, height=3cm]{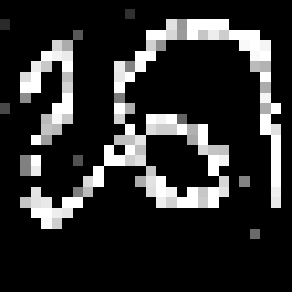}}
\subfigure[\scriptsize Lasso (difference)]{\includegraphics[width=3.4cm, height=3cm]{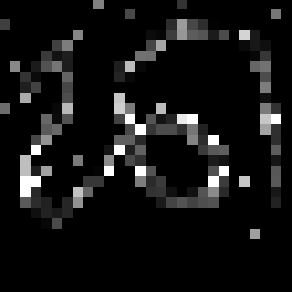}}
\subfigure[\scriptsize IHT (output)]{\includegraphics[width=3.4cm, height=3cm]{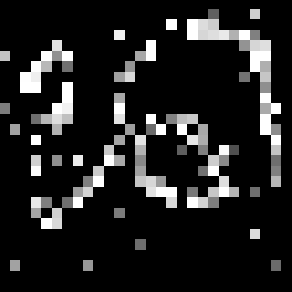}}
\subfigure[\scriptsize IHT (difference)]{\includegraphics[width=3.4cm, height=3cm]{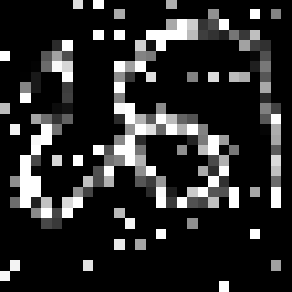}}
\subfigure[\scriptsize SP (output)]{\includegraphics[width=3.4cm, height=3cm]{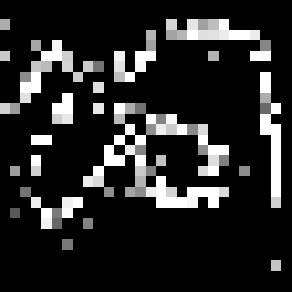}}
\subfigure[\scriptsize SP (difference)]{\includegraphics[width=3.4cm, height=3cm]{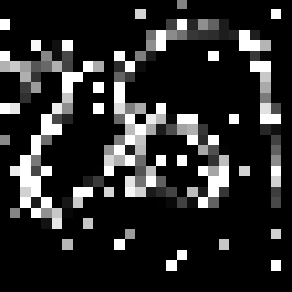}}
\subfigure[\scriptsize CoSaMP (output)]{\includegraphics[width=3.4cm, height=3cm]{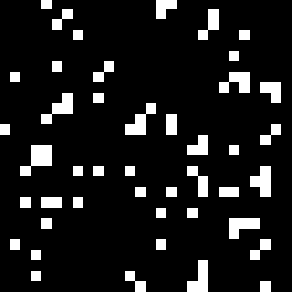}}
\subfigure[\scriptsize CoSaMP (difference)]{\includegraphics[width=3.4cm, height=3cm]{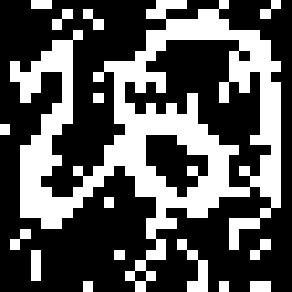}}
\subfigure[LVAMP (output)]{\includegraphics[width=3.4cm, height=3cm]{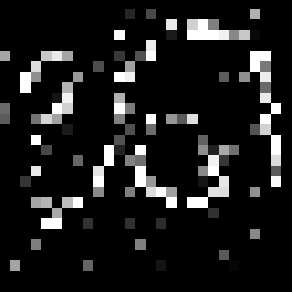}}
\subfigure[LVAMP (difference)]{\includegraphics[width=3.4cm, height=3cm]{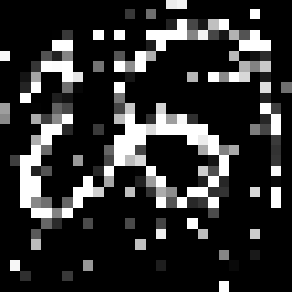}}
\caption{Example for reconstructing an OMNIGLOT image}
\label{omni_ex2}
\end{center}
\end{figure}

\begin{figure}
\begin{center}
\subfigure[\scriptsize Original image]{\includegraphics[width=3.4cm, height=3cm]{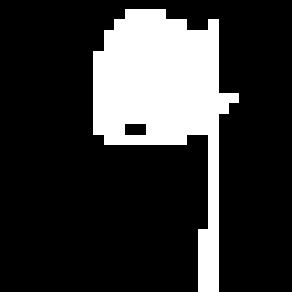}}
\subfigure[\scriptsize TSN (output)]{\includegraphics[width=3.4cm, height=3cm]{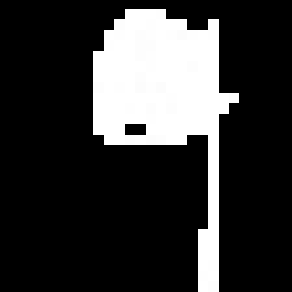}}
\subfigure[\scriptsize TSN (difference)]{\includegraphics[width=3.4cm, height=3cm]{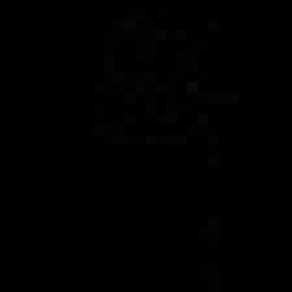}}
\subfigure[\scriptsize GFLSTM (output)]{\includegraphics[width=3.4cm, height=3cm]{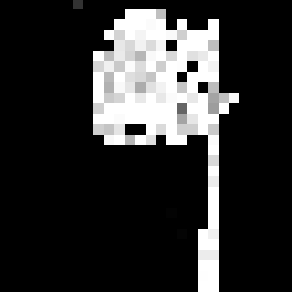}}
\subfigure[\scriptsize GFLSTM (difference)]{\includegraphics[width=3.4cm, height=3cm]{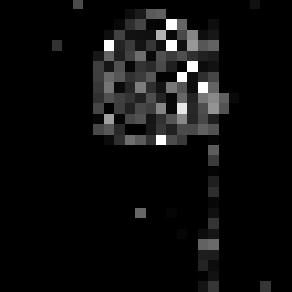}}
\subfigure[\scriptsize SBL (output)]{\includegraphics[width=3.4cm, height=3cm]{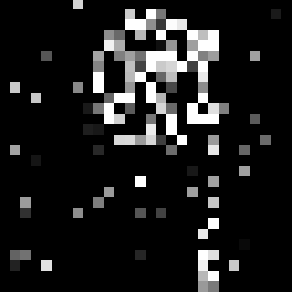}}
\subfigure[\scriptsize SBL (difference)]{\includegraphics[width=3.4cm, height=3cm]{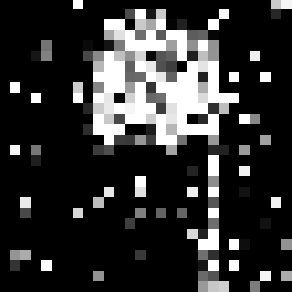}}
\subfigure[\scriptsize MMP (output)]{\includegraphics[width=3.4cm, height=3cm]{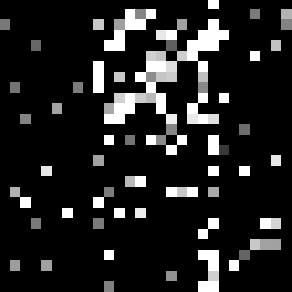}}
\subfigure[\scriptsize MMP (difference)]{\includegraphics[width=3.4cm, height=3cm]{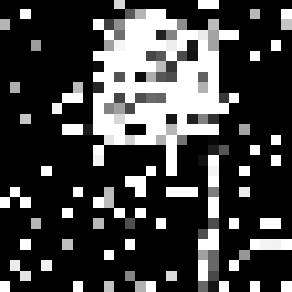}}
\subfigure[\scriptsize Lasso (output)]{\includegraphics[width=3.4cm, height=3cm]{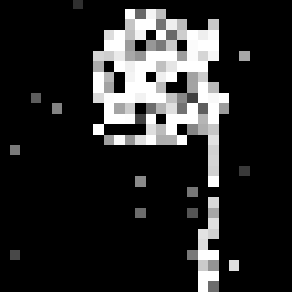}}
\subfigure[\scriptsize Lasso (difference)]{\includegraphics[width=3.4cm, height=3cm]{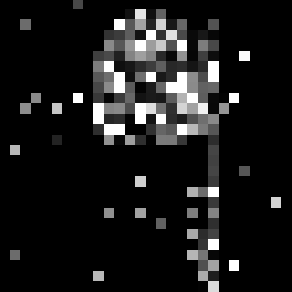}}
\subfigure[\scriptsize IHT (output)]{\includegraphics[width=3.4cm, height=3cm]{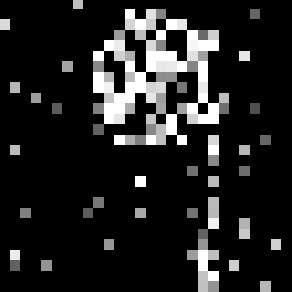}}
\subfigure[\scriptsize IHT (difference)]{\includegraphics[width=3.4cm, height=3cm]{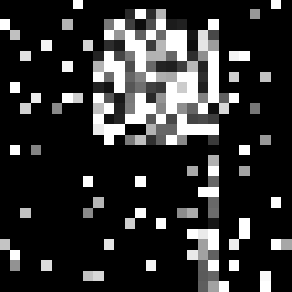}}
\subfigure[\scriptsize SP (output)]{\includegraphics[width=3.4cm, height=3cm]{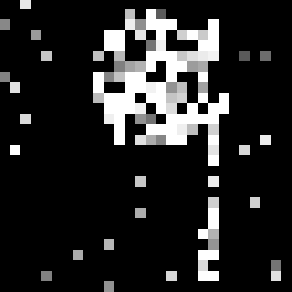}}
\subfigure[\scriptsize SP (difference)]{\includegraphics[width=3.4cm, height=3cm]{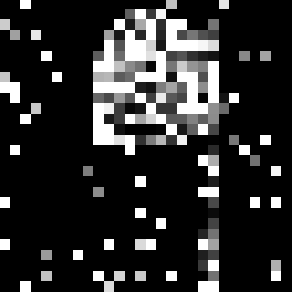}}
\subfigure[\scriptsize CoSaMP (output)]{\includegraphics[width=3.4cm, height=3cm]{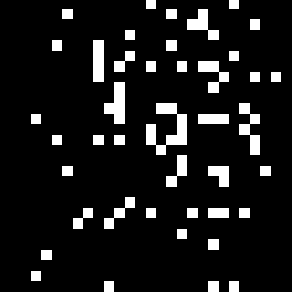}}
\subfigure[\scriptsize CoSaMP (difference)]{\includegraphics[width=3.4cm, height=3cm]{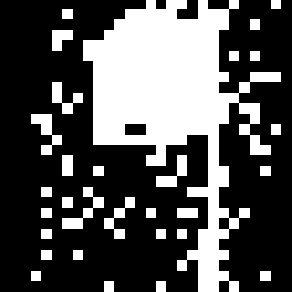}}
\subfigure[LVAMP (output)]{\includegraphics[width=3.4cm, height=3cm]{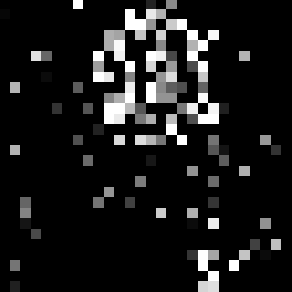}}
\subfigure[LVAMP (difference)]{\includegraphics[width=3.4cm, height=3cm]{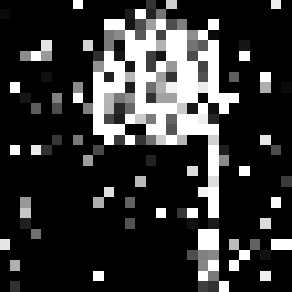}}
\caption{Example for reconstructing an OMNIGLOT image}
\label{omni_ex3}
\end{center}
\end{figure}

We adjusted the size of OMNIGLOT to $28 \times 28$ pixels. We have observed that 95\% of the MNIST or the resized OMNIGLOT images have sparsity below 215 and 160, respectively. Consequently, we randomly sampled the MNIST images with sparsity 215 and the OMNIGLOT images with sparsity 160, and evaluated the performance of SR algorithms. Figures \ref{mnist_ex1}--\ref{mnist_ex3} and  Figures \ref{omni_ex1}--\ref{omni_ex3} show some MNIST and OMNIGLOT images reconstructed by each SR algorithm, respectively.\footnote{We evaluated the performance of  TSN against the existing SR algorithms shown in Section \ref{ne} except for MMP$_2$, because of its long execution time; MMP shown in Table \ref{table_mnist} and Figures \ref{mnist_ex1}--\ref{mnist_ex3} indicates MMP$_1$ shown in Section \ref{ne}.} Images labeled with `difference' indicate the difference between the output image generated from the corresponding SR algorithm and the original image. Similar performance was observed for other MNIST or OMNIGLOT images. 

We used the GFLSTM network for DNN $f_{\theta}(\cdot)$ in TSN and trained the network using Algorithm \ref{alg1}, whose input tuple $(k_1,k_2,s_d,s_b,n_e,v_{\textup{SNR$_{\textup{dB}}$}})$ was set to $(1,250,6 \cdot 10^5,250,400,20)$. Note that DNN $f_{\theta}(\cdot)$ in TSN was trained without using MNIST images. Therefore, we assumed that each signal vector used as training data in Algorithm \ref{alg1} was generated such that each of its nonzero elements is independently and uniformly sampled from $0$ to $1$.\footnote{During testing, we normalized the MNIST image by dividing its elements by 256 to obtain values between 0 and 1.} We evaluated the proposed TSN (Algorithm \ref{alg5}) using the following set, $\boldsymbol{\tau}$, of input parameters:
\begin{align}\nonumber
\scriptsize
&\boldsymbol{\tau}:(q,z,\epsilon,\boldsymbol{l},\boldsymbol{g},t_{\textup{max}}) = (200,110,\bar \epsilon,(1,1),(1,1),\infty), 
\end{align}
where $\bar \epsilon:= \max[(\left \| y\right \| \cdot 10^{-\bar v_\textup{SNR$_{\textup{dB}}$}/20}),10^{-5}]$ is the signal error bound and $\bar v_\textup{SNR$_{\textup{dB}}$}=25$ is the SNR in decibels.  


\newpage
\bibliographystyle{IEEEtran}
\bibliography{IEEEabrv,bibdata3}

\end{document}